\documentclass[journal]{IEEEtran}
\usepackage{balance}
\usepackage{amsmath}
\usepackage{graphicx}
\usepackage{pdfpages}
\usepackage[noadjust]{cite}
\usepackage{subfig}
\usepackage[T1]{fontenc} 
\usepackage{array}
\usepackage{url}
\usepackage{float}

\usepackage{color}
\usepackage{wrapfig}
\usepackage{lipsum}
\usepackage[hidelinks]{hyperref}
\usepackage{multirow}
\usepackage{mathtools}
\usepackage{gensymb}
\usepackage{booktabs}

\begin{document}
\title{Color-based Segmentation of Sky/Cloud Images From Ground-based Cameras}
\author{Soumyabrata~Dev,~\IEEEmembership{Student Member,~IEEE},
        Yee~Hui~Lee,~\IEEEmembership{Senior Member,~IEEE},\\
        and Stefan~Winkler,~\IEEEmembership{Senior~Member, IEEE}
\thanks{Manuscript received 11-Dec-2015; revised 17-Mar-2016; accepted 09-Apr-2016.}        
\thanks{S. Dev and Y. H. Lee are with the School of Electrical and Electronic Engineering, Nanyang Technological University, Singapore (e-mail: soumyabr001@e.ntu.edu.sg, EYHLee@ntu.edu.sg). }
\thanks{S. Winkler is with the Advanced Digital Sciences Center (ADSC), University of Illinois at Urbana-Champaign, Singapore (e-mail: \mbox{Stefan.Winkler@adsc.com.sg}).}
\thanks{Send correspondence to S. Winkler, E-mail: Stefan.Winkler@adsc.com.sg.}
}

\markboth{IEEE Journal of Selected Topics in Applied Earth Observations and Remote Sensing, ~Vol.~XX, No.~XX, XX~2016}%
{Shell \MakeLowercase{\textit{et al.}}: Bare Demo of IEEEtran.cls for Journals}

\maketitle

\begin{abstract}
Sky/cloud images captured by ground-based cameras (a.k.a.\ whole sky imagers) are increasingly used nowadays because of their applications in a number of fields, including climate modeling, weather prediction, renewable energy generation, and satellite communications. Due to the wide variety of cloud types and lighting conditions in such images, accurate and robust segmentation of clouds is  challenging. In this paper, we present a supervised segmentation framework for ground-based sky/cloud images based on a systematic analysis of different color spaces and components, using partial least squares (PLS) regression. Unlike other state-of-the-art methods, our proposed approach is entirely learning-based and does not require any manually-defined parameters. In addition, we release the \textbf{S}ingapore \textbf{W}hole Sky \textbf{IM}aging \textbf{SEG}mentation Database (SWIMSEG), a large database of annotated sky/cloud images, to the research community.
\end{abstract}


\IEEEpeerreviewmaketitle

\setlength{\fboxsep}{0pt}
\setlength{\fboxrule}{0.2pt}

\section{Introduction}
\IEEEPARstart {C}{louds} have been extensively studied in the research community over the past few decades. The analysis of clouds and their features is important for a wide variety of applications. For example, it has been used for nowcasting to deliver accurate weather forecasts \cite{nowcast}, rainfall and satellite precipitation estimates~\cite{SP_estimate}, in the study of contrails~\cite{contra}, and various other day-to-day meteorological applications~\cite{met_app}.  Yuan et al.\ have been investigating the clouds' vertical structure \cite{cloud_model_compare} and cloud attenuation for optimizing satellite links \cite{radiosonde2014}.

Sky/cloud imaging can be performed in different ways. Satellite imagery \cite{PALSAR,sat_reg} and aerial photographs~\cite{air1} are popular in particular for large-scale surveys; airborne light detection and ranging (LiDAR) data are extensively used for aerial surveys~\cite{LIDAR1}. However, these techniques rarely provide sufficient temporal and/or spatial resolution for localized and short-term cloud analysis over a particular area. This is where ground-based whole sky imagers (WSIs) offer a compelling alternative~\cite{WAHRSIS}. The images obtained from these devices provide high-resolution data about local cloud formation, movement, and other atmospheric phenomena. 

Segmentation is one of the first steps in sky/cloud image analysis.  It remains a challenging task because of the non-rigid, feature-less, and poorly-defined structure of clouds, whose shape also changes continuously over time. Thus, classical image segmentation approaches based on shape priors \cite{wrapper_Jain} are not suitable. Furthermore, the wide range of lighting conditions (direct sunlight to completely covered skies) adds to the difficulty.

\subsection{Related Work}
As color is the most discriminating feature in sky/cloud images, most works in the literature use color for cloud segmentation. Long et al.\ \cite{Long} showed that the ratio of red and blue channels from RGB color space is a good candidate for segmentation and tuned corresponding thresholds to create binary masks. Heinle et al.\ \cite{Heinle2010} exploited the difference of red and blue channels for successful detection and subsequent labeling of pixels. Liu et al.\ \cite{LiuSP2015} also used the difference of red and blue channels in their superpixel-based cloud segmentation framework. Souza et al.\ \cite{Souza} used the Saturation (S) channel for calculating cloud coverage. Mantelli-Neto et al. \cite{Sylvio} investigated the locus of cloud pixels in the RGB color model. Li et al.\ \cite{Li2011} proposed cloud detection using an adaptive threshold technique in the normalized blue/red channel. Yuan et al.\ ~\cite{BOW-cloud} proposed a cloud detection framework using superpixel classification of image features.

In these existing methods in the literature for cloud segmentation, the selection of color models and channels has not been studied systematically. Many existing approaches \cite{Calbo2008,Heinle2010,LiuSP2015,Li2011,LiuGC2015} use combinations of red and blue channels, which is a sensible choice, because the sky is predominantly blue due to the Rayleigh scattering of light at shorter wavelengths. However, we are not aware of any experimental analysis presented regarding the efficacy of these color channels in sky/cloud image segmentation. Furthermore, all of the above methods rely on manually-defined parameters and case-based decisions for segmentation. These make the methods somewhat ad-hoc and prone to errors. Finally, most of them assign binary labels by design, which further reduces their flexibility and robustness.

\subsection{Outline of Our Contribution}
The motivation of this paper is to propose a robust framework for color-based cloud segmentation under any illumination conditions, including a systematic analysis of color channels. The framework is based on partial least squares (PLS) regression and provides a straightforward, parameter-free supervised segmentation method. We show that our approach is robust and offers a superior performance across two different databases as compared to current state-of-the-art algorithms. Furthermore, it allows annotating each pixel with a degree of \emph{belongingness} to the sky or cloud category, instead of the usual binary labeling.

In our previous work \cite{ICIP1_2014}, we presented an analysis of color channels for sky/cloud images captured by whole-sky cameras, which is an important pre-requisite for better segmentation. The fuzzy c-means clustering method we used in that work however suffers from similar shortcomings as other existing cloud segmentation methods. The main novel contributions of the present manuscript compared to our earlier work include: 
\begin{itemize}
\item Introduction of a large public sky/cloud image database with segmentation masks;
\item Extensive evaluation of color components and selection of appropriate color channels on two different sky/cloud image databases;
\item Robust learning-based framework for sky/cloud segmentation that outperforms existing methods.
\end{itemize}

The rest of this paper is organized as follows. Section \ref{sec:color-spaces} introduces the color spaces under consideration and describes the statistical tools used for subsequent evaluation. Section \ref{sec:prob-segment} discusses the supervised probabilistic segmentation framework. The sky/cloud databases used for evaluation, including our new SWIMSEG database, are presented in Section \ref{sec:database}. An exhaustive analysis of color channels is performed in Section \ref{sec:results}. Section~\ref{sec:result-segment} presents the experimental evaluation of the segmentation framework, followed by a discussion of the results in Section \ref{sec:Discussion}.  Section \ref{sec:conc} concludes the paper.

\section{Approach \& Tools}
\label{sec:color-spaces}

In this section, we describe the color models and channels we consider in this paper and present the statistical tools for evaluating their usefulness in sky/cloud image analysis.
Specifically, we use Principal Component Analysis (PCA) to check the degree of correlation between the color channels and to identify those that capture the most variance.  Loading factors from the primary principal component as well as the Receiver Operating Characteristic (ROC) curve for a simple thresholding applied directly to the color values of a given channel serve as indicators of a channel's suitability for cloud classification.

\subsection{Color Channels}
We consider a set of $16$ color channels and components $c_1, c_2,...,c_{16}$ (see Table \ref{ps}). They comprise color spaces RGB, HSV, YIQ, $\mbox{L}^{*}\mbox{a}^{*}\mbox{b}^{*}$, different red-blue combinations ($R/B$, $R-B$, $\frac{B-R}{B+R}$), and chroma $C=\mbox{max}(R,G,B)-\mbox{min}(R,G,B)$.

\begin{table}[htb]
\small
\centering
\setlength{\tabcolsep}{4pt} 
\begin{tabular}{c|c||c|c||c|c||c|c||c|c||c|c}
  \hline
  $c_{1}$ & R & $c_{4}$ & H & $c_{7}$ & Y & $c_{10}$ & $L^{*}$ & $c_{13}$ & $R/B$ & $c_{16}$ & $C$\\
  $c_{2}$ & G & $c_{5}$ & S & $c_{8}$ & I & $c_{11}$ & $a^{*}$ & $c_{14}$ & $R-B$& $ $ & $ $\\
  $c_{3}$ & B & $c_{6}$ & V & $c_{9}$ & Q & $c_{12}$ & $b^{*}$ & $c_{15}$ & $\frac{B-R}{B+R}$ & $ $ & $ $\\
  \hline
\end{tabular}
\caption{Color spaces and components used for analysis.}
\label{ps}
\end{table}

In the existing literature, mainly color channels $c_{1-9}$ \cite{Souza,Sylvio} and $c_{13-15}$ \cite{Long,Heinle2010,Li2011} have been used for sky/cloud image segmentation. In addition to these, we also consider $\mbox{L}^{*}\mbox{a}^{*}\mbox{b}^{*}$ space ($c_{10-12}$) because of its perceptual uniformity properties as well as chroma ($c_{16}$), because clouds tend to be achromatic.

\subsection{Principal Component Analysis}
\label{sec:pca-theory}
We use Principal Component Analysis (PCA) to determine the underlying structure of an image represented by the $16$ color channels from Table \ref{ps} and analyze the inherent correlations amongst these components. Consider a sample image $\mathbf{X}_{i}$ of dimension $m \times n$ from a dataset consisting of \emph{N} images ($i=1,2,...,N$). The individual color channels $c_{1-16}$ are extracted for $\mathbf{X}_{i}$ and reshaped into column vectors $\mathbf{\widetilde{c}}_j \in {\rm I\!R}^{mn \times 1}$ where $j=1,2,..,16$. The $\mathbf{\widetilde{c}}_j$ obtained from the different color channels of the sample image $\mathbf{X}_{i}$ are stacked alongside, to form the matrix  $\hat{\textbf{X}_{i}} \in {\rm I\!R}^{mn \times 16}$.
\begin{equation}
\label{eq:eq1}
\hat{\textbf{X}_{i}}=[\mathbf{\widetilde{c}}_1, \mathbf{\widetilde{c}}_2,..,\mathbf{\widetilde{c}}_j..,\mathbf{\widetilde{c}}_{16}],\mbox{where } i=1,..,N
\end{equation}

We normalize $\hat{\textbf{X}_{i}}$ with the mean $\bar{c_{j}}$ and standard deviation $\sigma_{c_{j}}$ of the individual color channels of the image $\mathbf{X}_{i}$, and represent it by $\ddot{\mathbf{X}_{i}}$:
\begin{equation}
\label{eq:eq3}
\ddot{\mathbf{X}_{i}}= [\frac{\widetilde{\mathbf{c}_{1}}-\bar{c_{1}}}{\sigma_{c_{1}}}, \frac{\widetilde{\mathbf{c}_{2}}-\bar{c_{2}}}{\sigma_{c_{2}}},..,\frac{\widetilde{\mathbf{c}_{j}}-\bar{c_{j}}}{\sigma_{c_{j}}},..,\frac{\widetilde{\mathbf{c}_{16}}-\bar{c_{16}}}{\sigma_{c_{16}}}]
\end{equation}

Subsequently, we compute the covariance matrix $\mathbf{M}_i$ for each of $\ddot{\mathbf{X}_{i}}$. The covariance matrices $\mathbf{M}_i$ for each of the images provide interesting insights on how correlated the different color channels are, and on the inherent relationships between the various color channels. 

The eigenvalue decomposition of $\mathbf{M}_i$ yields 
\begin{gather}
\mathbf{M}_i\mathbf{e}^{st}=\lambda^{t}\mathbf{e}^{st},
\end{gather}
where $\mathbf{e}^{st}$ represents the $s^{th}$ eigenvector for $t^{th}$ eigenvalue, and $\lambda^{t}$ represents the $t^{th}$ eigenvalue. These eigenvalues and eigenvectors are used for computing the relative contributions of different color channels to the new orthogonal axes.                                                                                                                             

As an indication of which color channel may be most suitable for sky/cloud segmentation, we analyze the relative loading factors of the different color channels on the primary principal component. We calculate the loading factors by considering the relative contribution of the different color channels on the primary principal component. We consider the eigenvector $\mathbf{e}^{s1}$ corresponding to the primary principal component, and compute the absolute values of the $16$ components of this eigenvector as the respective loading factors of $16$ color channels.

\subsection{Discrimination Threshold}
\label{sec:bimodality}
In addition to principal component analysis, we also use the Receiver Operating Characteristics (ROC) curve~\cite{ROC_main,microarray-ROC} to identify the important color components in sky/cloud image segmentation. The ROC curve represents the fraction of true positives and false positives samples in the sky/cloud binary classification problem for varying values of discrimination threshold. The area between this ROC curve and the random classifier slope is widely used as a measure for the discriminatory performance in a classification framework. 

We consider the set of $16$ color components and compute the area under the ROC curve for each, which is subsequently used to identify the best color components for sky/cloud segmentation.

\section{Probabilistic Segmentation}
\label{sec:prob-segment}
Using statistical tools, we identify the most discriminative color channels for cloud segmentation. Out of $16$ color channels, the best performing color channels may be subsequently used as discriminatory feature $\mathbf{X}^{f}_{i}$ in the segmentation framework. 

Our objective is to predict the label of a pixel as either sky or cloud, given a set of training images with manually annotated ground truths. We formulate this task of sky/cloud image segmentation in a probabilistic manner using Partial Least Squares (PLS) regression. Both PCA and PLS are based on the same underlying motivation, namely to extract components that accounts for maximum variation in input data. 

We start with the assumption that the sky/cloud segmentation problem is dependent on a large number of explanatory independent variables (color channels in this case). Using dimensionality reduction techniques like PCA, we observe the degree of correlation amongst the variables and the amount of variance captured. PCA provides this information based only on the independent input color channels and project the data into new sets of orthogonal axes. In PLS however, both the input color channels and the output labels (sky/cloud) are considered. Using a set of manually defined ground truth images, we model the relationship between the input features and the output label. In the testing stage, we use this estimated model to predict the output labels of the pixels of a given test image.

Suppose, for a sample $\mathbf{X}_{i} \in {\rm I\!R}^{m \times n}$, the feature vector is denoted by $\mathbf{X}^{f}_{i} \in {\rm I\!R}^{mn \times k}$ and the corresponding binary ground-truth image as $\mathbf{Y}_{i} \in {\rm I\!R}^{mn \times 1}$.  The feature vector $\mathbf{X}^{f}_{i}$ consists of $k$ color channels from the set of $16$ color channels. We decompose the feature vector $\mathbf{X}^{f}_{i}$ and labels $\mathbf{Y}_{i}$ as 
\begin{align}
\mathbf{X}^{f}_{i} = \mathbf{T}_{i}\mathbf{P}_{i}^{T} + \mathbf{E}_{i},\\
\mathbf{Y}_{i} = \mathbf{U}_{i}\mathbf{Q}_{i}^{T} + \mathbf{F}_{i},
\label{eq:pls1}
\end{align}
where $\mathbf{T}_{i} \in {\rm I\!R}^{mn \times p}$ and $\mathbf{U}_{i} \in {\rm I\!R}^{mn \times p}$ are $p$ extracted latent matrices.  $\mathbf{P}_{i} \in {\rm I\!R}^{k \times p}$ and $\mathbf{Q}_{i} \in {\rm I\!R}^{1 \times p}$ are the loading matrices, and $\mathbf{E}_{i} \in {\rm I\!R}^{mn \times k}$ and $\mathbf{F}_{i} \in {\rm I\!R}^{mn \times 1}$ are the residual matrices. Henceforth, for the sake of brevity and without loss of generality we drop the suffix $i$ in the notations. 

In the generalized partial least squares regression sense, we decompose $\mathbf{X}^{f}$ and $\mathbf{Y}$ so as to maximize the covariance between $\mathbf{T}$ and $\mathbf{U}$ \cite{SIMPLS}.
Under the assumption that the input matrix $\mathbf{X}^{f}$ can be expressed as a linear combination of a few latent variables, we can assume the existence of a linear relationship between $\mathbf{T}$ and $\mathbf{U}$ such that \begin{align}
\label{eq:pls2}
\mathbf{U} = \mathbf{T}\mathbf{D}+\mathbf{H},
\end{align}
where $\mathbf{D} \in {\rm I\!R}^{p \times p}$ is a diagonal matrix and $\mathbf{H}$ is the residual matrix. Combining Eqs.\ (\ref{eq:pls1}) and (\ref{eq:pls2}), we get
\begin{gather}
\label{eq:pls3}
\mathbf{Y}=\mathbf{T}\mathbf{D}{\mathbf{Q}}^{T} + (\mathbf{H}\mathbf{Q}^{T} + \mathbf{F}),\\
\mathbf{Y}=\mathbf{T}\mathbf{C}^{T} + \mathbf{F}^{*},
\end{gather}
where $\mathbf{C}^{T}$ is the matrix of regression coefficients and $\mathbf{F}^{*}$ is the residual matrix. We extract score vectors $\mathbf{t}$ and $\mathbf{u}$ from predictors and response variables respectively, and express as: 
\begin{gather}
\mathbf{t}=\mathbf{Xw},\\
\mathbf{u}=\mathbf{Yc},
\end{gather}
where $\mathbf{w}$ and $\mathbf{c}$ are the corresponding weight vectors of $\mathbf{X}$ and $\mathbf{Y}$. These score vectors are necessary to define the response and predictor variables in the PLS setting. $\mathbf{W}$ represents a set of orthogonal basis vectors, such that $\mathbf{W}=(\mathbf{w}_1, \mathbf{w}_2, \ldots, \mathbf{w}_p)$. From \cite{Manne1987}, we can express the regression coefficient matrix as 
\begin{align}
\label{eq:pls4}
\mathbf{T} = \mathbf{X}\mathbf{W}(\mathbf{P}^{T}\mathbf{W})^{-1}.
\end{align}

Using Eqs.~(\ref{eq:pls3}) and (\ref{eq:pls4}), we get
\begin{gather}
\label{eq:pls5}
\mathbf{Y}=\mathbf{X}\mathbf{W}(\mathbf{P}^{T}\mathbf{W})^{-1}\mathbf{C}^{T} + \mathbf{F}^{*}, \\
\label{eq:pls6}
\mathbf{Y}=\mathbf{XB} + \mathbf{F}^{*},
\end{gather}
where $\mathbf{B}=\mathbf{W}(\mathbf{P}^{T}\mathbf{W})^{-1}\mathbf{C}^{T}$ represents the matrix of regression coefficients. 

In our proposed approach, we select a set of training images from the database and obtain the regression coefficient matrix $\mathbf{B}$ as per Eq.\ (\ref{eq:pls6}). In the testing stage, we obtain the corresponding segmentation result as
\begin{align}
\label{eq:testing_eqn}
\mathbf{Y}_{\mathrm{\mathrm{test}}}=\mathbf{X}_{\mathrm{\mathrm{test}}}\mathbf{B}
\end{align}
by using the obtained regression coefficient matrix $\mathbf{B}$. However, because of the residual matrices and the inherent imperfections in modeling, the values of $\mathbf{Y}_{\mathrm{\mathrm{test}}}$ are not discrete in nature. We do not employ any thresholding at this stage, but rather normalize these values to the range $[0,1]$ ($0=\mbox{sky}, \mbox{ }1=\mbox{cloud}$). 
These normalized values provide a probabilistic  indication of the ``belongingness'' of a particular pixel to a specific class (i.e.\ cloud or sky) instead of a binary decision. 

\section{Sky/Cloud Image Databases}
\label{sec:database}
One of the most important pre-requisites for rigorous evaluation of any segmentation algorithm is an annotated dataset of suitable images. Most research groups working on sky/cloud segmentation use their own proprietary images that are not publicly available. In this section, we mention the currently available database containing ground truth images and describe the need for a more comprehensive and consistent database.

\subsection{HYTA Database}
\begin{figure}[htb]
\centering
\includegraphics[height=0.64in]{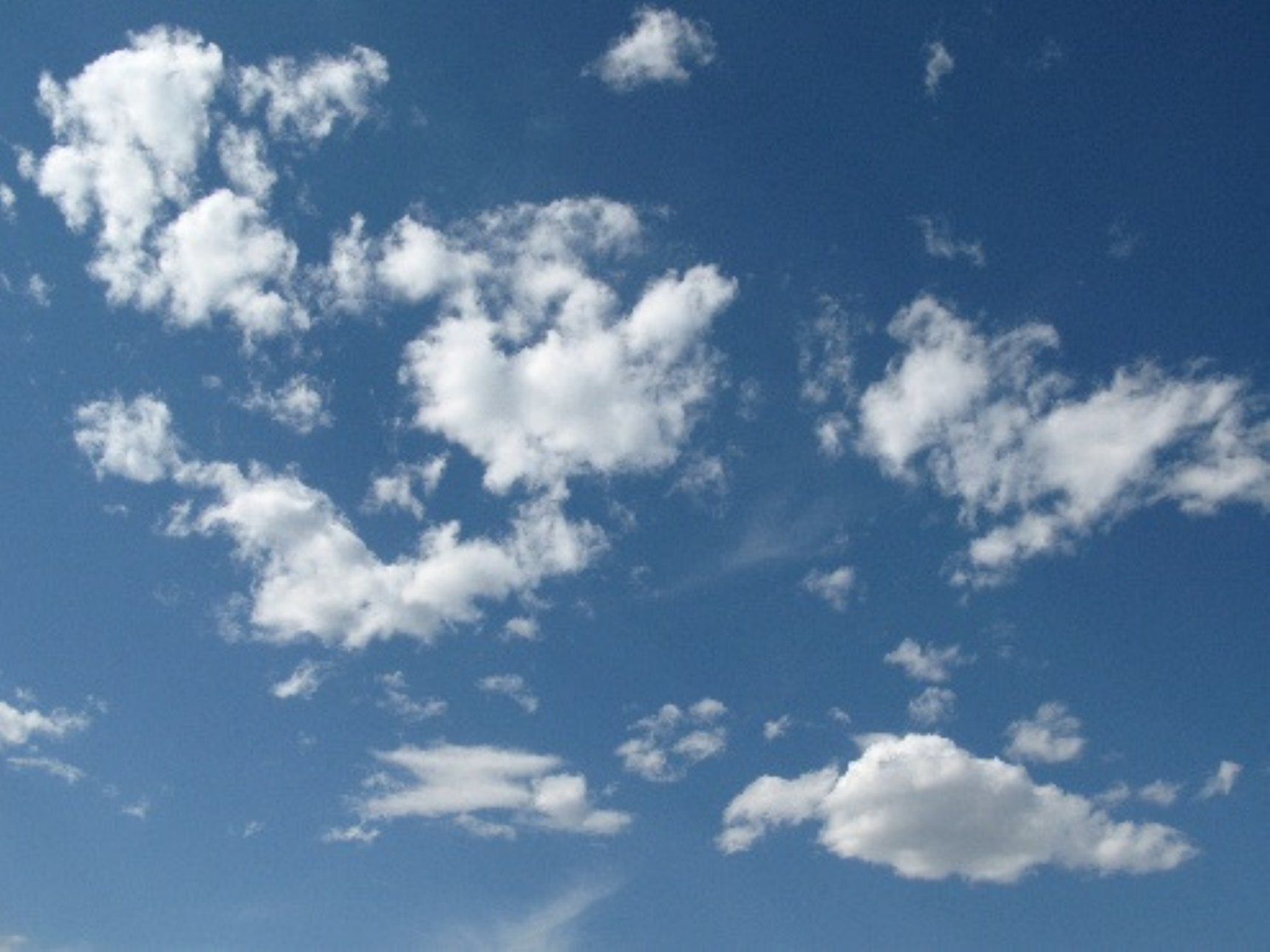}\hspace{0.5mm}
\includegraphics[height=0.64in]{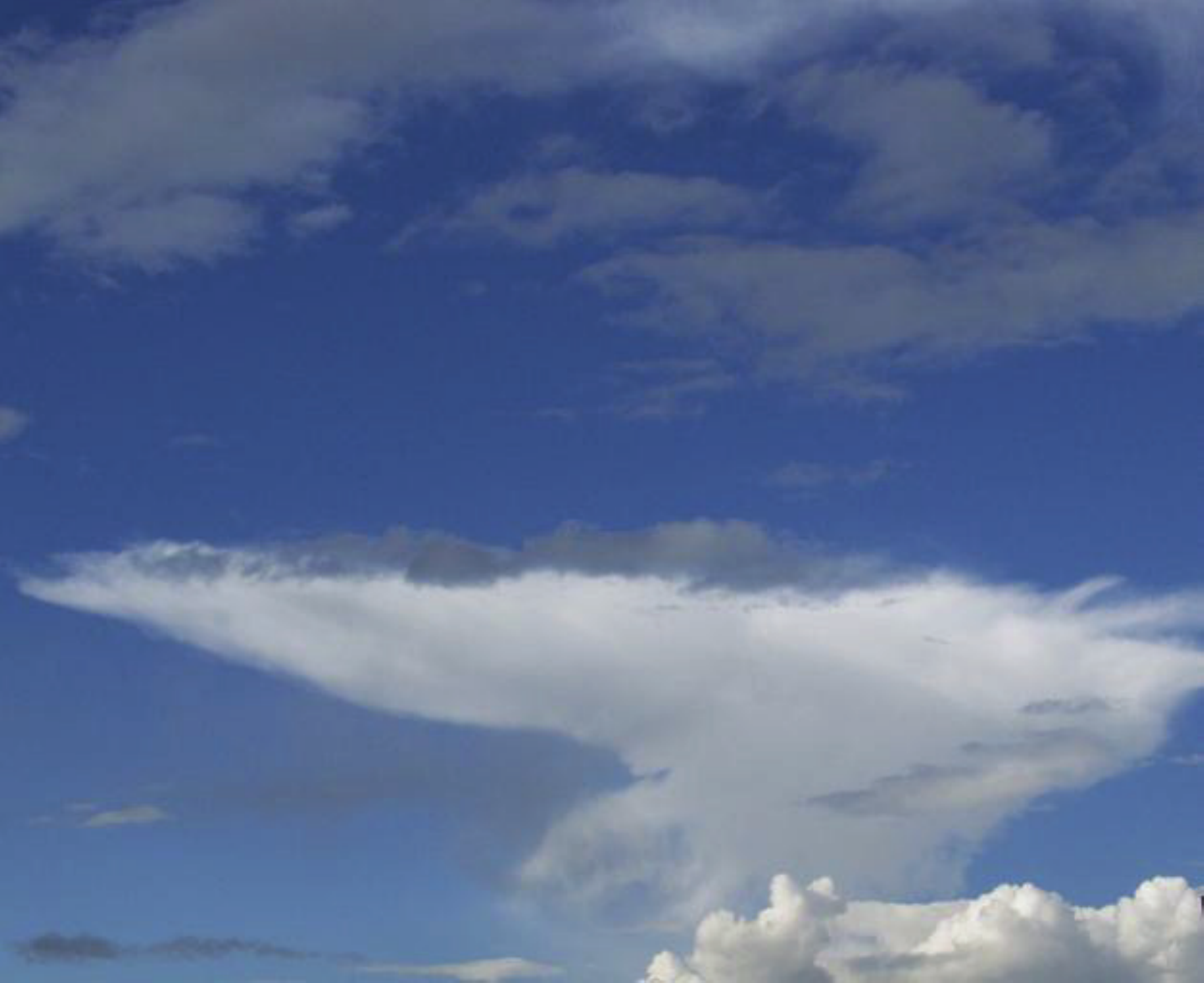}\hspace{0.5mm}
\includegraphics[height=0.64in]{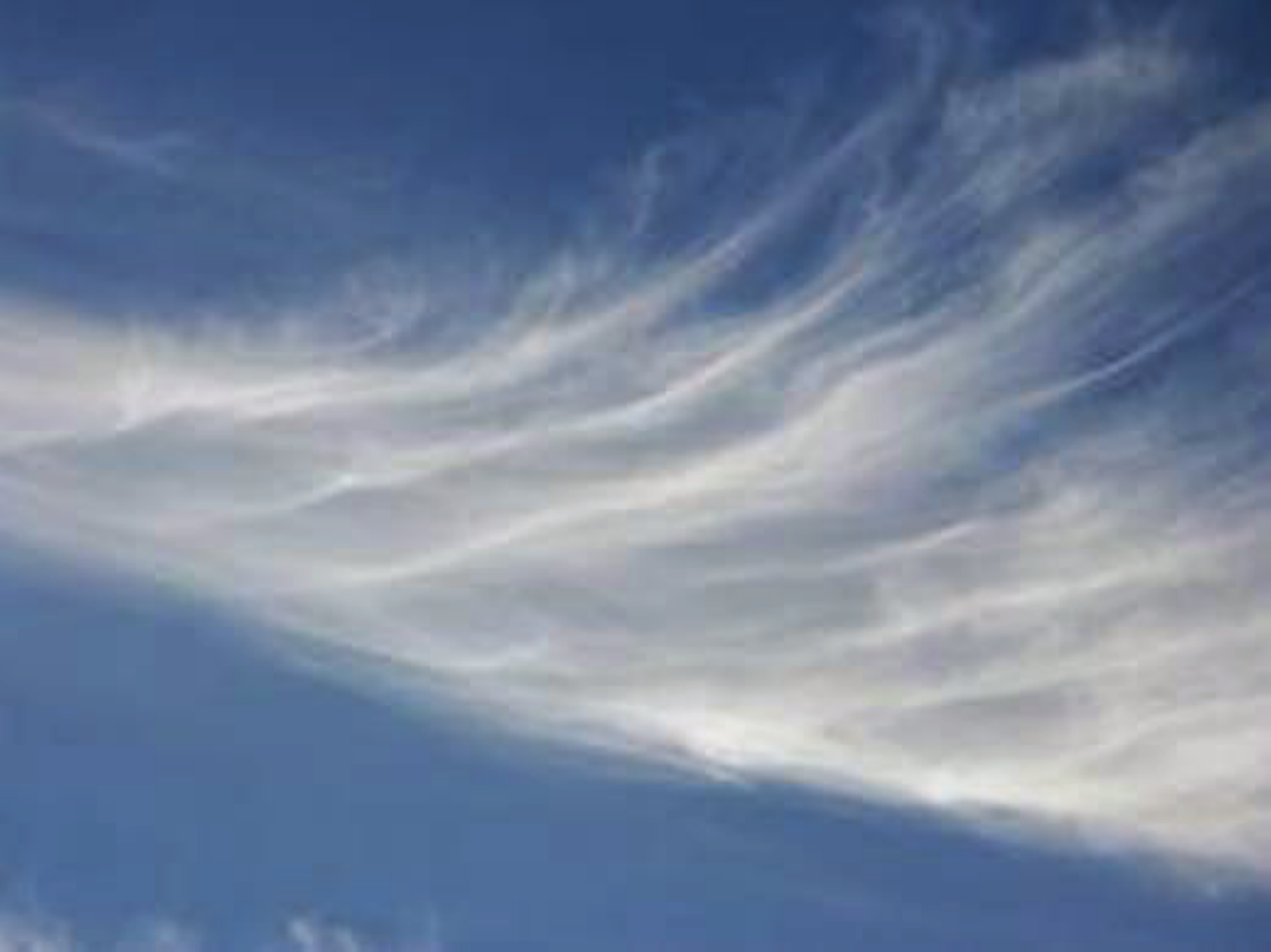}\hspace{0.5mm}
\includegraphics[height=0.64in]{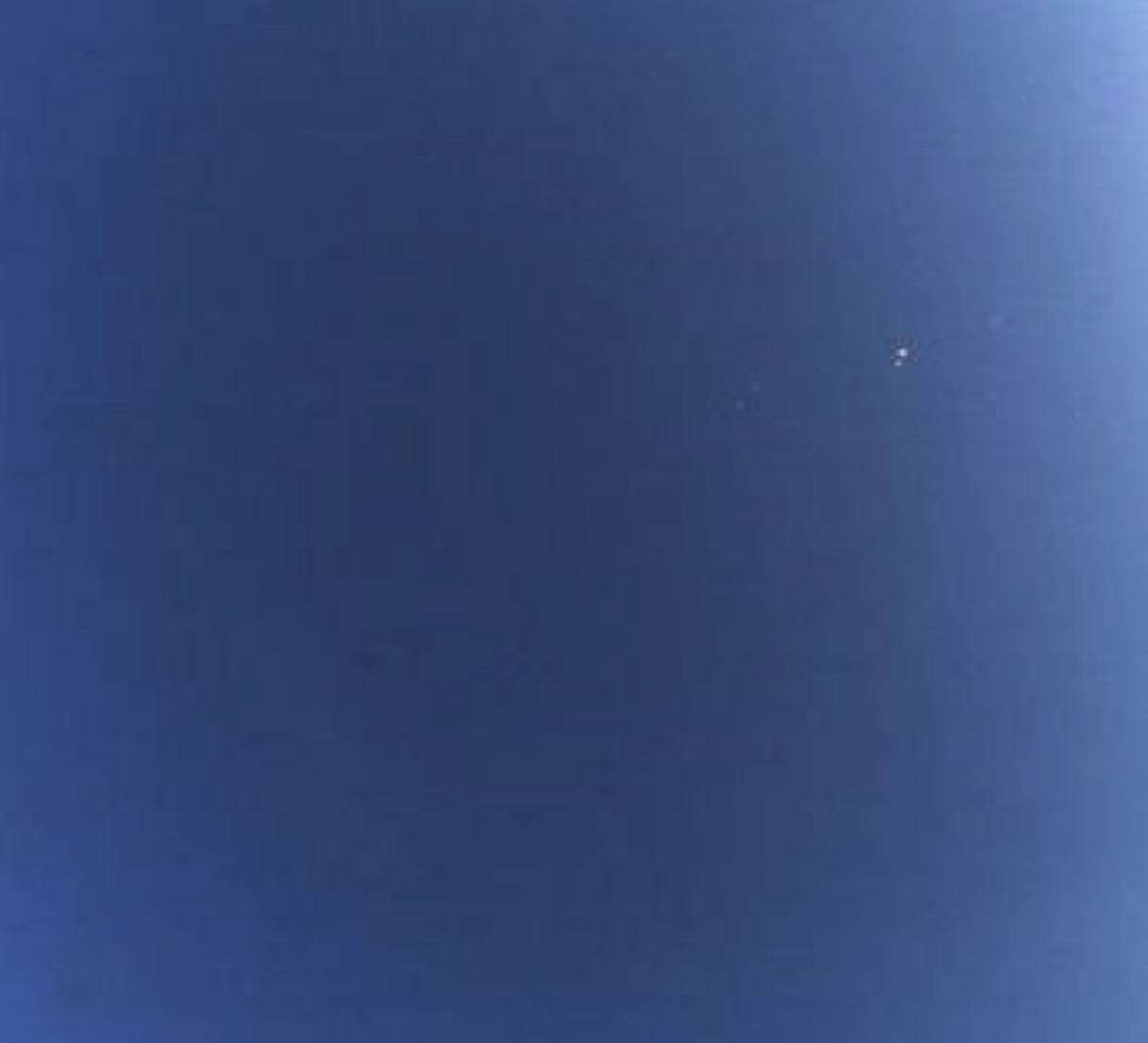}\\\vspace{1mm}
\includegraphics[height=0.64in]{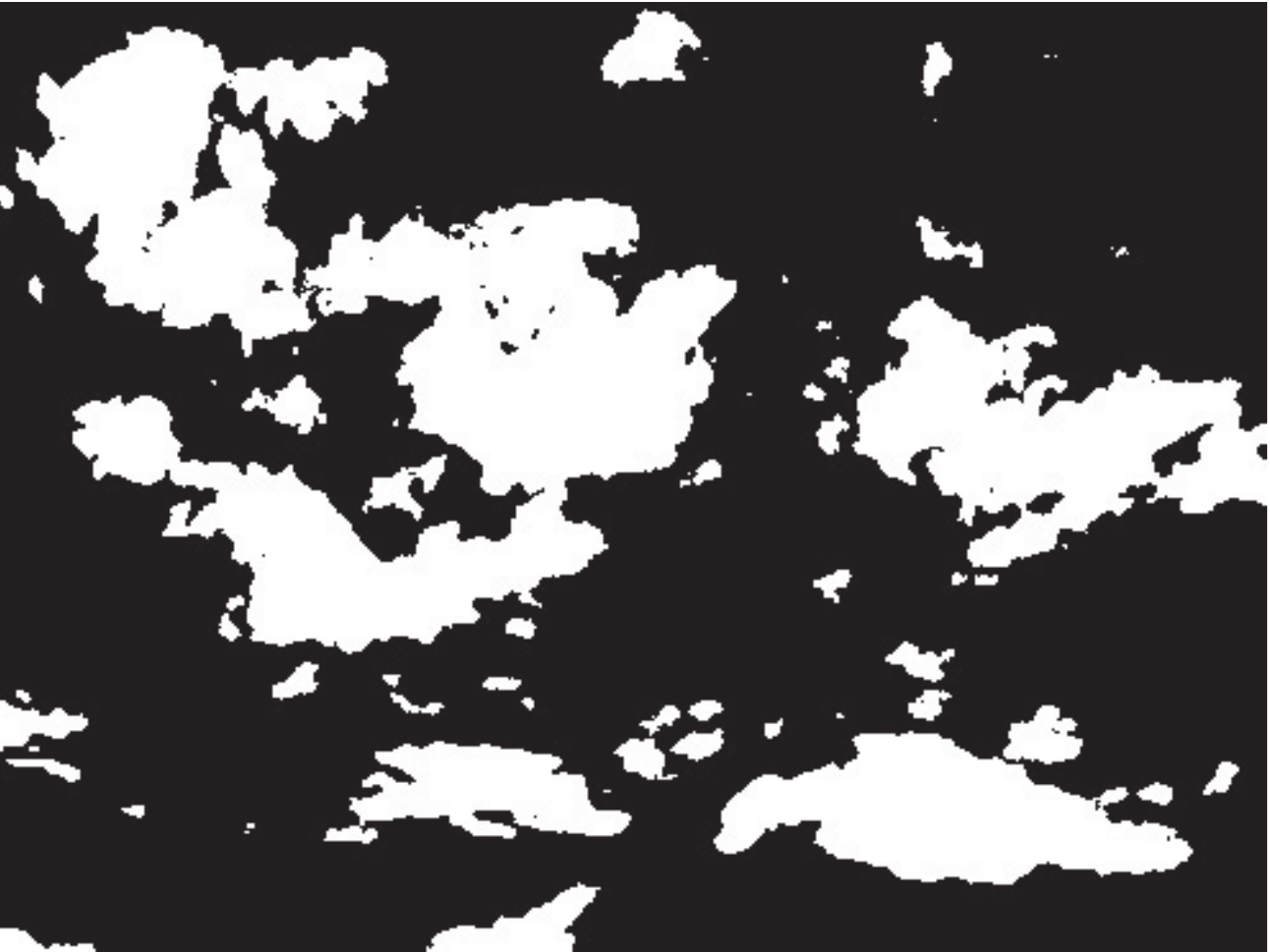}\hspace{0.5mm}
\includegraphics[height=0.64in]{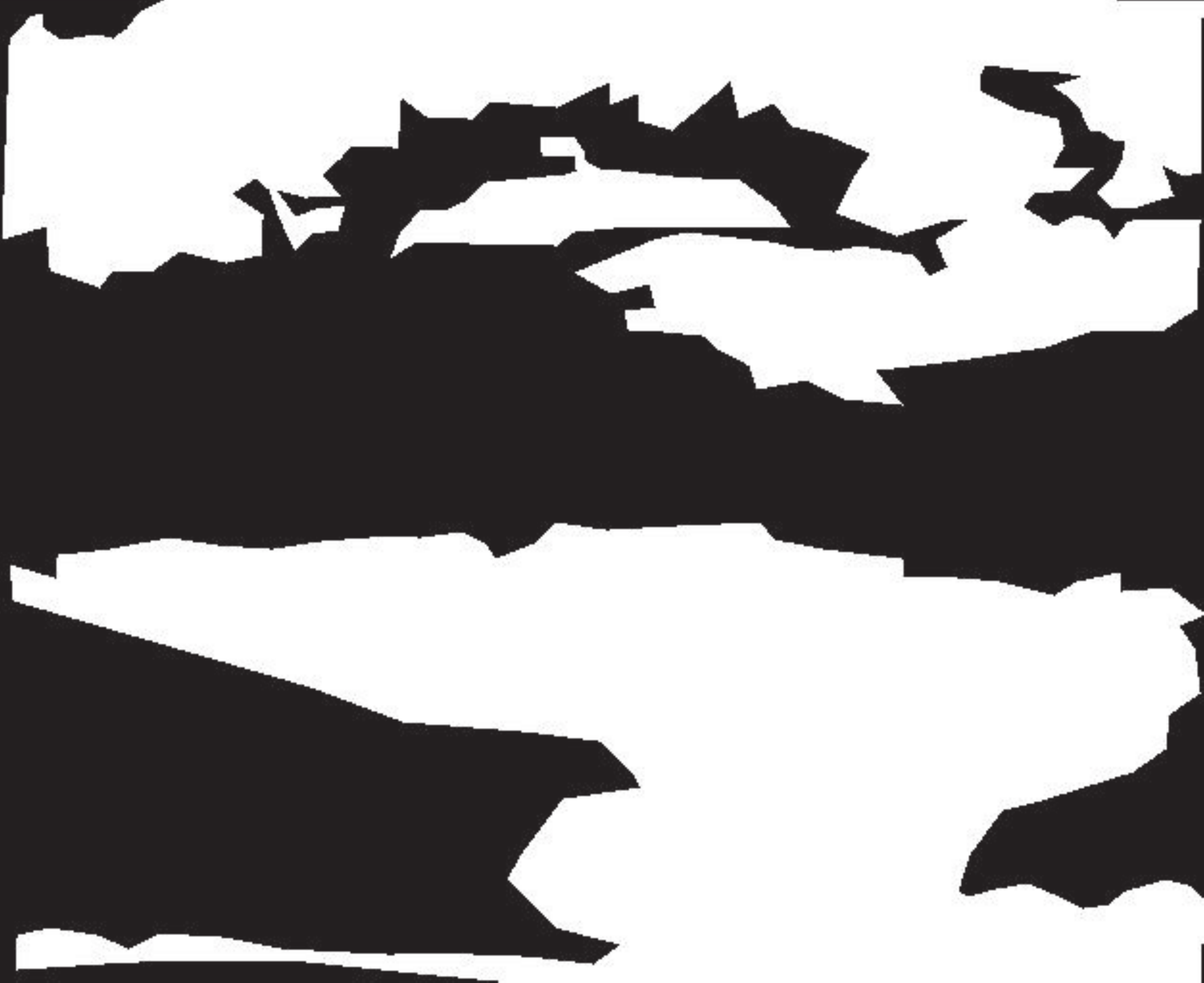}\hspace{0.5mm}
\includegraphics[height=0.64in]{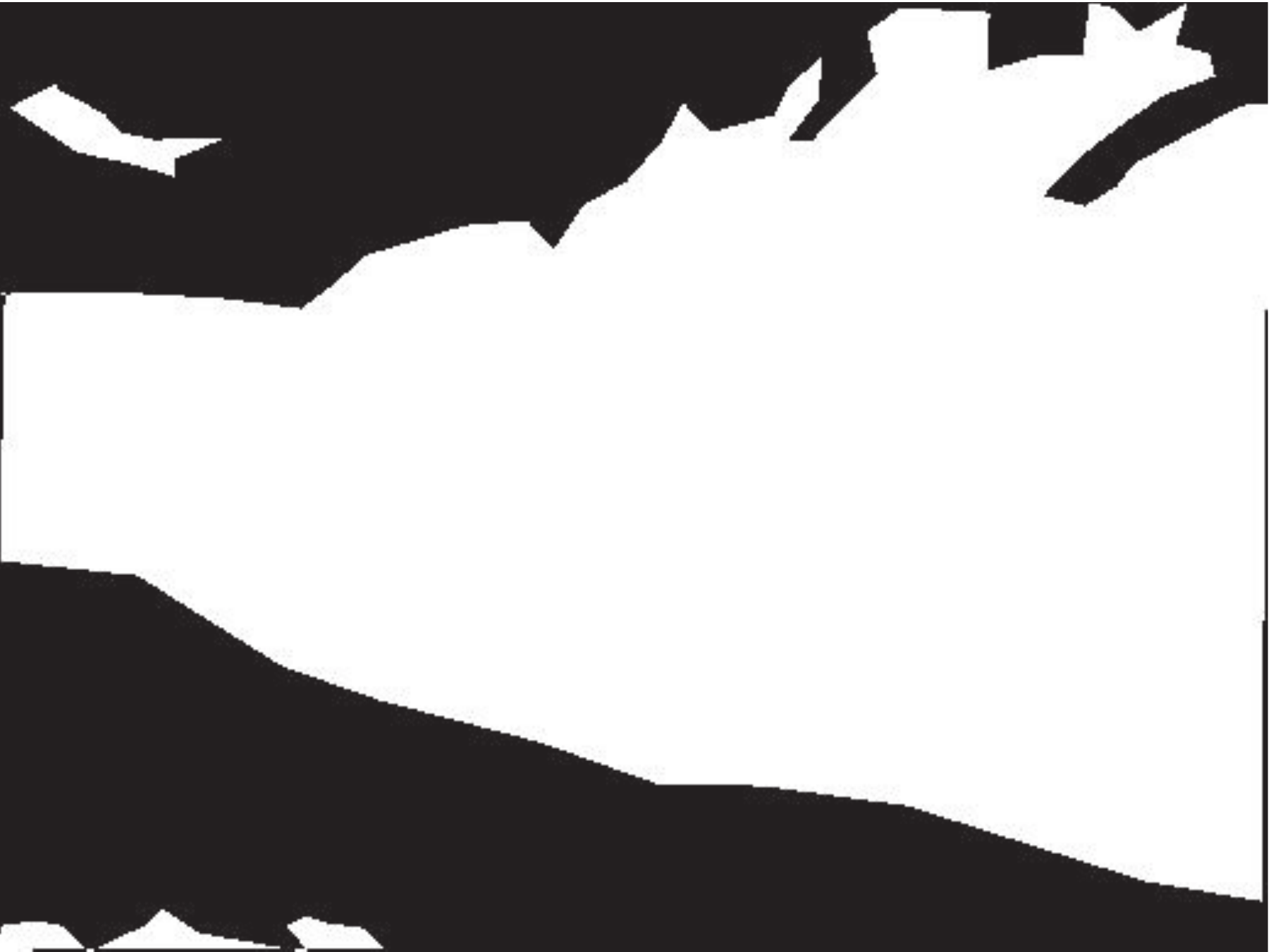}\hspace{0.5mm}
\includegraphics[height=0.64in]{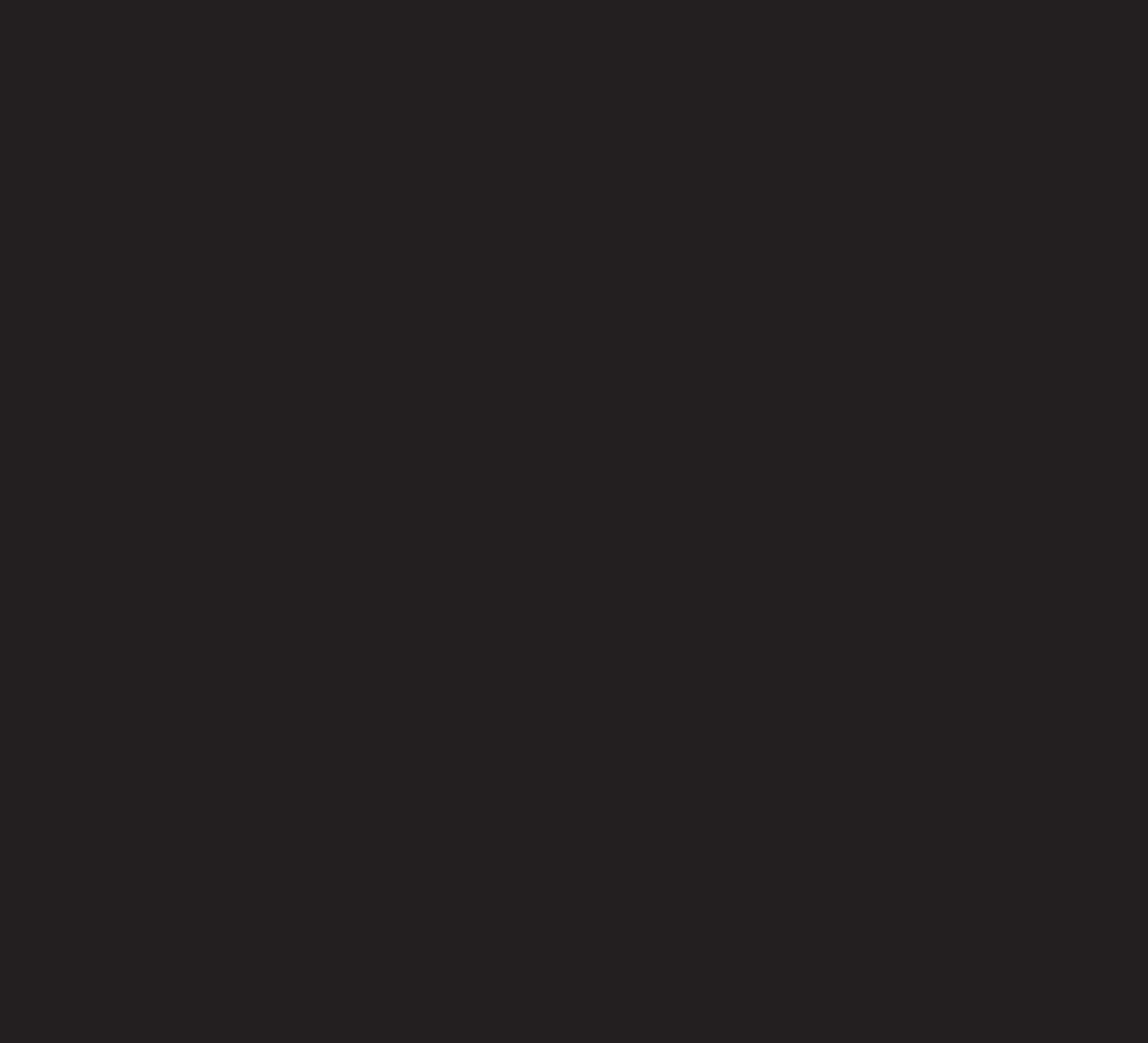}
\caption{Sample images from the HYTA database (top row) along with corresponding sky/cloud segmentation ground truth (bottom row).}
\label{fig:sample_HYTA}
\end{figure}

Till date, the only publicly available sky/cloud image dataset annotated with segmentation masks is the HYTA database by Li et al.\ \cite{Li2011}. It consists of $32$ distinct images with different sky/cloud conditions along with their binary ground-truth segmentation masks. These images are collected from two main sources, namely ground-based sky cameras located at Beijing and Conghua, China. Additionally, several images found on the Internet from other cameras and locations are included. The ground truth segmentation masks of each of the images are generated using a Voronoi polygonal region generator. The images in HYTA are of various dimensions; the average is $682 \times 512$ pixels. A few sample images of HYTA along with their ground truth are shown in Fig.\ \ref{fig:sample_HYTA}.

\subsection{SWIMSEG Database}
Because of the dearth of public databases for sky/cloud images, we created SWIMSEG, the \textbf{S}ingapore \textbf{W}hole Sky \textbf{IM}aging \textbf{SEG}mentation Database.\footnote{~The SWIMSEG database can be downloaded from \url{http://vintage.winklerbros.net/swimseg.html}.} It consists of $1013$ images that were captured by a custom-designed ground-based sky imager called WAHRSIS (Wide Angle High Resolution Sky Imaging System), developed and deployed at Nanyang Technological University in Singapore~\cite{WAHRSIS}. 
The imaging system of our ground-based sky camera consists of a Canon EOS Rebel T3i (a.k.a. EOS 600D) camera body and a Sigma 4.5mm F2.8 EX DC HSM Circular Fish-eye Lens with a field of view of 180 degrees.

\begin{figure}[htb]
\centering
   \subfloat[Original image]{\includegraphics[width=0.16\textwidth]{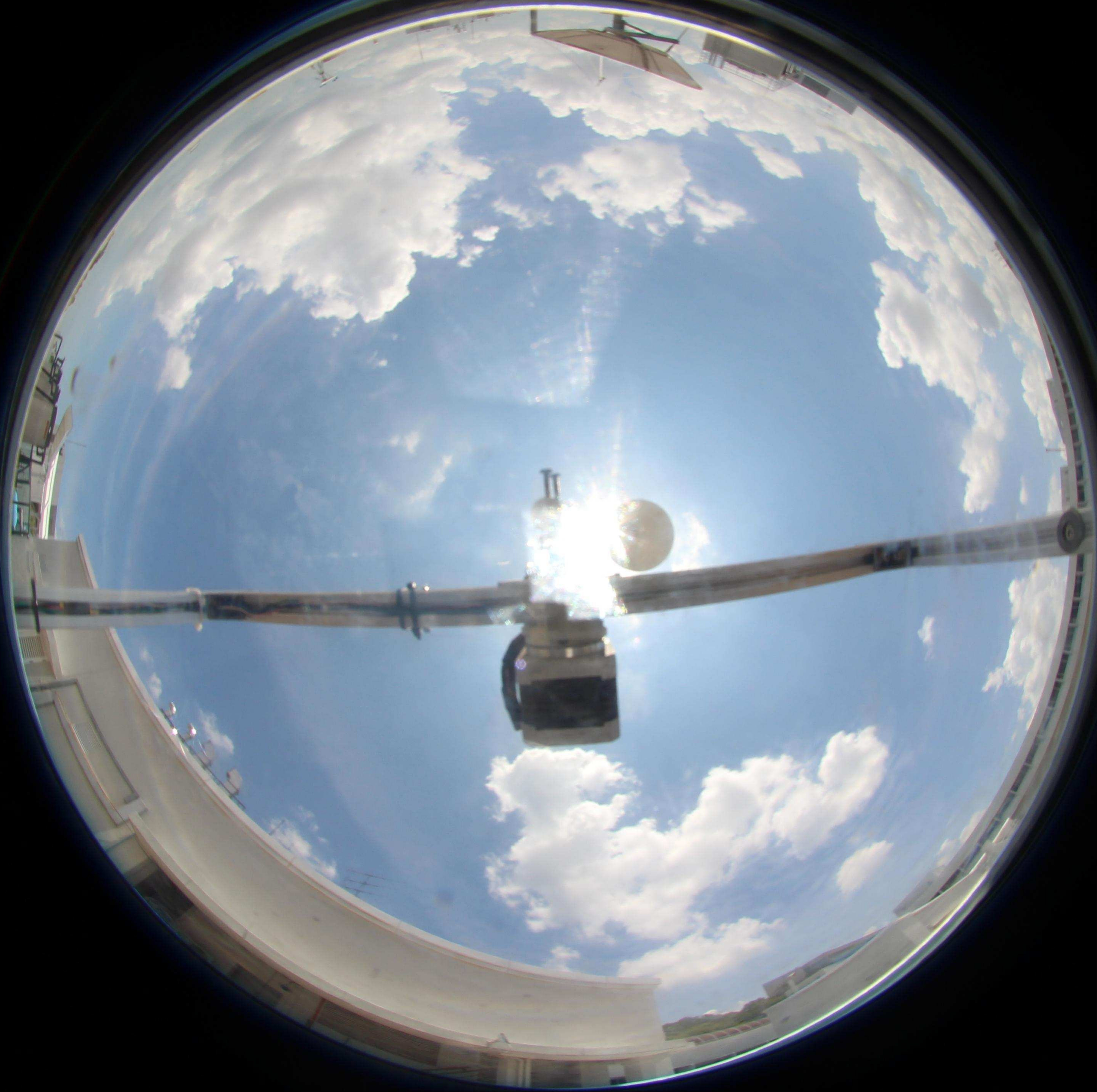}}
   \subfloat[Vignetting correction]{\includegraphics[width=0.16\textwidth]{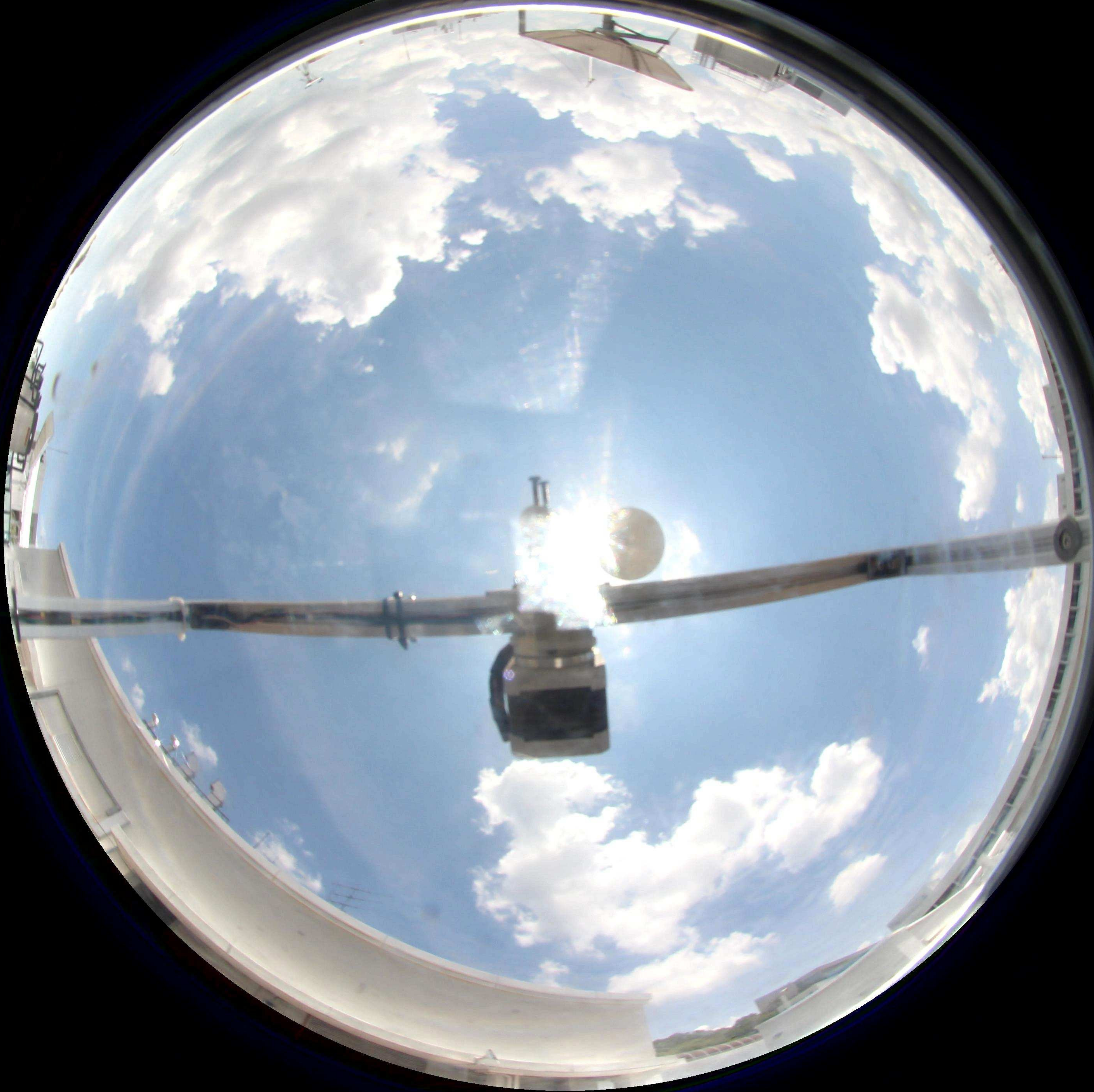}}
   \subfloat[Color correction]{\includegraphics[width=0.16\textwidth]{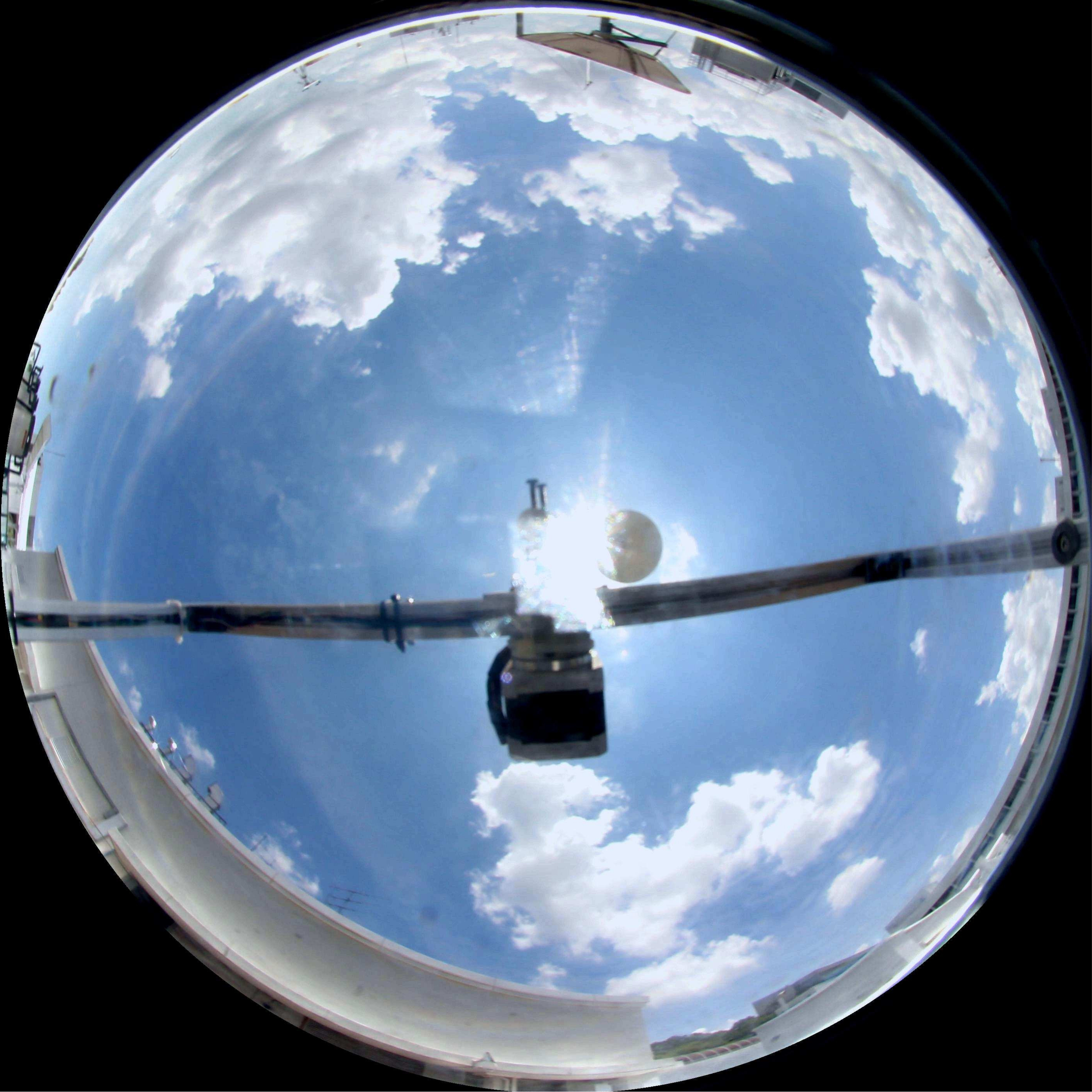}}
\caption{Image correction w.r.t.\ illumination and color.}\label{fig:pre-process}
\end{figure}

The images captured by this system suffer from distortions in illumination and color, which need to be corrected.  An example is shown in Fig.~\ref{fig:pre-process}. First, because of a phenomenon called vignetting, the images are comparatively darker at the edges as compared to the center. Vignetting occurs because the incident light rays reach the camera sensor with varying angles and is particularly noticeable for fish-eye lenses. We analyze vignetting in our imaging system using an integrating sphere \cite{WAHRSIS}  and perform the necessary correction in the image, the result of which is shown in Fig.~\ref{fig:pre-process}(b). 

Second, the colors in the captured images vary due to the different weather conditions and capturing hours of the images  and often do not correspond to the actual colors in the scene. We use a color checkerboard with $18$ different color patches as a reference, whose exact $L^{*}a^{*}b^{*}$ values are specified by the manufacturer. Using the mapping between captured and actual color, we thereby obtain the color-corrected image, which is shown in Fig.~\ref{fig:pre-process}(c).
 
\begin{figure}[htb]
\centering
   \subfloat[]{\includegraphics[width=0.20\textwidth]{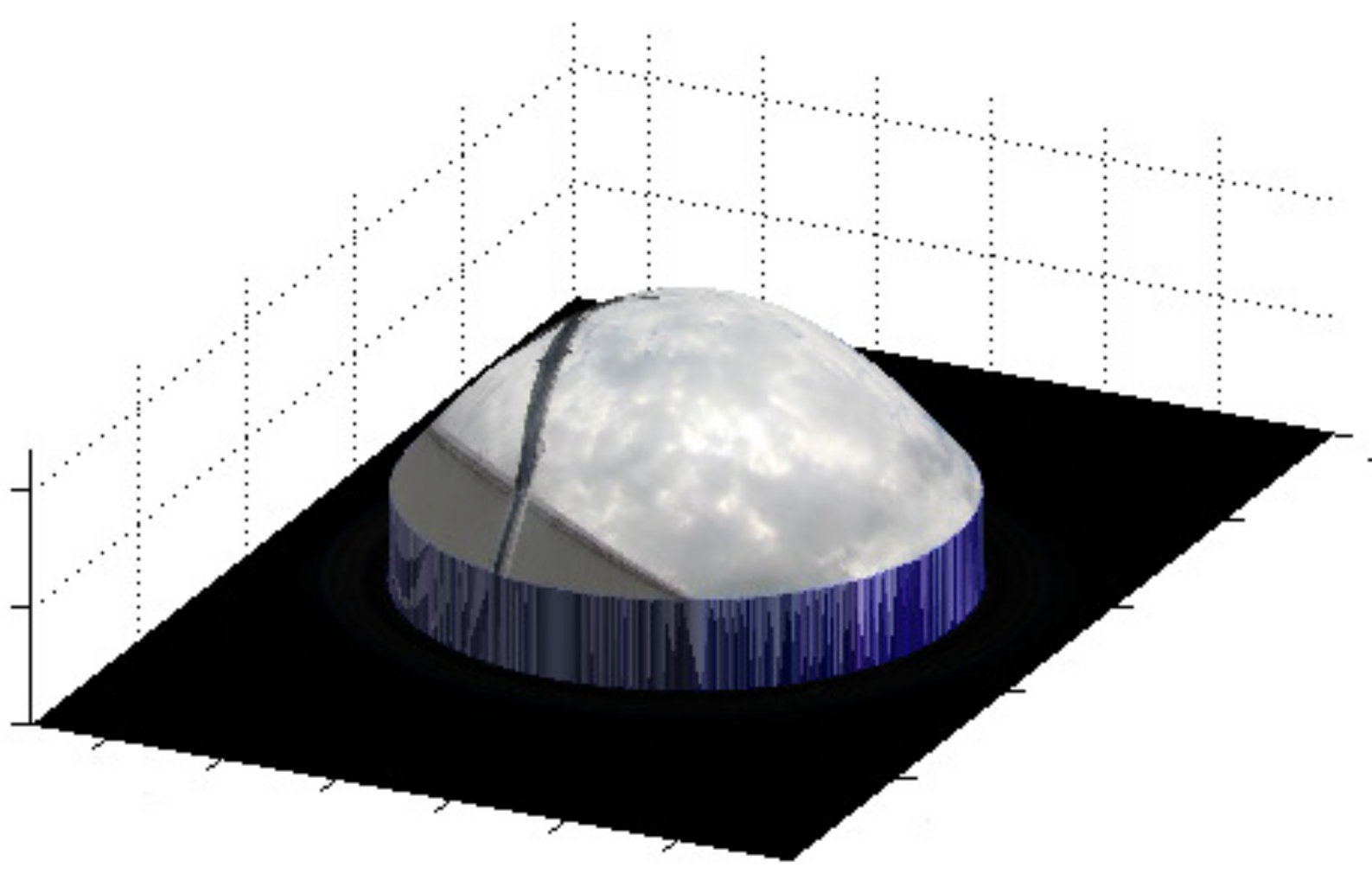}}
   \subfloat[]{\includegraphics[width=0.14\textwidth]{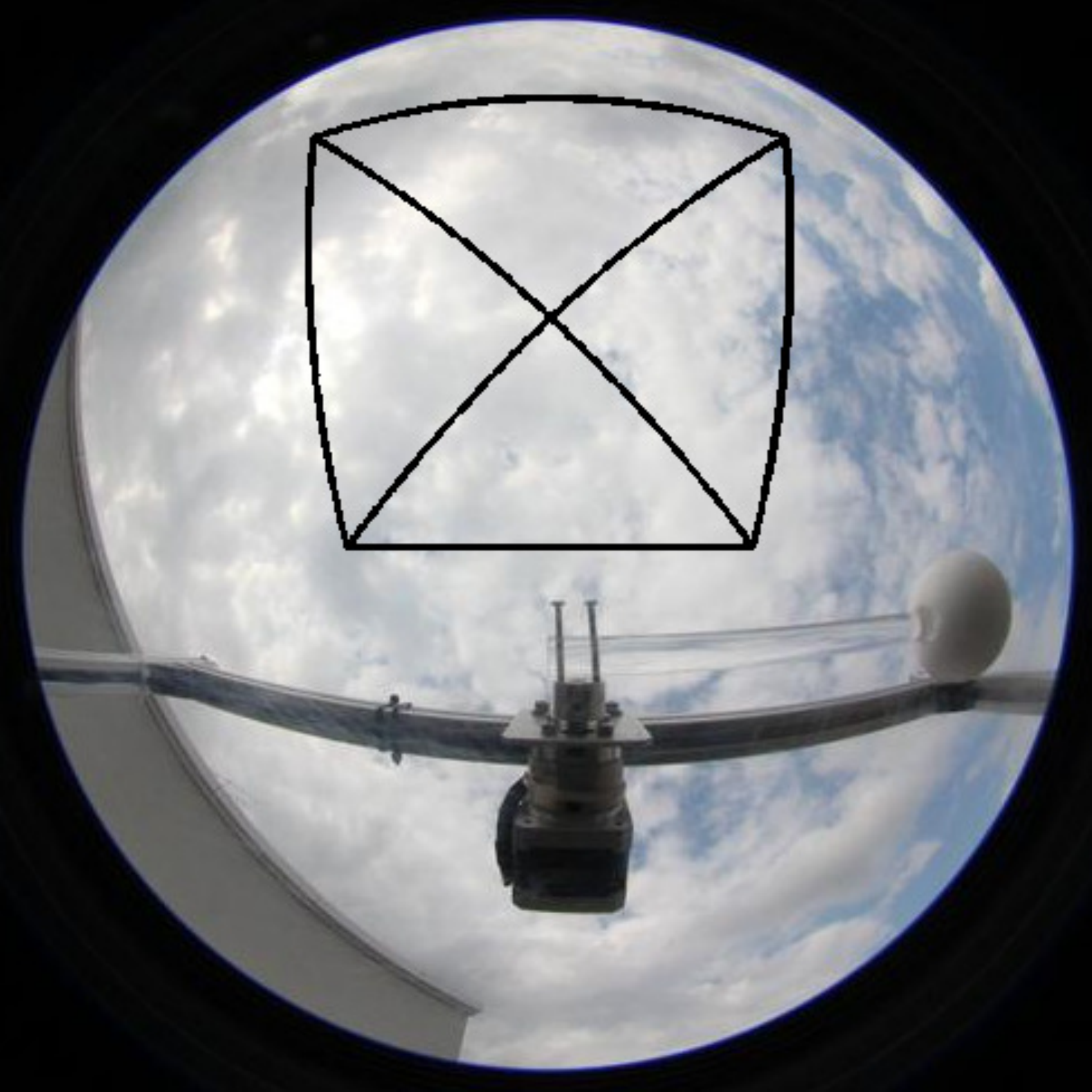}}
   \subfloat[]{\includegraphics[width=0.14\textwidth]{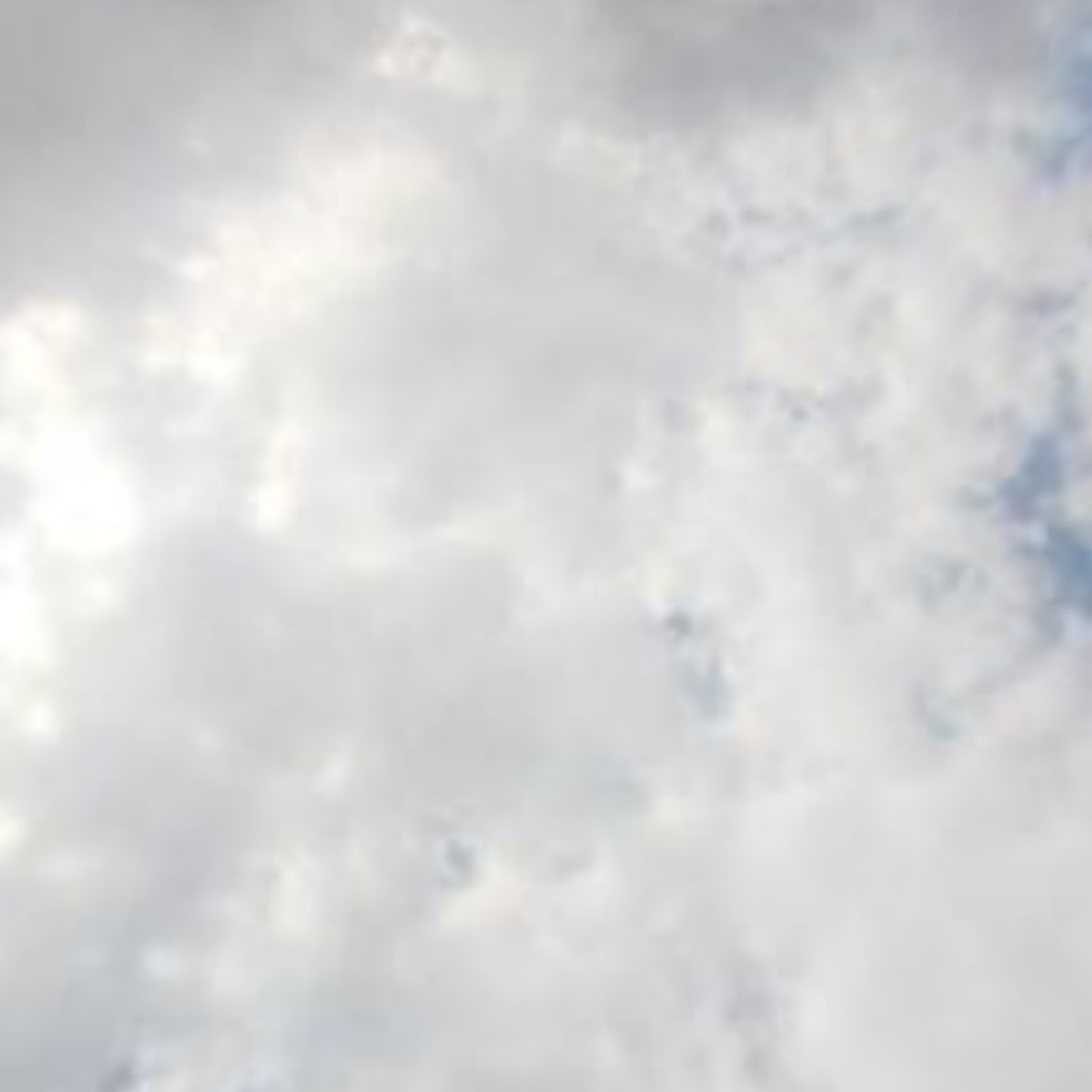}}
\caption{Generation of undistorted images using ray-tracing. (a) Projection of the image on a hemisphere; (b) Input image with borders and diagonals of the output image; (c) Undistorted output image. }\label{fig:undistortion-results}
\end{figure}

Finally, the images from the database are undistorted using the geometric calibration function, which models the effect of the lens. This function relates each pixel of the captured image to the azimuth and elevation angle of the corresponding incident light ray. Using that information, we can project the image onto a hemisphere whose center is the focal point. This is shown in Fig.~\ref{fig:undistortion-results}(a). We generate undistorted pictures using a ray tracing approach by considering an imaginary camera (with a standard lens) at center of the hemisphere, which points towards a user defined direction with a given azimuth and an elevation angle. In order to simulate this camera, we consider an output image with dimension $600 \times 600$. Each output image corresponds to a viewing angle of $62^{\circ}$. The rays passing through each pixel will intersect the hemisphere and then converge towards the center. The value of a pixel is then equal to the color of the hemisphere at its intersection point. We perform this process at varying azimuth angles for a particular angle of elevation to generate several undistorted images. Figure~\ref{fig:undistortion-results} shows the output image as well as the lines corresponding to its borders and diagonals on the original image.

The corresponding sky/cloud segmentation masks of the images were created in consultation with cloud experts from the Singapore Meteorological Services. A few representative sample images from this database are shown in Fig.\ \ref{fig:WAHRSISdb600}. 

\begin{table}[htb]
\begin{tabular}{ccccc}
\includegraphics[width=0.1\textwidth]{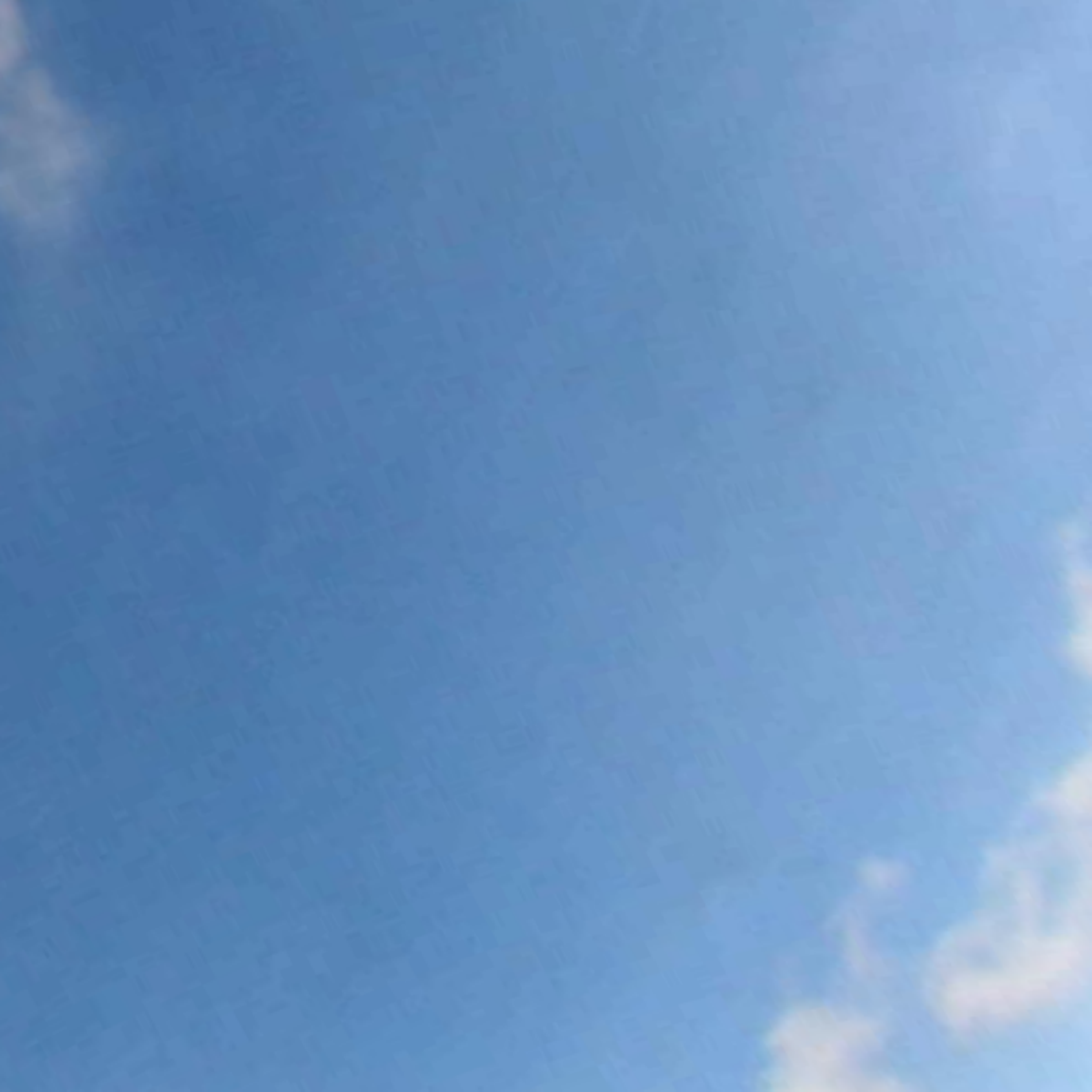} & 
\includegraphics[width=0.1\textwidth]{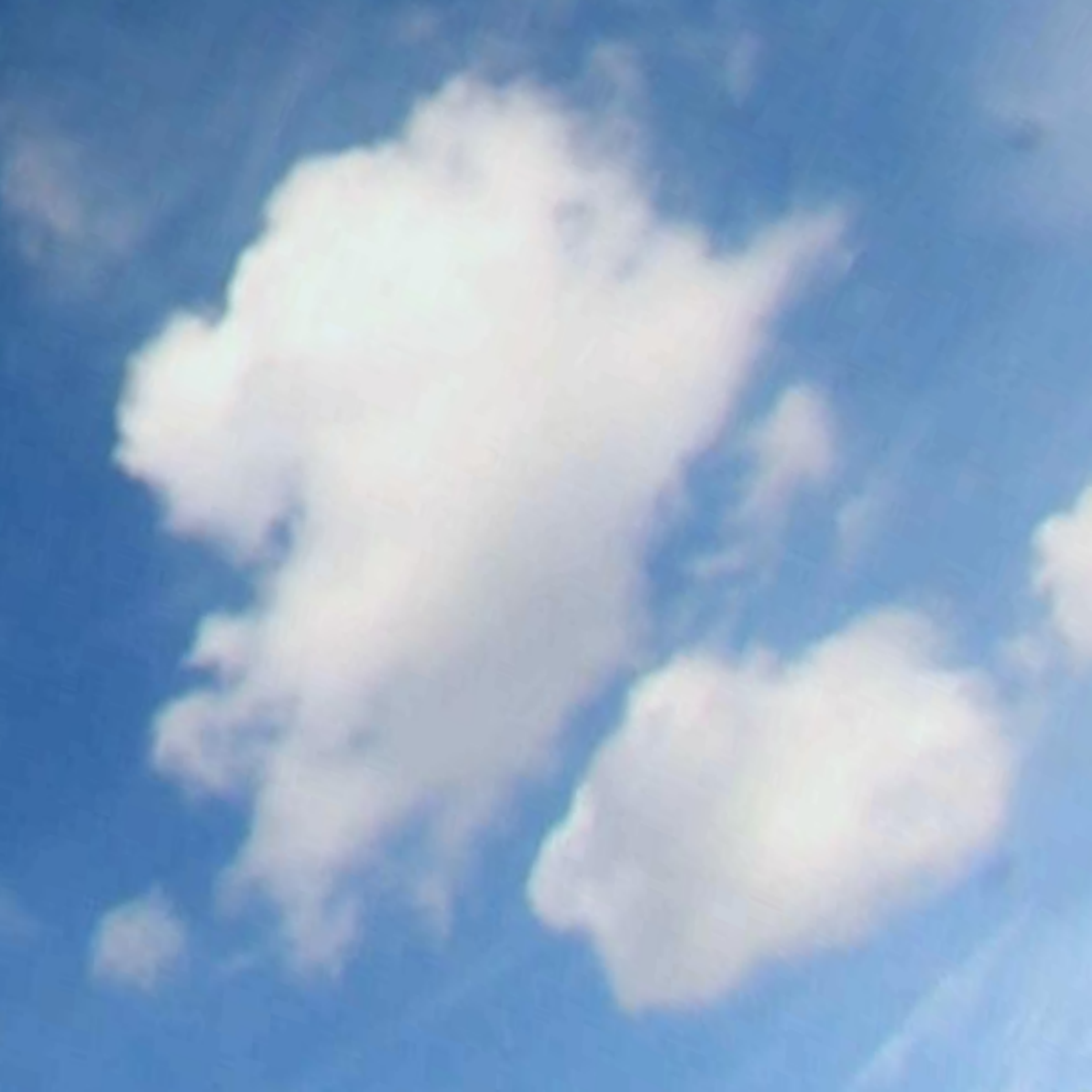} &  
\includegraphics[width=0.1\textwidth]{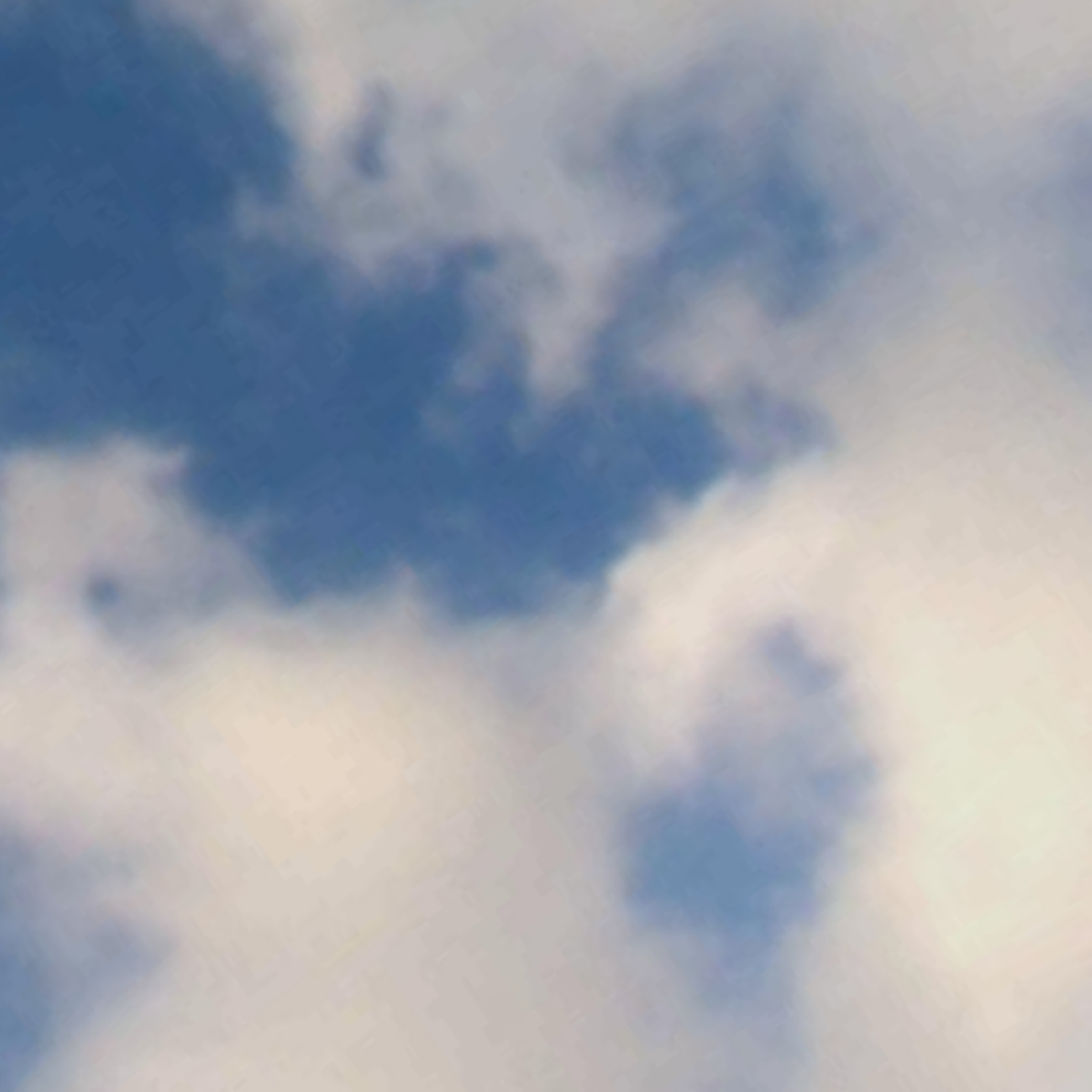} &
\includegraphics[width=0.1\textwidth]{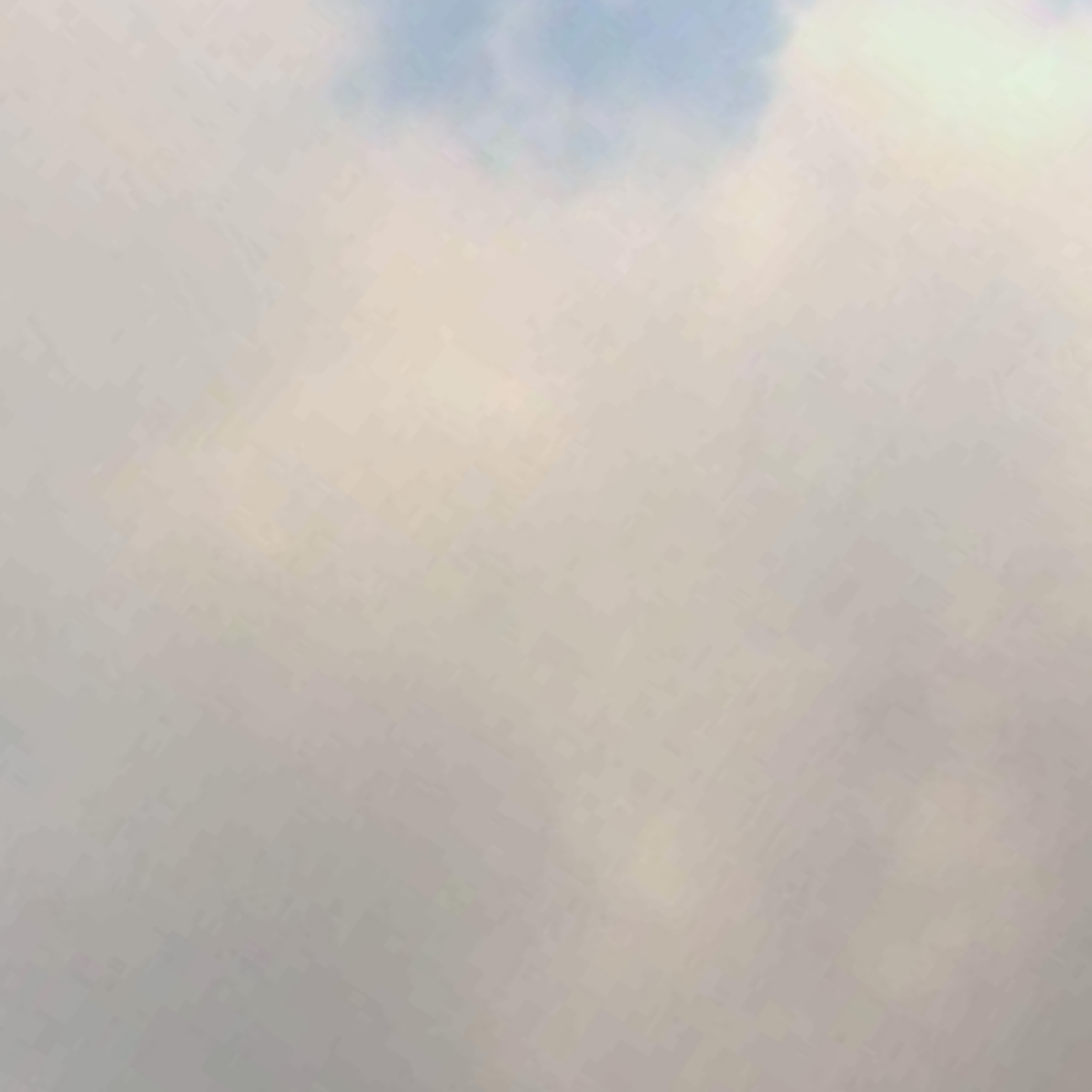}\\

\fbox{\includegraphics[width=0.1\textwidth]{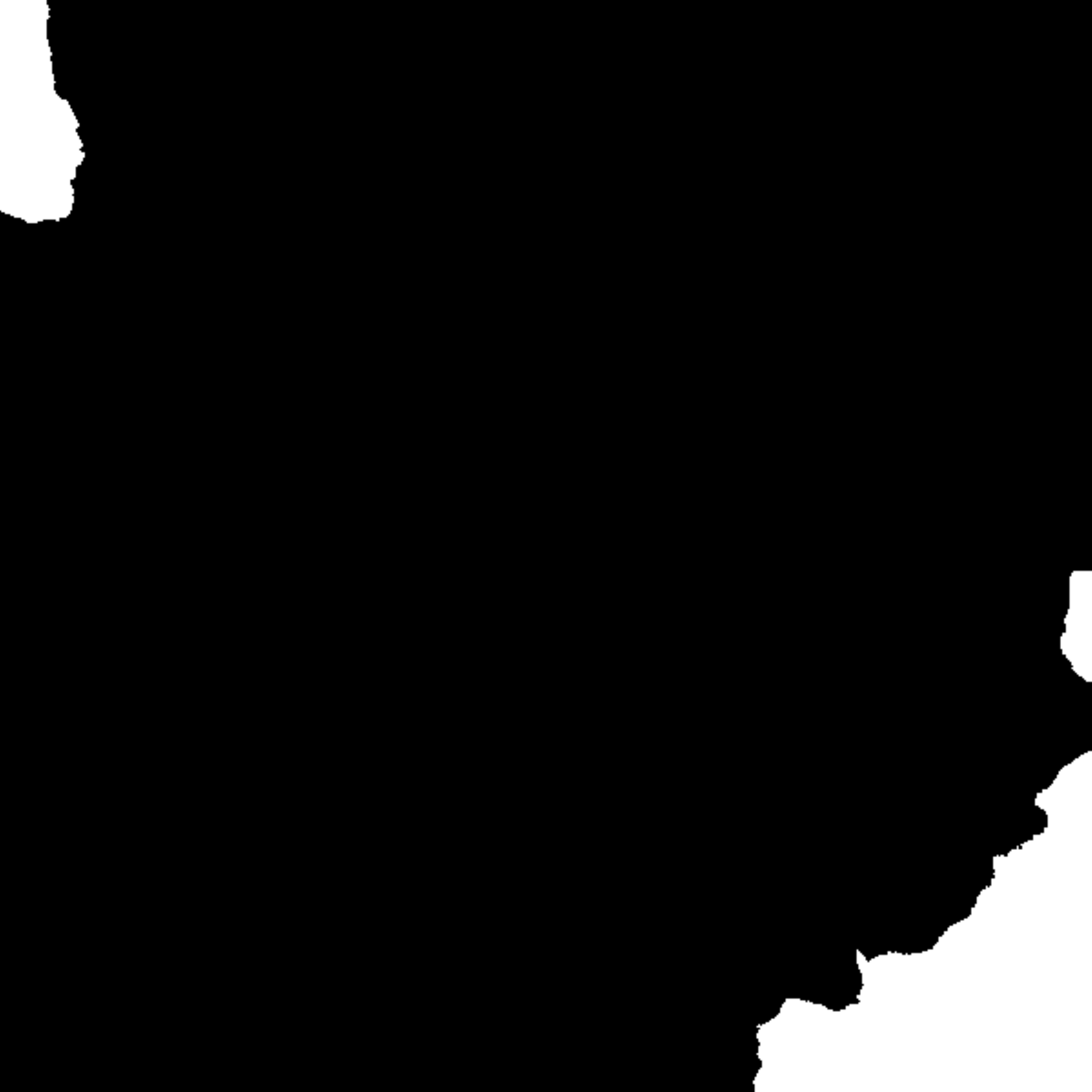}} &
\fbox{\includegraphics[width=0.1\textwidth]{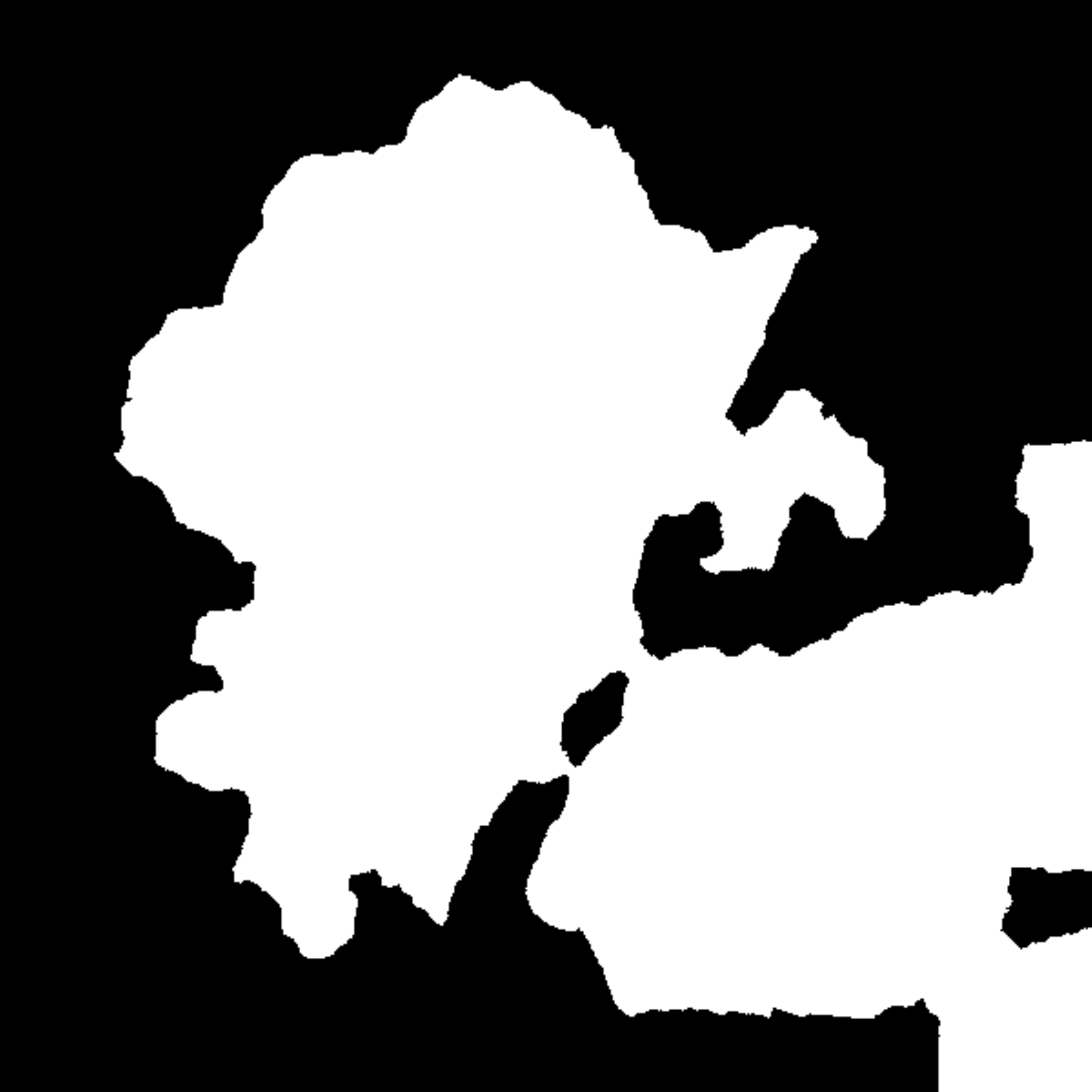}} &
\fbox{\includegraphics[width=0.1\textwidth]{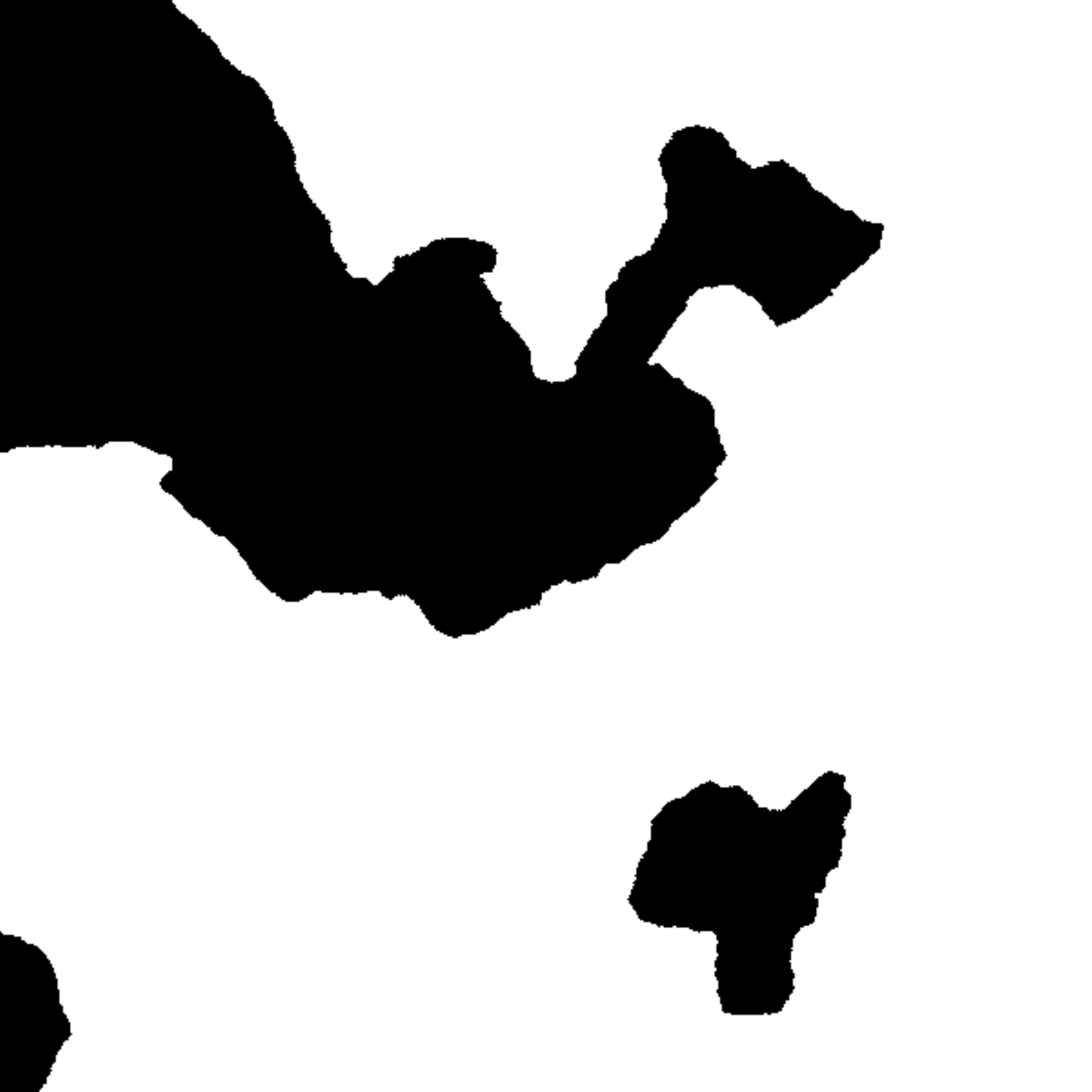}} & 
\fbox{\includegraphics[width=0.1\textwidth]{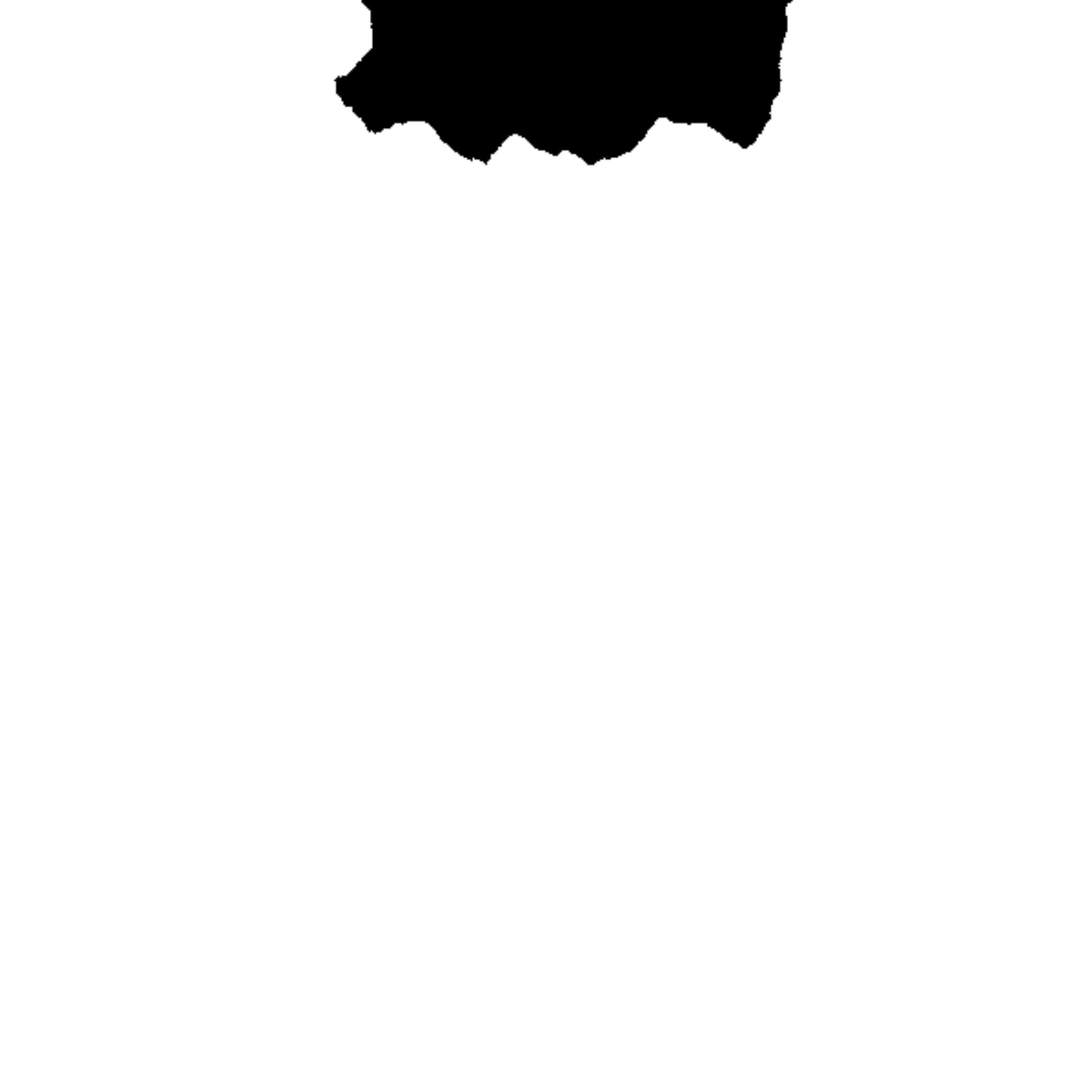}} \\

{\fontsize{0.3cm}{2em}\selectfont $6.2\%$} & 
{\fontsize{0.3cm}{2em}\selectfont $48.7\%$} & 
{\fontsize{0.3cm}{2em}\selectfont $72.5\%$} & 
{\fontsize{0.3cm}{2em}\selectfont $95.0\%$} \\

\end{tabular}
\captionof{figure}[foo]{Sample images from the SWIMSEG database (top row), along with corresponding sky/cloud segmentation ground truth (bottom row) and percentage of cloud coverage.}
\label{fig:WAHRSISdb600}
\end{table}

\begin{figure*}[htb]
\centering
\subfloat[Time of day]{\includegraphics[height=0.23\textwidth]{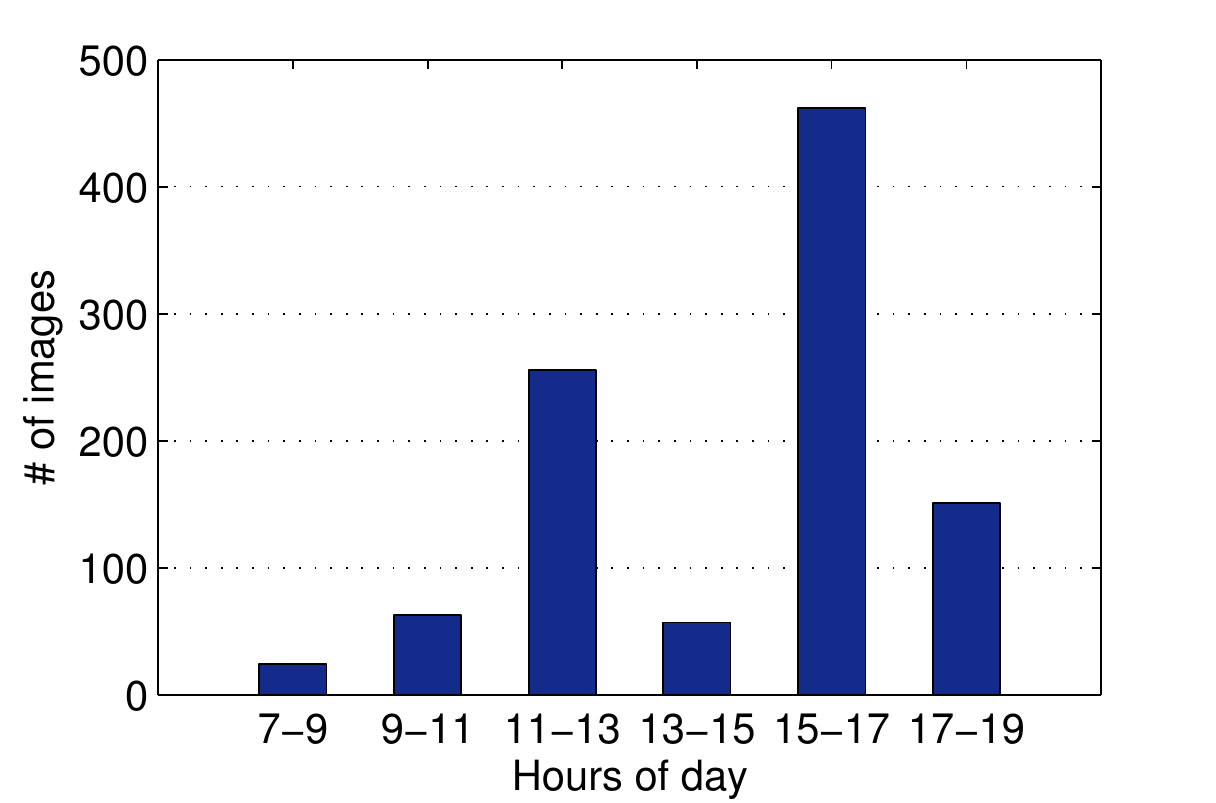}}\hspace{-0.5cm}
\subfloat[Cloud coverage]{\includegraphics[height=0.23\textwidth]{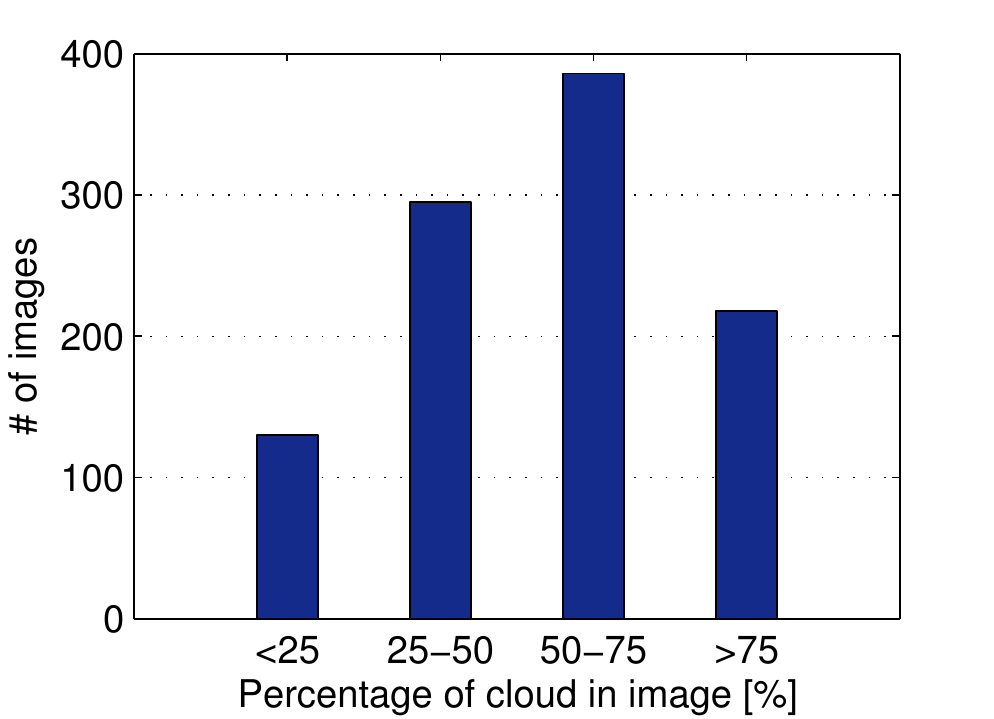}}\hspace{-0.5cm}
\subfloat[Distance from the sun]{\includegraphics[height=0.23\textwidth]{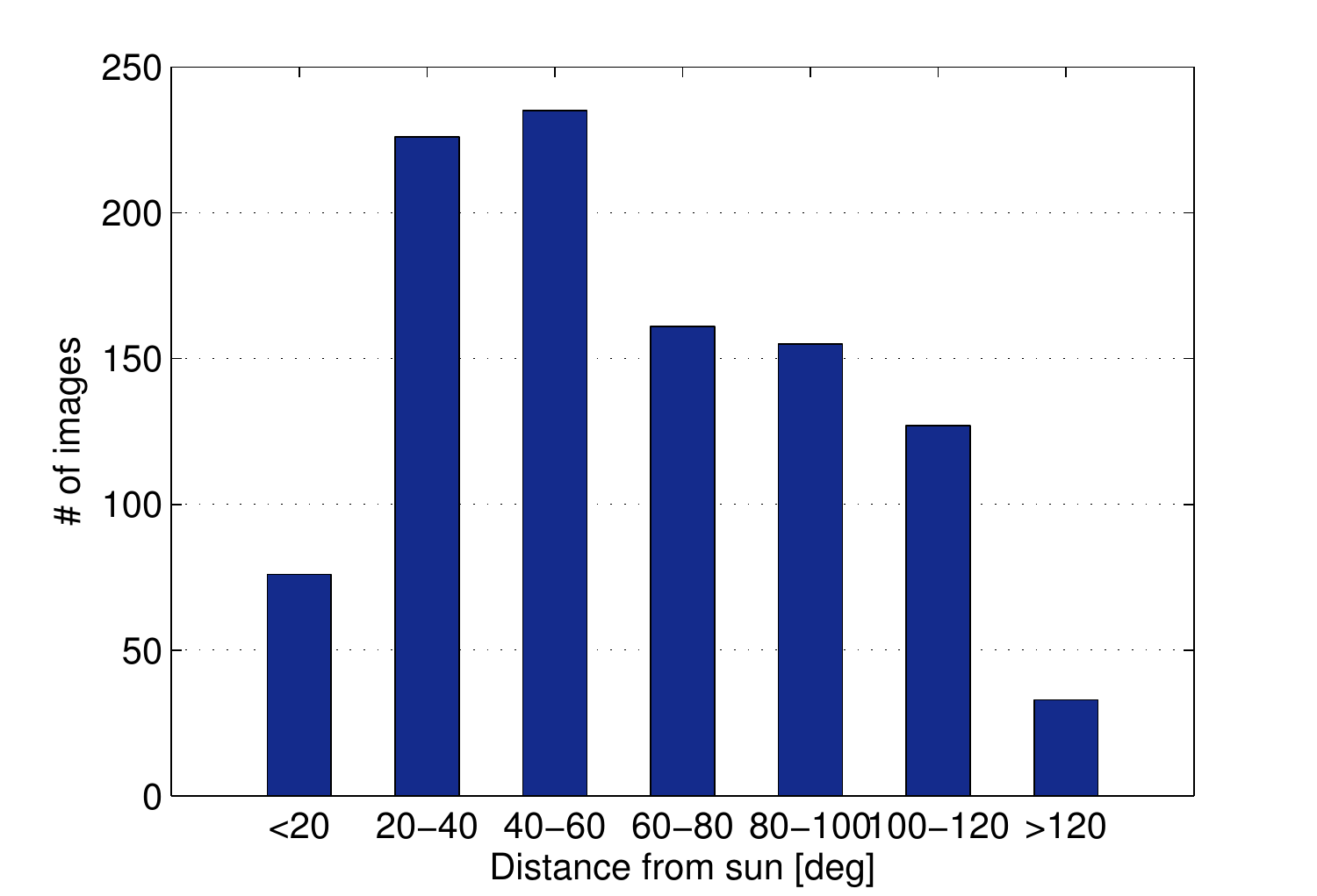}}
\caption{Distribution of images in SWIMSEG according to various characteristics.}\label{fig:image-stats}
\end{figure*}

SWIMSEG consists of $1013$ patches that were handpicked carefully from images taken over the period from October 2013 till July 2015 in Singapore. In the selection process, different parameters (viz.\ azimuth and elevation angle of sun, time of day, cloud coverage percentage, camera parameters) were considered. 
The camera elevation angle is set to three distinct angles ($36^{\circ}$, $45^{\circ}$, and $60^{\circ}$), whereas the azimuth angle is sampled at various angles around the unit hemisphere. The camera is set to auto mode while capturing pictures, and we observe that the settings tend to remain similar across all images. This is primarily because weather conditions in Singapore remain fairly similar throughout the year.

Fig.~\ref{fig:image-stats} characterizes the images in the dataset according to the time of day they were taken, cloud coverage, and distance from the sun. The sun rises and sets at around the same time throughout the year in Singapore, namely at 7am and 7pm local time.  The number of clear sky patches is relatively low as compared to moderately-cloudy or overcast conditions. This is because Singapore experiences a tropical climate in which clear-sky conditions are rare. Although most of the image patches are sampled at a certain distance from the sun to avoid saturation effects, we still include  a substantial number of images  close to sun.

\section{Experimental Evaluation}
\label{sec:results}

\subsection{Principal Component Analysis (PCA)}
\label{sec:pca}
We use PCA as described in Section \ref{sec:pca-theory} to check the degree of correlation between the 16 color channels, and also to identify the ones that capture most of the variance of the input data. 

\begin{figure}[htb]
\centering
\subfloat[HYTA]{\includegraphics[width=0.5\textwidth]{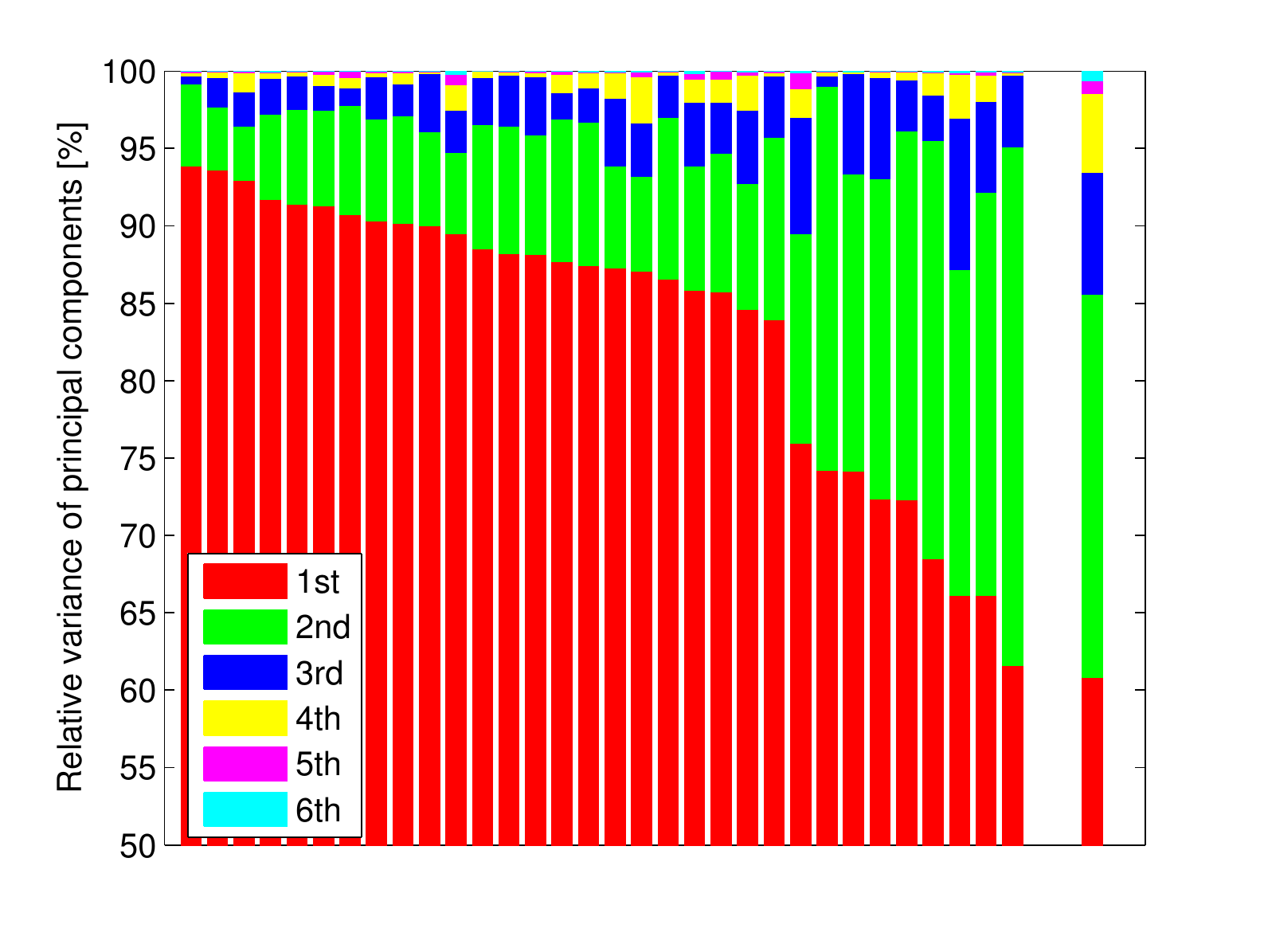}}\\
\vspace{-0.2cm}
\subfloat[SWIMSEG]{\includegraphics[width=0.5\textwidth]{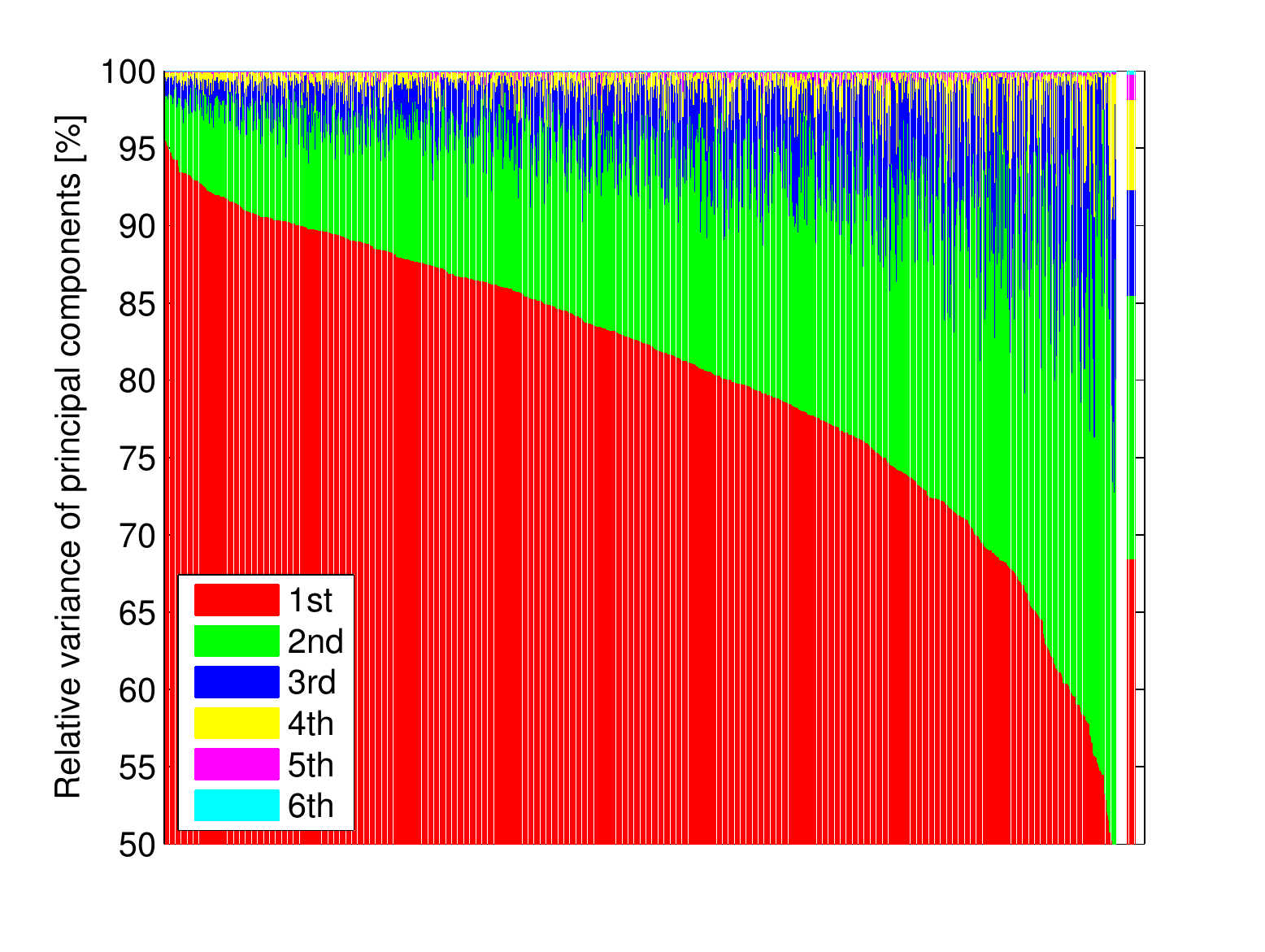}}
\caption{Distribution of variance across the principal components for all images in the HYTA and SWIMSEG databases. Each bar represents a single image. The rightmost bar in each plot shows the variance for the concatenation of all images (cf.\ Table \ref{table:PCA-table}; best viewed in color).}\label{fig:PCA_HYTA_WIMSEG}
\end{figure}

Figure~\ref{fig:PCA_HYTA_WIMSEG} shows the amount of variance captured by the most important principal components for HYTA and SWIMSEG databases. Table~\ref{table:PCA-table} shows the variance captured by each principal component for the concatenated distributions. In both databases, the first two principal components capture a large part of the variance across all the images, namely $85.6\%$ and $85.4\%$ of the total variance, respectively. 

\begin{table}[htb]
\small
\centering
\setlength{\tabcolsep}{3pt} 
\begin{tabular}{c||c|c|c|c|c|c}
  \hline
  \textbf{Database} & $\mathbf{1^{st}}$ & $\mathbf{2^{nd}}$ & $\mathbf{3^{rd}}$ & $\mathbf{4^{th}}$ & $\mathbf{5^{th}}$ & $\mathbf{6^{th}}$ \\
  \hline
  HYTA & $60.8\%$ & $24.8\%$ & $7.86\%$ & $5.11\%$ & $0.84\%$ & $0.36\%$ \\
  SWIMSEG & $68.44\%$ & $17.01\%$ & $6.82\%$ & $5.82\%$ & $1.64\%$ & $0.18\%$ \\
  \hline
\end{tabular}
\caption{Percentage of variance captured across different principal components for HYTA and SWIMSEG databases.}
\label{table:PCA-table}
\end{table}

\begin{figure}[htb]
\centering
\includegraphics[width=0.5\textwidth]{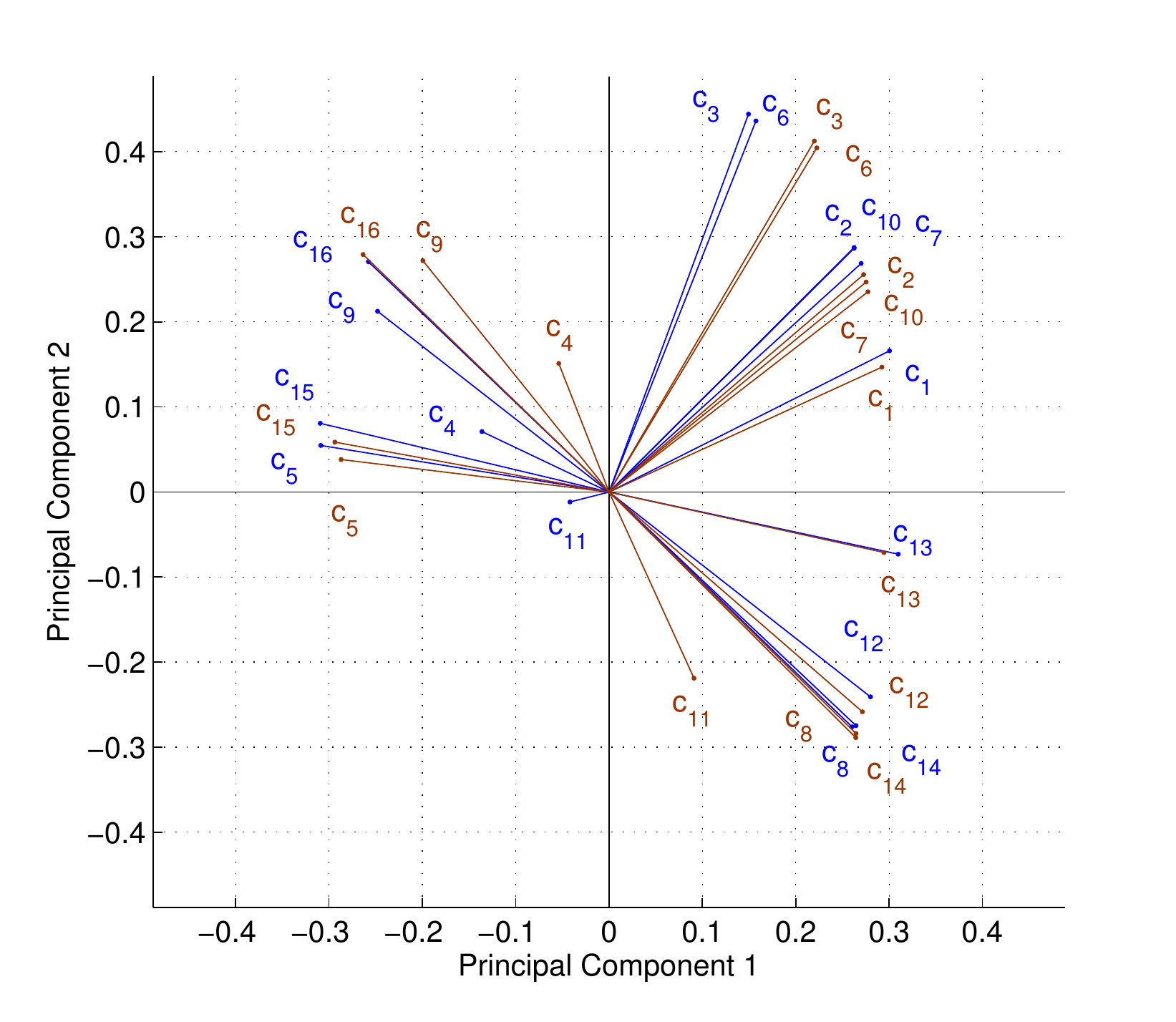}    
\caption{Bi-plot representation of $16$ color channels for HYTA (in blue) and SWIMSEG (in red) databases.}\label{fig:biplot_HYTA_WIMSEG}
\end{figure}

Fig.~\ref{fig:biplot_HYTA_WIMSEG} shows the bi-plots of $16$ color channels for HYTA and SWIMSEG. The bi-plots are consistent across both databases, with only small differences. We can also infer that certain color channel pairs such as $c_5$ and $c_{15}$, $c_8$ and $c_{14}$, etc. are highly correlated with each other. The most noticeably different channel is $c_{11}$ ($a^{*}$), largely because the red/green axis is not very useful for distinguishing sky from cloud and contains mostly noise. 

\subsection{Loading Factors and ROC Curves}
\label{sec:which-color}

In order to determine the most appropriate color channel(s) for effective segmentation of sky/cloud images, we calculate the loading factors, i.e.\ the projection of the input vectors in the bi-plot onto the $1^{st}$ principal component axis,  as described in Section \ref{sec:pca-theory}. The channels with the highest loading factors are  most discriminatory and thus favorable for segmentation.  Furthermore, we use the area between the ROC curve and the random classifier slope to identify the important color channels in the sky/cloud binary image classification task, as described in Section \ref{sec:bimodality}.

\begin{figure}[htb]
\centering
\subfloat[HYTA]{\includegraphics[width=0.5\textwidth]{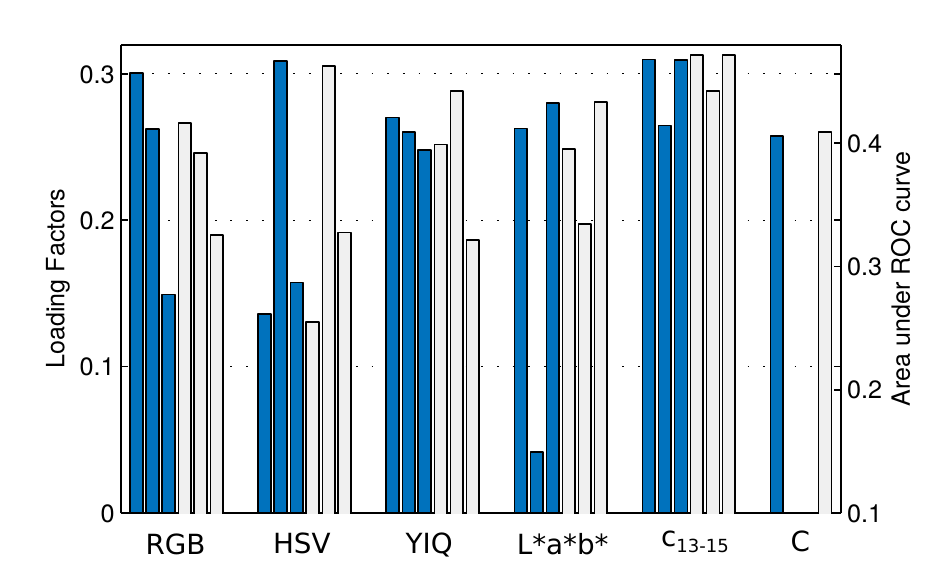}}\\
\subfloat[SWIMSEG]{\includegraphics[width=0.5\textwidth]{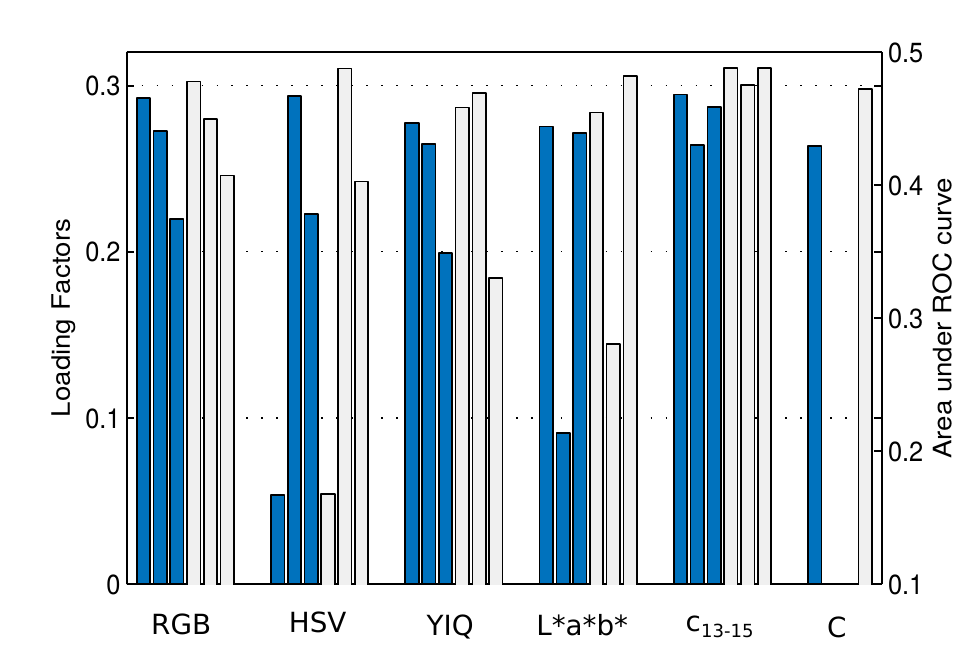}}
\caption{Selection of color channels for effective sky/cloud image segmentation. We represent the loading factors in blue bars (primary y-axis), and area under the ROC curve in gray bars (secondary y-axis), for HYTA and SWIMSEG databases. The $16$ color channels are grouped by color space. Each individual bar represents a single color channel.}
\label{fig:select-CC}
\end{figure}

Figure~\ref{fig:select-CC} shows the results obtained for the loading factors and the area under ROC curve for all color channels of the concatenated distribution in HYTA and SWIMSEG databases. We see a clear trend amongst these values across the different color channels. Red/blue ratio channels $c_{13,15}$ and saturation channel $c_5$ rank high. On the other hand, color channels like $c_4$ and $c_{11}$ are not discriminatory and thereby rank lower. Also, as could be expected from the bi-plots, the results are again perfectly consistent across both databases (except for $c_{11}$). The color channels with the maximum contribution to the $1^{st}$ principal component are $c_{13}$, $c_{15}$ and $c_5$. Therefore we consider these color channels good candidates for our segmentation problem.

\section{Segmentation Results}
\label{sec:result-segment}
In the previous Section, we identified suitable candidate color components for segmentation, such as $c_{5}$ (saturation) and various blue-red ratios ($c_{13}$, $c_{14}$, $c_{15}$). These color channels are expected to perform better than others. We now evaluate the segmentation results for all $16$ color channels, by individually considering them as the discriminatory feature vector $\mathbf{X}^{f} \in {\rm I\!R}^{mn \times 1}$ for segmentation. We showed previously \cite{ICIP1_2014} that there is hardly any performance improvement when using more than $1$ color channel in the feature matrix $\mathbf{X}_f$.

We perform extensive experiments of our proposed approach on the HYTA and SWIMSEG databases. Each of the images in the two datasets are expressed in the form of $\hat{\mathbf{X}_{i}}$, as shown in Eq.\ (\ref{eq:eq1}). The corresponding feature vector $\mathbf{X}_i^{f}$ is computed for both training and testing stages.  The training images are used to calculate the matrix of regression coefficients $\mathbf{B}$ as per Eq.\ (\ref{eq:pls6}). In our experiments, we use \texttt{plsregress}, the MATLAB implementation of the SIMPLS algorithm \cite{SIMPLS} for this. 
We set $p=k=1$, as we use a single color channel for evaluation.
During the testing phase, $\mathbf{X}_{\mathrm{test}}$ of the testing image is computed. Using the matrix of regression coefficients $\mathbf{B}$, the corresponding $\mathbf{Y}_{\mathrm{test}}$ is computed. 

In the ideal scenario, the value of $\mathbf{Y}_{\mathrm{test}}$ is either $0$ or $1$. However, in the actual implementation, the values are continuous in nature. Suppose we denote the obtained values of $\mathbf{Y}_{\mathrm{test}}$ as $\{v_1, v_2, \ldots, v_{mn} \}$. Let $\theta$ be the function that normalizes the obtained values of $\mathbf{Y}_{\mathrm{test}}$ to the interval $[0,1]$:
$$\theta(v_i)=\frac{v_i - \min v_i}{\max v_i-\min v_i}.$$ 
We perform \emph{soft} thresholding by normalizing these values in the range [0,1]. Therefore, this normalized value indicates the belongingness of a pixel to belong to the cloud (or sky) category. 

The objective evaluation is based on the true positive (\emph{TP}), true negative (\emph{TN}), false positive (\emph{FP}), and false negative (\emph{FN}) samples in a binary image. We report Precision, Recall, F-score, and Misclassification rate for the different benchmarking algorithms, which are defined as follows: 
\begin{gather*}
\mbox{Precision}=TP/(TP+FP),\\
\mbox{Recall}=TP/(TP+FN),\\
\mbox{F-score}=\frac{2\times\mbox{Precision}\times\mbox{Recall}}{{\mbox{Precision}+\mbox{Recall}}},\\
\mbox{Misclassification rate}=\frac{FP+FN}{TP+TN+FP+FN}.
\end{gather*}

\subsection{Choice of Color Channels}
As described earlier, the segmentation performance is dependent on the choice of color channels. In order to understand the effect of color channels, we use the individual color channels to constitute the feature vector $\mathbf{X}^{f}$ in this experiment. We choose training and test set sizes as per Table~\ref{table:train-test}. 

\begin{table}[htb]
\small
\centering
\begin{tabular}{r||c|c|c}
  \hline
  \textbf{Database}  & \textbf{Training Set} & \textbf{Test Set} & \textbf{Total}  \\
  \hline
  HYTA & 17 & 15 & 32 \\
  SWIMSEG & 500 & 513 & 1013 \\
  \hline
\end{tabular}
\caption{Number of training and test images for HYTA and SWIMSEG databases.}
\label{table:train-test}
\end{table}

\begin{figure}[htb]
\centering
   \subfloat[HYTA]{\includegraphics[width=0.5\textwidth]{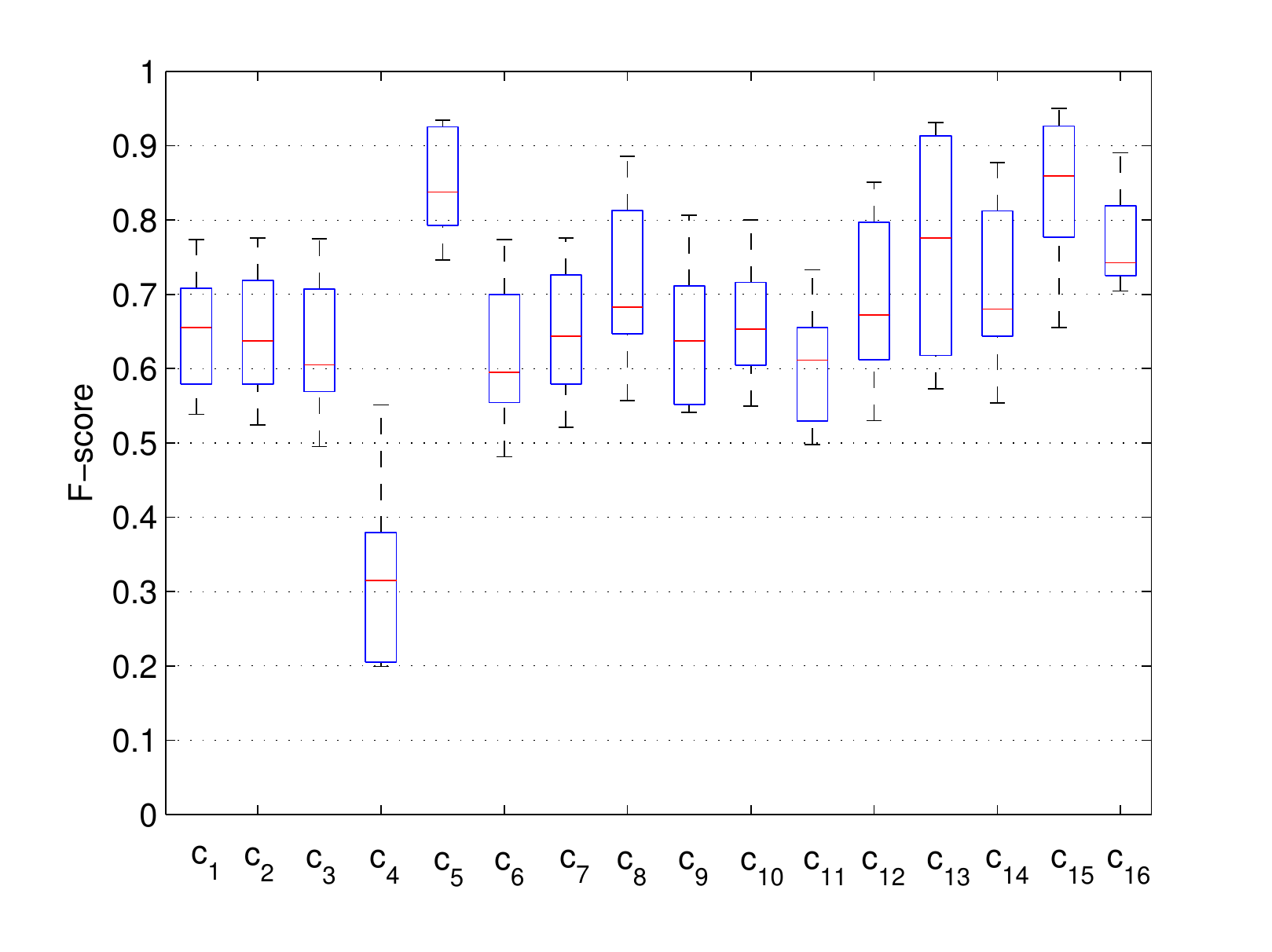}}\vspace{-0.4cm}\\
   \subfloat[SWIMSEG]{\includegraphics[width=0.5\textwidth]{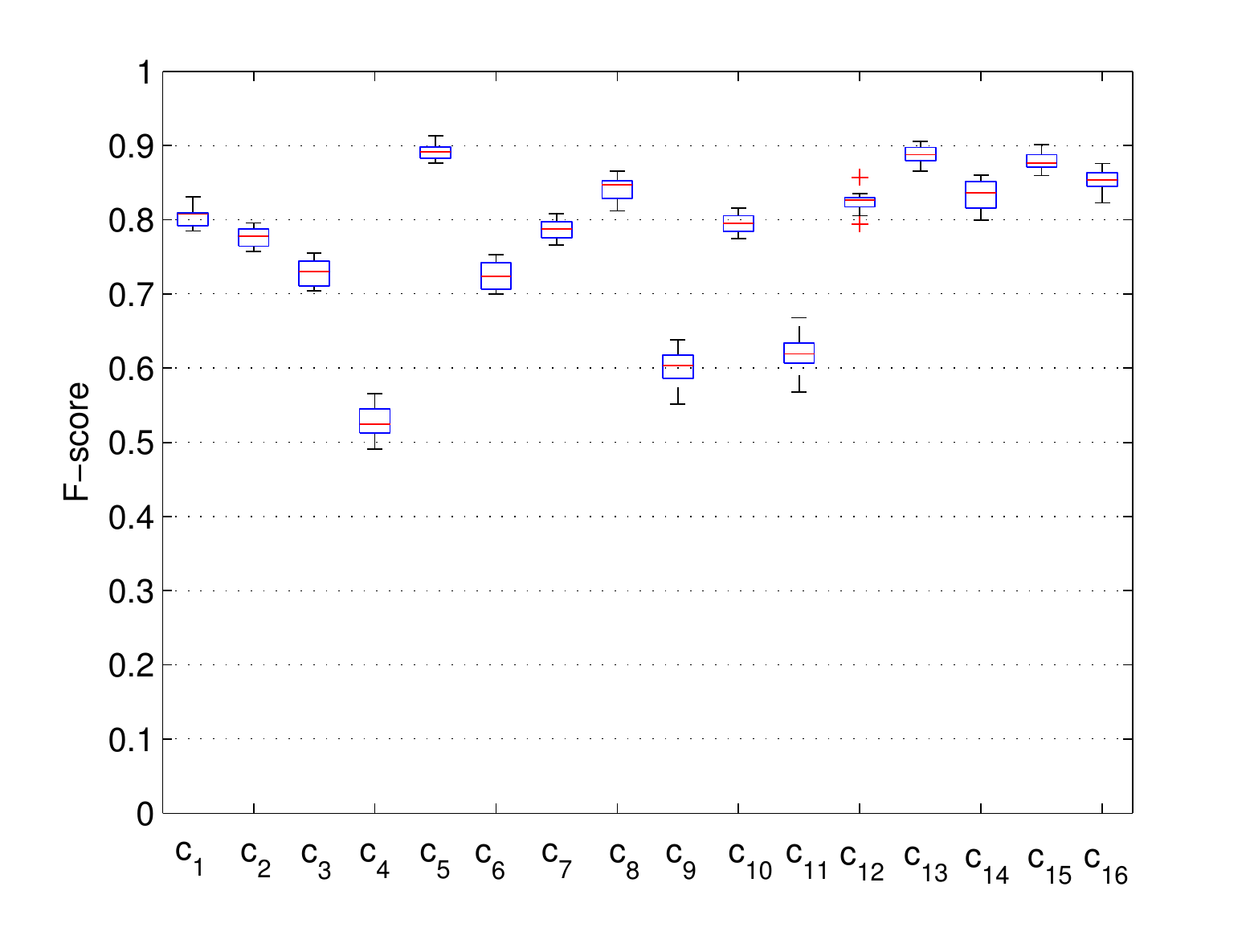}}
\caption{Binary classification performance for different color channels.}\label{fig:CV-result}
\end{figure}

We compute F-scores (the harmonic mean of precision and recall) to characterize the segmentation performance of each color channel. We perform a $5$-fold and $10$-fold cross validation in HYTA and SWIMSEG datasets, respectively. In each of these experiments, $1$ fold is used for testing and the remaining folds are used in the training stage. Figure~\ref{fig:CV-result} shows the results for HYTA and SWIMSEG datasets. 

The classification performance for individual color channels follows the same trend in both HYTA and SWIMSEG dataset. Color channels $c_5$ and the red-blue ratio channels $c_{13,15}$ perform best, which is consistent with what we found in Section~\ref{sec:results}.  Based on these results, we conduct the subsequent evaluations with $c_{15}$ as the feature vector $\mathbf{X}^{f}$.

\subsection{Effect of Database Characteristics}
\begin{figure*}[htb!]
\centering
\subfloat[Time of day]{\includegraphics[height=0.24\textwidth]{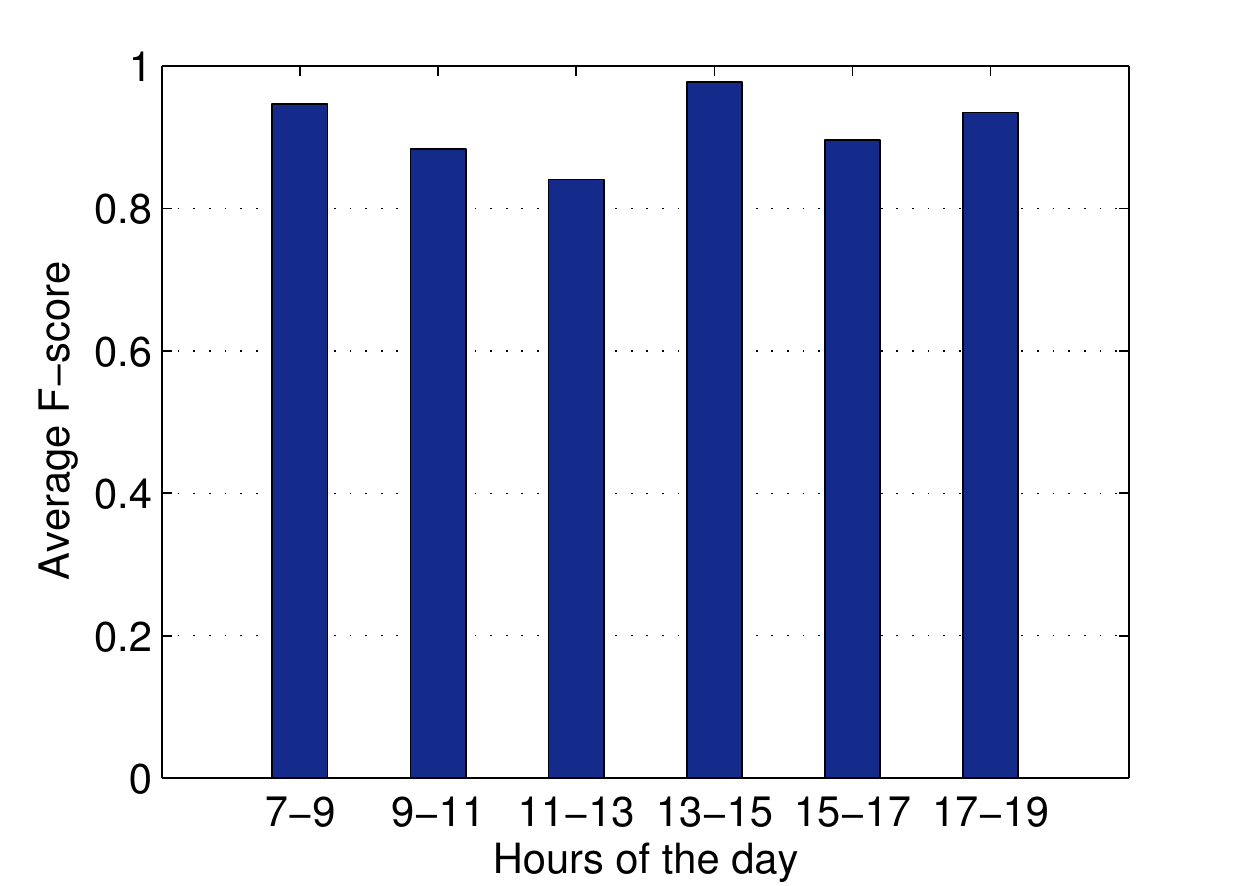}}\hspace{-0.5cm}
\subfloat[Cloud coverage]{\includegraphics[height=0.24\textwidth]{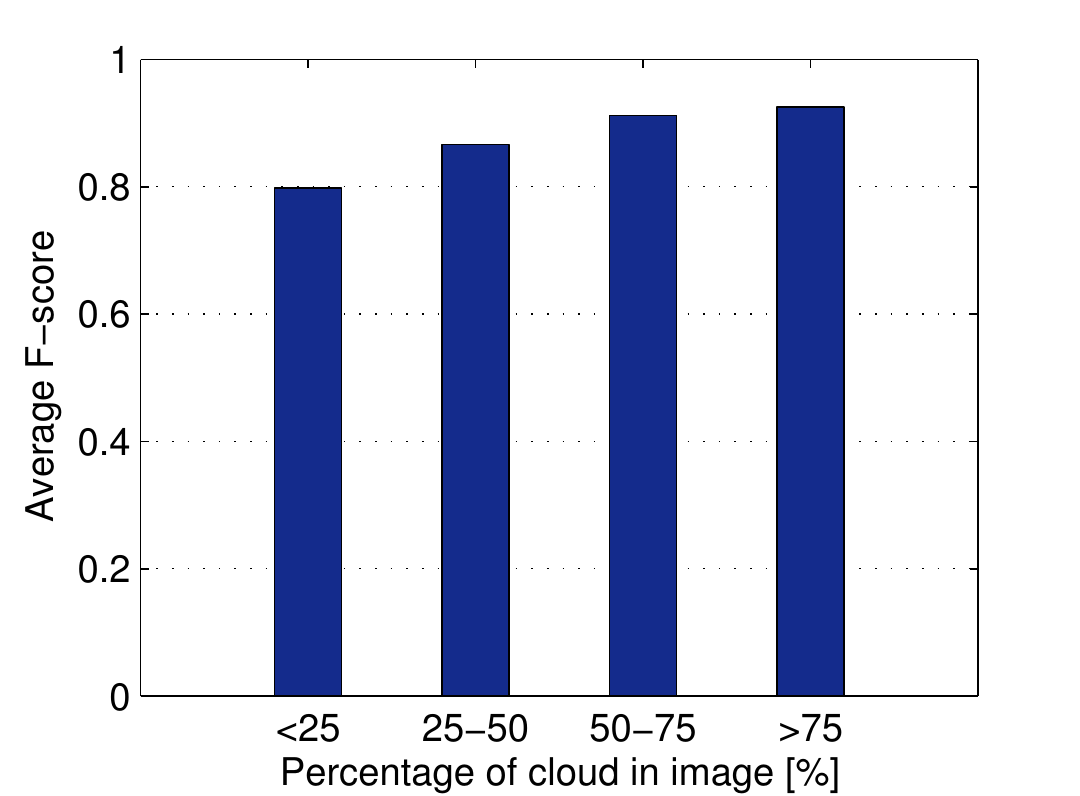}}\hspace{-0.5cm}
\subfloat[Distance from the sun]{\includegraphics[height=0.24\textwidth]{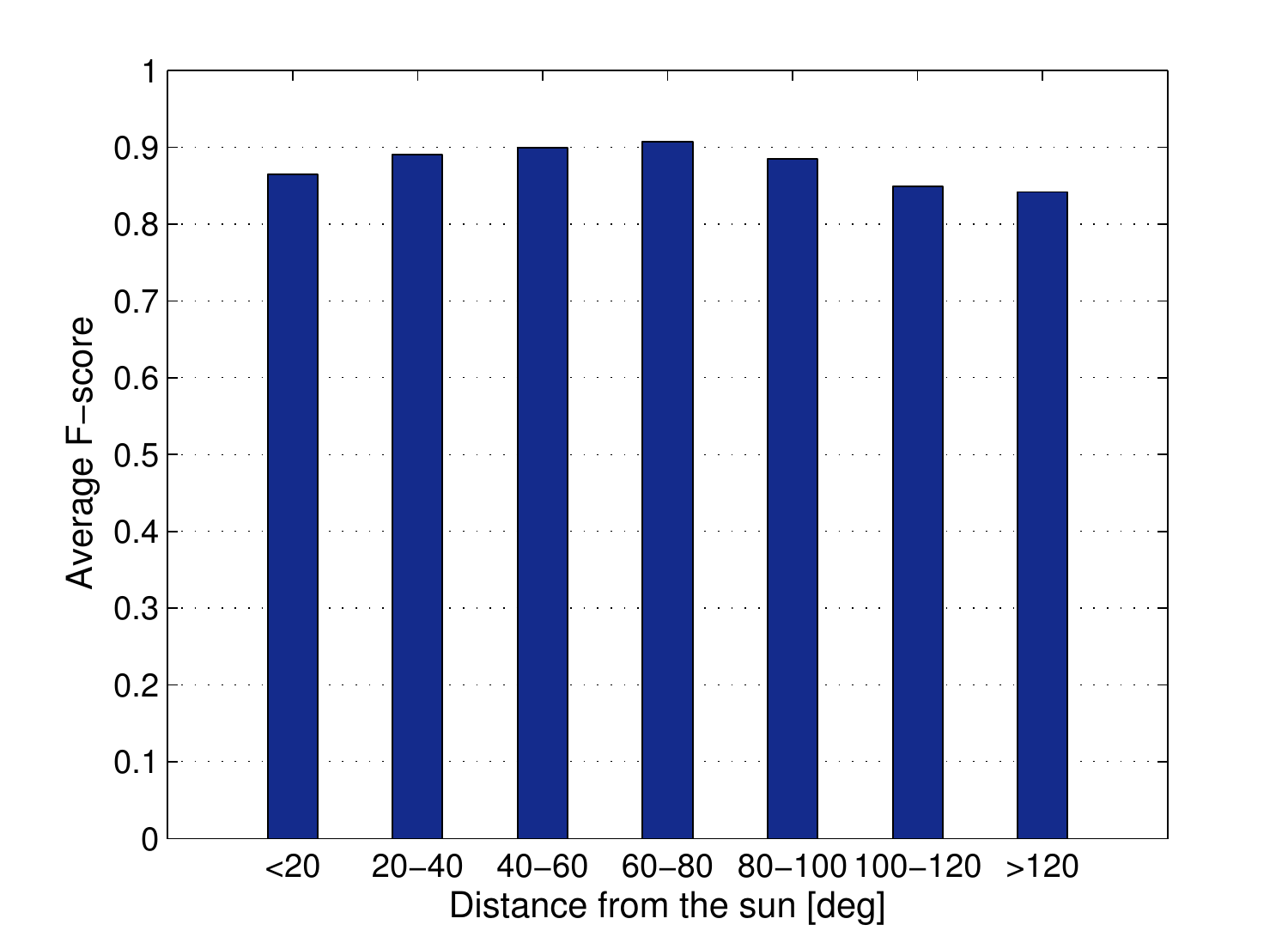}}
\caption{Distribution of average F-score value as a function of various characteristics of the images in SWIMSEG.}\label{fig:fscore-distribution}
\end{figure*}

We study the effect of the different characteristics of the database on the obtained F-score values of our proposed segmentation algorithm by checking whether the obtained segmentation results are biased towards a particular set of image parameters, specifically time of day, cloud coverage, and distance from the sun. We present the distribution of F-score values across these parameters in Fig.~\ref{fig:fscore-distribution}. 

The average F-score is slightly high in the morning and late afternoon, while it drops during mid-day. This is because the images captured during mid-day have a higher tendency to get over-exposed because of the presence of direct sun. On the other hand, the images captured at the morning and evening hours of the day receive slant sun-rays, and contain more clouds. The variation of the F-score values w.r.t. the percentage of cloud in an image is shown in Fig.~\ref{fig:fscore-distribution}(b). We observe that the performance is marginally better for higher percentage of cloud coverage in an image. Finally, we note that there is no clear relationship of the distance of the image patch from the sun on the obtained F-score values, which is illustrated in Fig.~\ref{fig:fscore-distribution}(c).

\subsection{Benchmarking}
We benchmark our proposed approach with current state-of-the-art cloud detection algorithms: Li et al.\ \cite{Li2011}, Souza et al.\ \cite{Souza}, Long et al.\ \cite{Long}, and Mantelli-Neto et al.\ \cite{Sylvio}. We also compare against another popular image segmentation algorithm viz. SLIC superpixels \cite{SLIC}. 

Li et al.\ \cite{Li2011} use a Hybrid Thresholding Algorithm (HYTA) that combines both adaptive and fixed threshold approach. Souza et al.\ \cite{Souza} use the saturation component of the intensity, hue, and saturation (IHS) color model in defining appropriate thresholds to obtain the final segmentation mask. Long et al.\ \cite{Long} use the ratio of red and blue channels to model the molecular scattering in the atmosphere. Mantelli-Neto et al.\ \cite{Sylvio} use multi-dimensional Euclidean geometric distance and Bayesian approaches to identify sky and cloud pixels. 

The popular superpixel segmentation Simple Linear Iterative Clustering (SLIC) divides the input image into a number of superpixels. We combine it with the DBSCAN algorithm~\cite{DBSCAN} to cluster the superpixels and subsequently categorize them into sky/cloud based on the average ratio of red and blue of a particular superpixel. 

In addition to these state-of-the-art cloud segmentation algorithms, we also compare our proposed algorithm with other typical machine-learning based methods. We extract discriminative features from image, and use them in a classifier model. In our case, we employ SVM classifier along with various image features. 

We use several feature descriptors viz. Local Binary Patterns (LBP)~\cite{LBP_PAMI}, Color Histogram (colorHIST)~\cite{colorhist-CVPR}, and Dense Scale Invariant Feature Transform (dSIFT)~\cite{SIFT} for extracting discriminatory features from sky/cloud images. Quite recently, Yuan et al.\ \cite{BOW-cloud} used Bag-Of-Words (BOW) model to model cloud- and non-cloud- satellite images from dSIFT local features. Yuan et al.\ \cite{BOW-cloud} over segments a satellite image into sub-images using SLIC superpixel generation method. Image descriptors are thereby extracted from these sub-patches using dense SIFT extractor. Subsequently, the BOW model is used to model the cloud- and non-cloud images and the corresponding codebook is generated. A Support Vector Machine (SVM) classifier is finally used for the super-pixel classification into cloud- and non-cloud- parts \cite{BOW-cloud}.
In addition to these features, we also use texture descriptors (using Schmid filter~\cite{Schmid_CVPR}) along with BOW and SVM as a benchmarking cloud segmentation algorithm.
Furthermore, we also use the raw gray-scale (GRAY) color channel values as discriminatory feature in feature extraction stage. These features are used in training a SVM classifier. In the testing stage, these descriptors are again computed for the test images, and the trained SVM classifier classifies them to either sky or cloud pixels. Therefore, these different image features together with SVM classifier form a collection of machine-learning based sky/cloud classification algorithms.

\begin{figure}[htb]
\centering
   \subfloat[HYTA]{\includegraphics[height=0.21\textwidth]{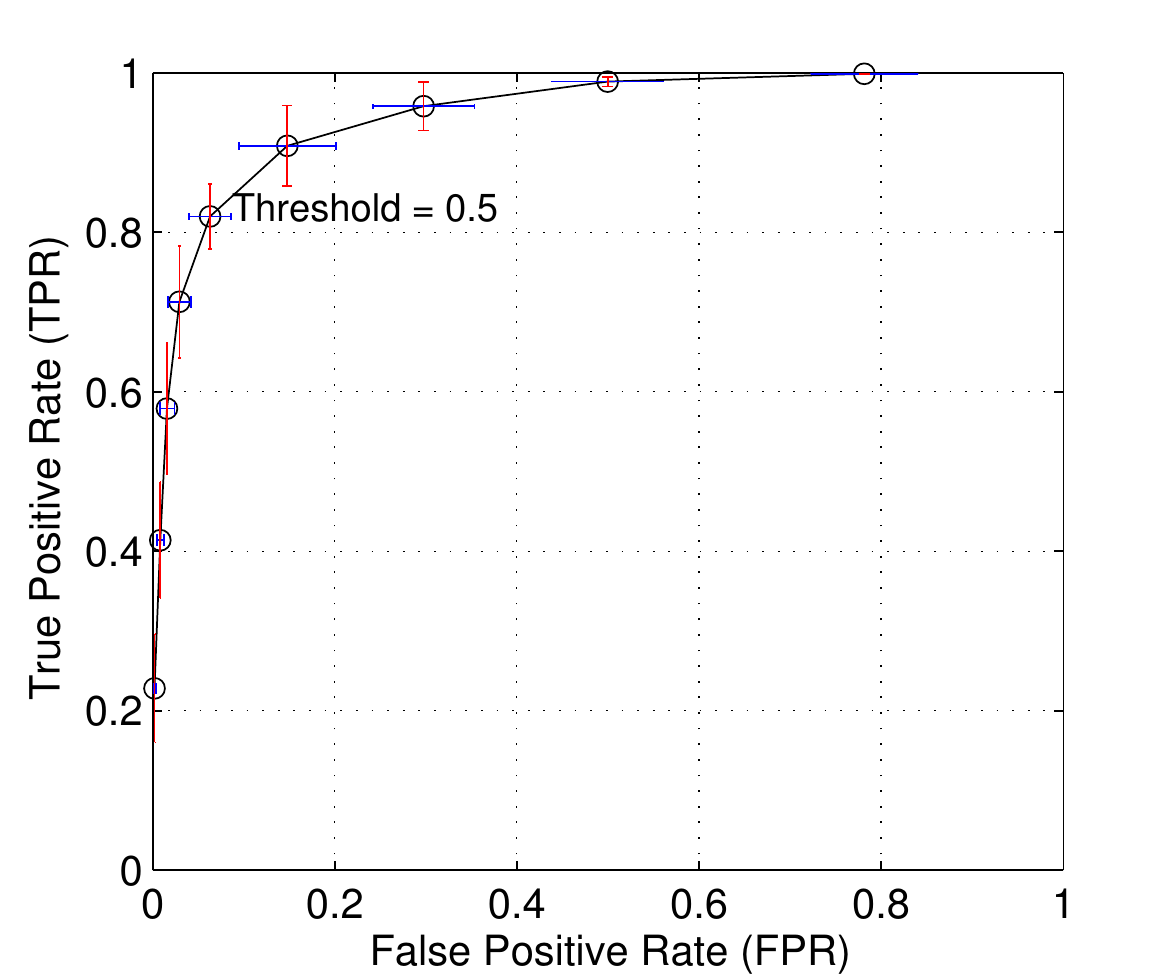}}
   \subfloat[SWIMSEG]{\includegraphics[height=0.21\textwidth]{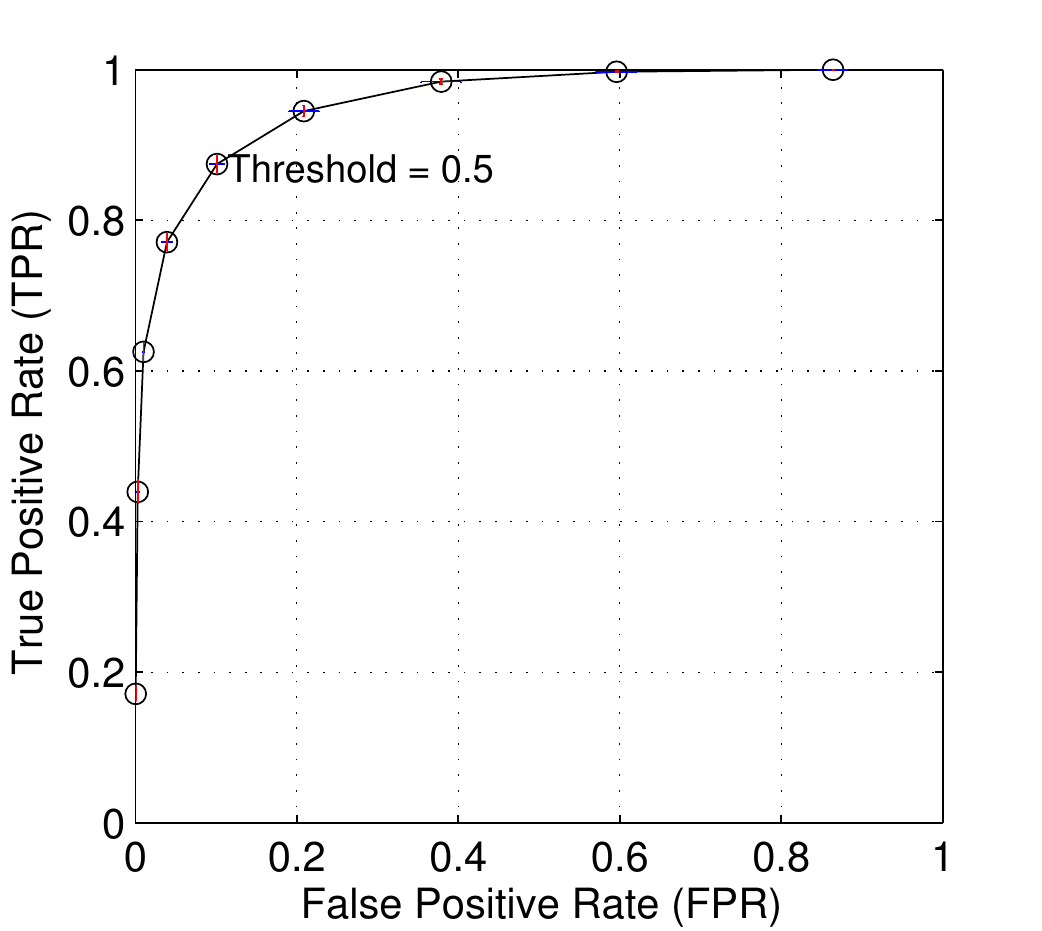}}
\caption{ROC curves of our proposed algorithm, including error bars in X- and Y-directions.}
\label{fig:ROC-prop}
\end{figure}

In our proposed approach, we assign a membership value for the cloud category to each pixel. However, as the available ground-truth masks are binary, we convert our probabilistic map into a binary map by simple thresholding in order to perform the objective evaluations.
The threshold value can be any value in the closed interval [0,1]; a lower value indicates a higher tendency for a pixel to classify as \emph{cloud} and a higher value indicates a higher tendency to classify as \emph{cloud} pixel. Figure~\ref{fig:ROC-prop} shows the ROC curve for our proposed algorithm by changing the threshold value in the closed interval [0,1] for both HYTA and SWIMSEG databases.  $20$ trials were performed for each threshold value, with training and testing sets of images chosen randomly for each trial.  The subsequent results were obtained with a threshold of $0.5$.

Figure~\ref{fig:sample-results} shows a few sample images from the HYTA and SWIMSEG databases together with the segmentation ground truth and the corresponding output images obtained by our proposed approach and other existing methods.  Most of the approaches work quite well when there is a clear contrast between sky and cloud regions, as is often the case for the HYTA database. The method of Mantelli-Neto et al.\ \cite{Sylvio} fails on images that contain non-uniform illumination of cloud mass from sunlight. The detection of thin clouds is  the most problematic: Long et al.\ \cite{Long} and SLIC+DBSCAN methods classify the entire image patch as cloud in such cases. The method of Li et al.\ \cite{Li2011} fails in cases where the boundary between sky and cloud is ill-defined. We observe from the last two examples in Fig.~\ref{fig:sample-results} that Li et al.\ over-segments the corresponding input images. Except for Souza et al. \cite{Souza}, most of the approaches fail for images of the SWIMSEG dataset with generally less well-defined sky/cloud separation. 
Similarly, the other machine-learning based methods viz. GRAY+SVM and dSIFT+BOW+SVM ~\cite{BOW-cloud} perform poorly for the wide range of images. This is because the Bag-of-words model obtained for cloud- and non-cloud- superpixels fails to discriminate between the superpixels in the test images.
The results obtained by our PLS-based approach show that it works well across diverse illumination conditions and for both databases. 

\begin{table*}[htb!]
\begin{tabular}{m{1.3cm}cccccc}
{\fontsize{0.4cm}{1em}\selectfont Input Image} \vspace{1.5cm}& 
\includegraphics[height=0.08\textheight]{B1.pdf} &
\includegraphics[height=0.08\textheight]{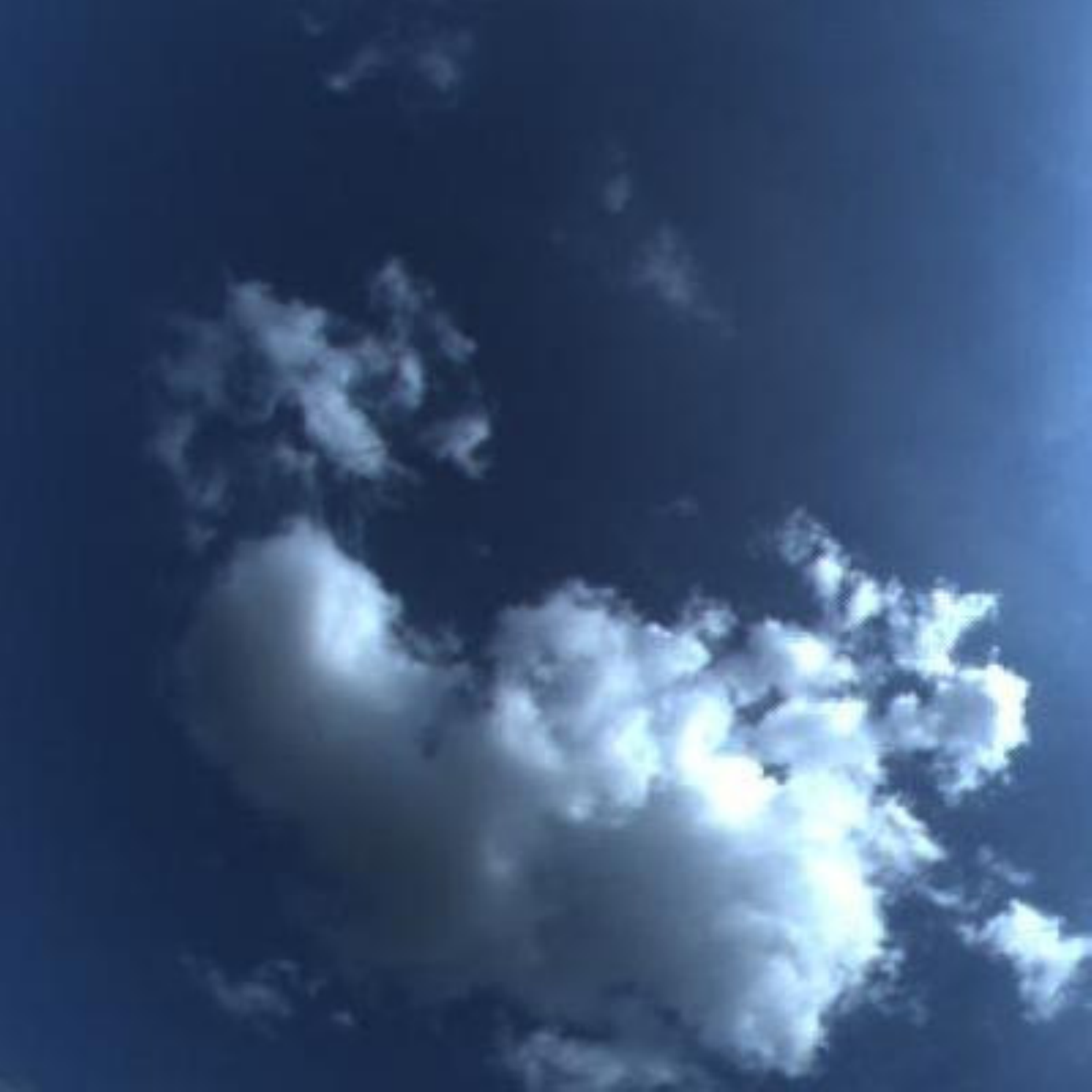} &
\includegraphics[height=0.08\textheight]{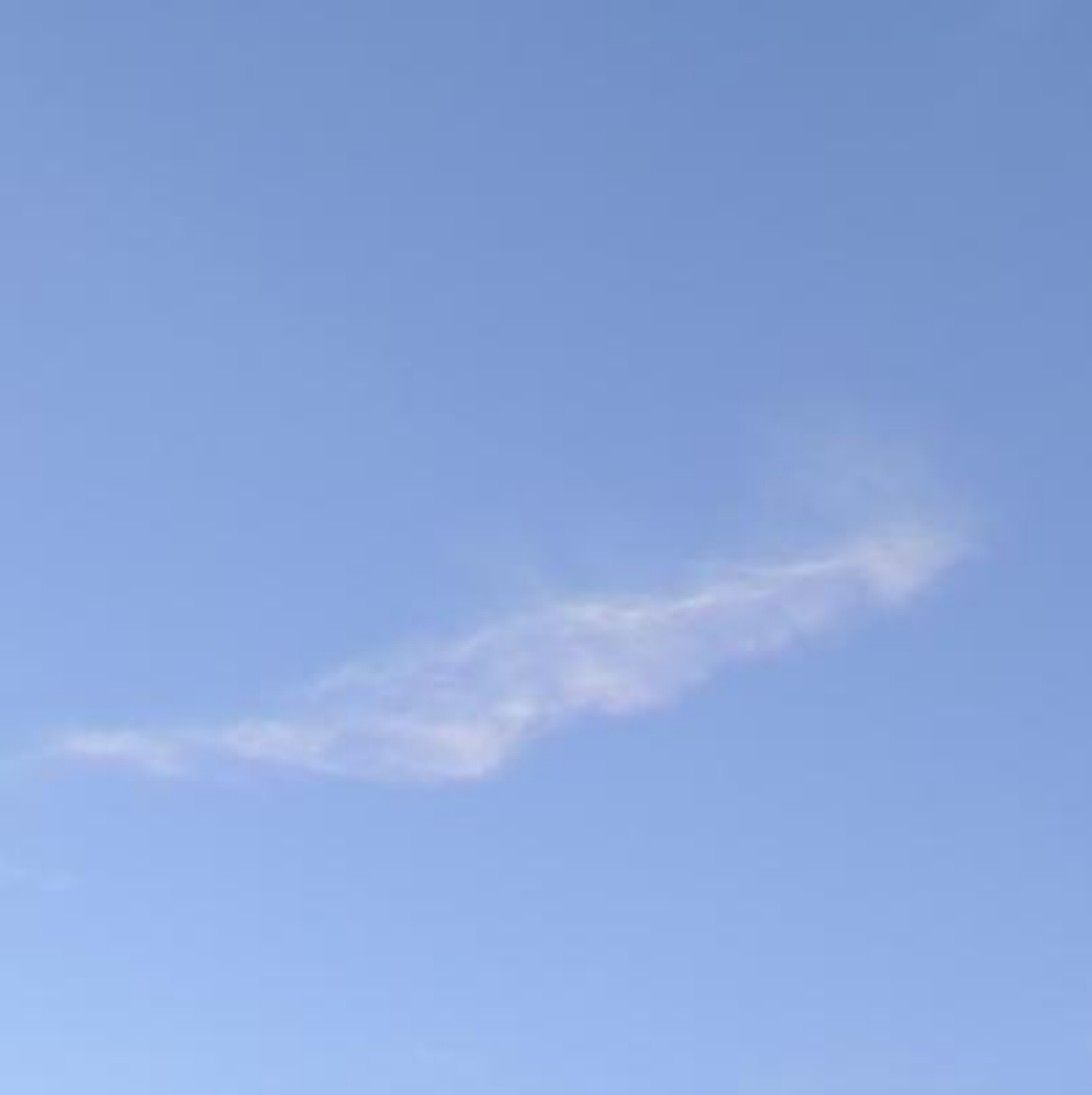} &
\includegraphics[height=0.08\textheight]{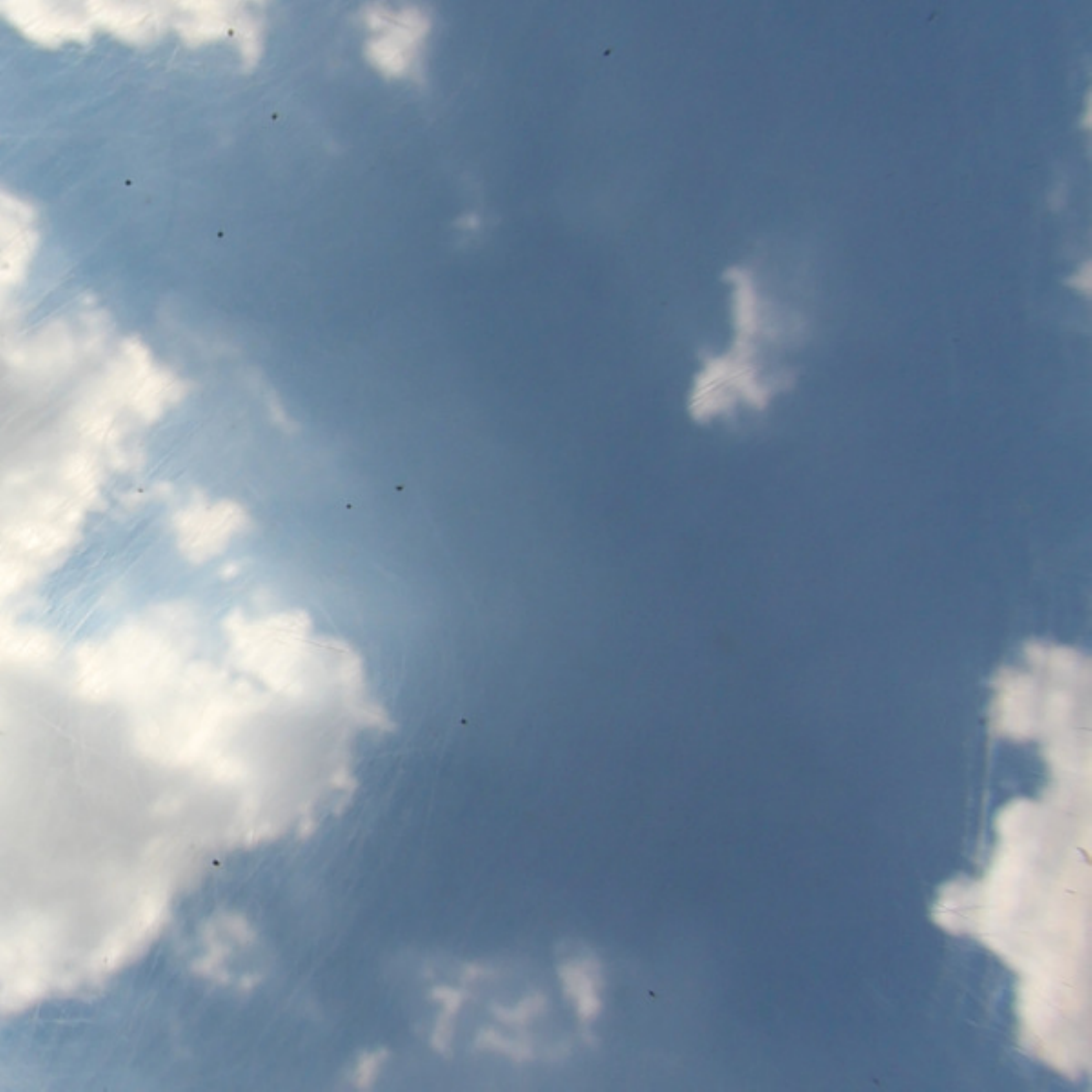} &
\includegraphics[height=0.08\textheight]{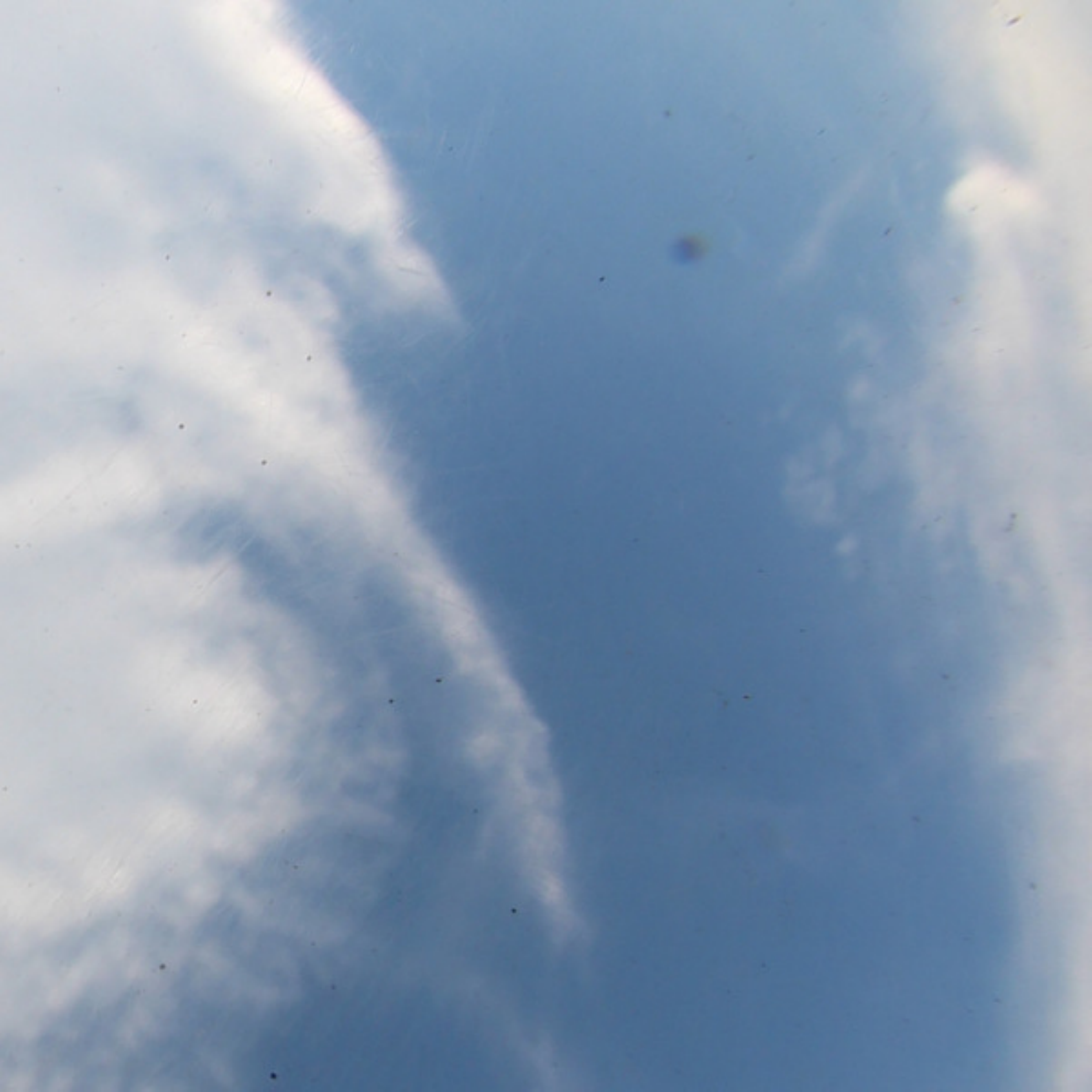} &
\includegraphics[height=0.08\textheight]{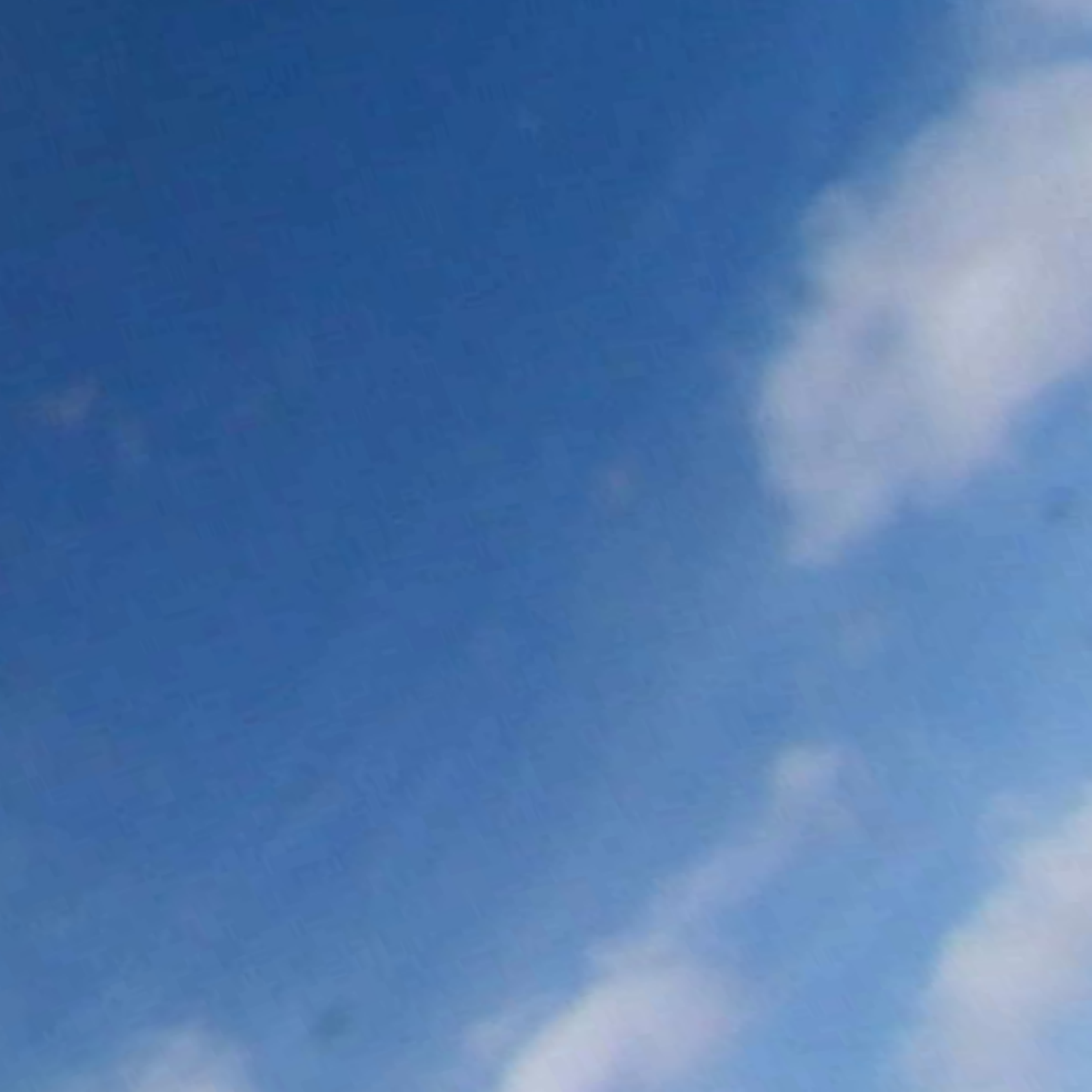} \vspace{-0.39in}\\

{\fontsize{0.4cm}{1em}\selectfont Ground Truth} \vspace{1.5cm}& 
\includegraphics[height=0.08\textheight]{B1_GT.pdf} &
\includegraphics[height=0.08\textheight]{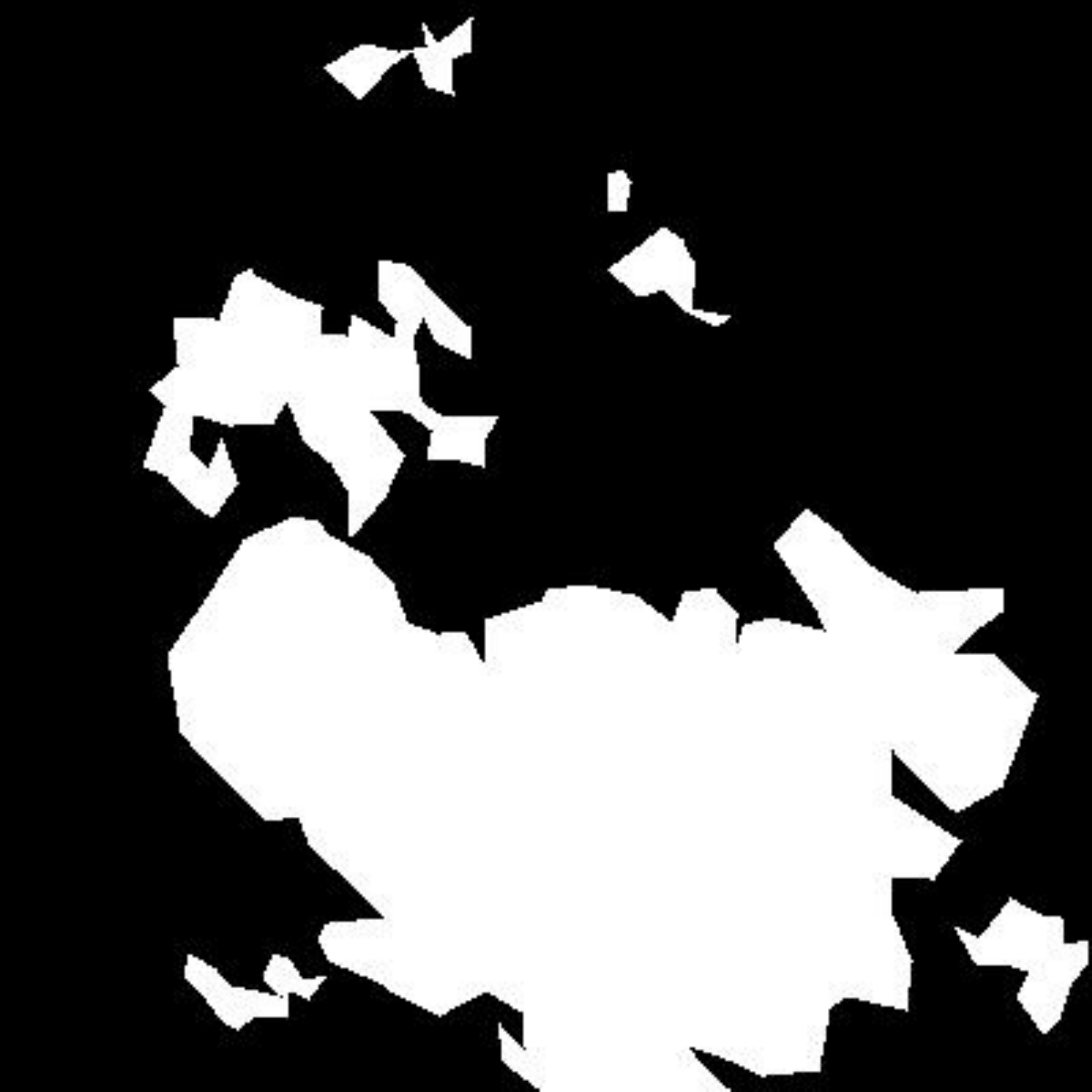} &
\includegraphics[height=0.08\textheight]{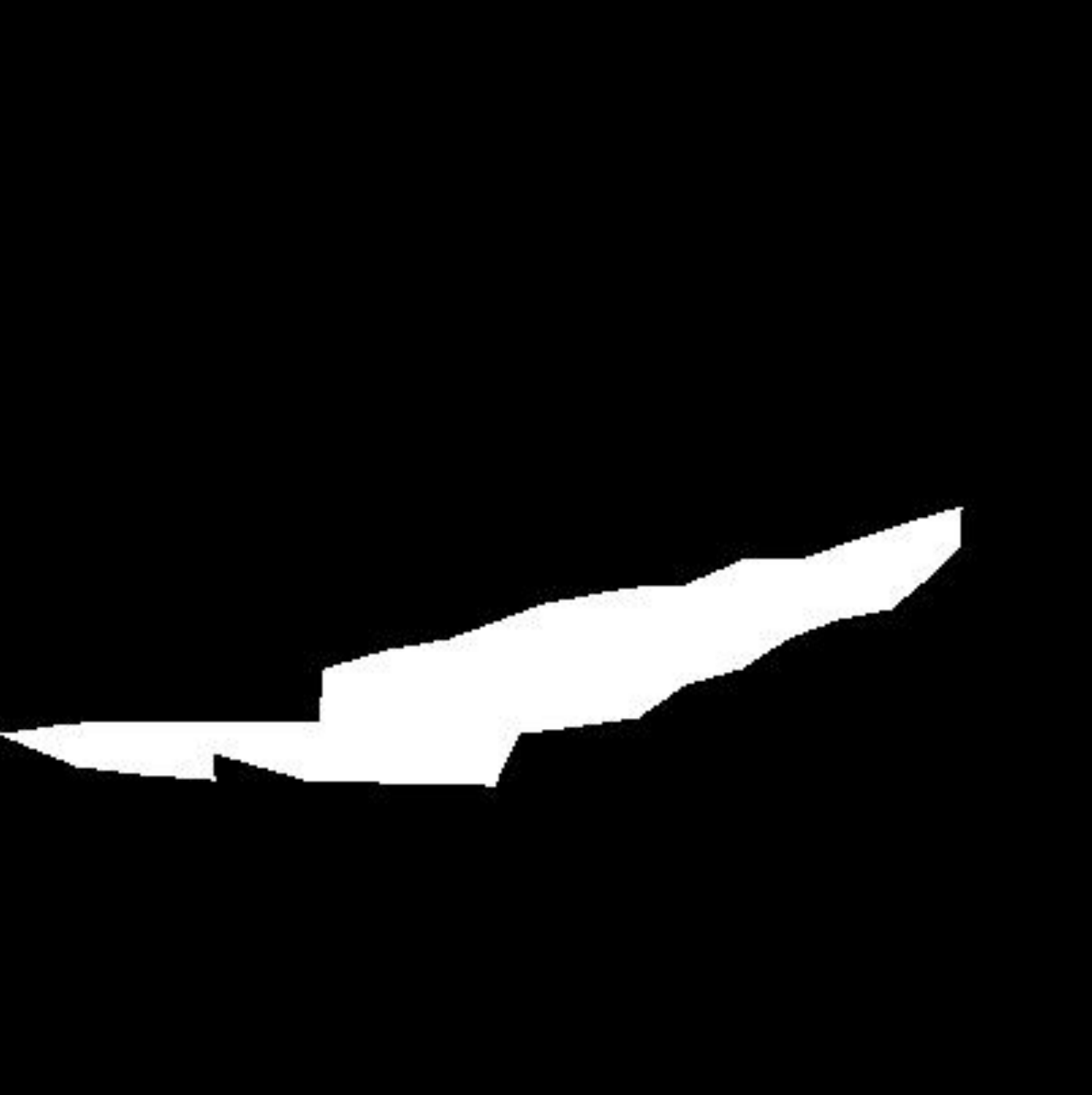} &
\fbox{\includegraphics[height=0.08\textheight]{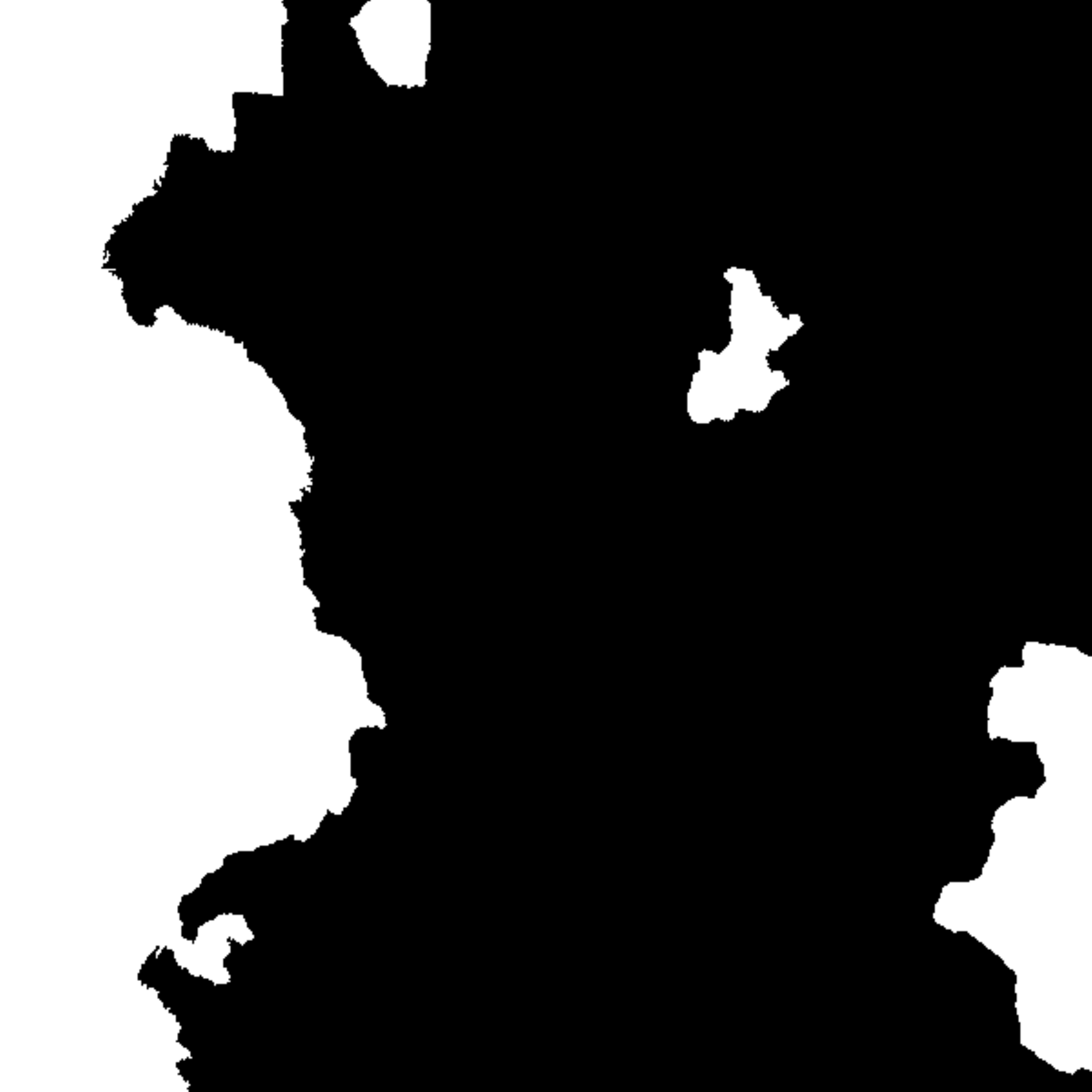}} &
\fbox{\includegraphics[height=0.08\textheight]{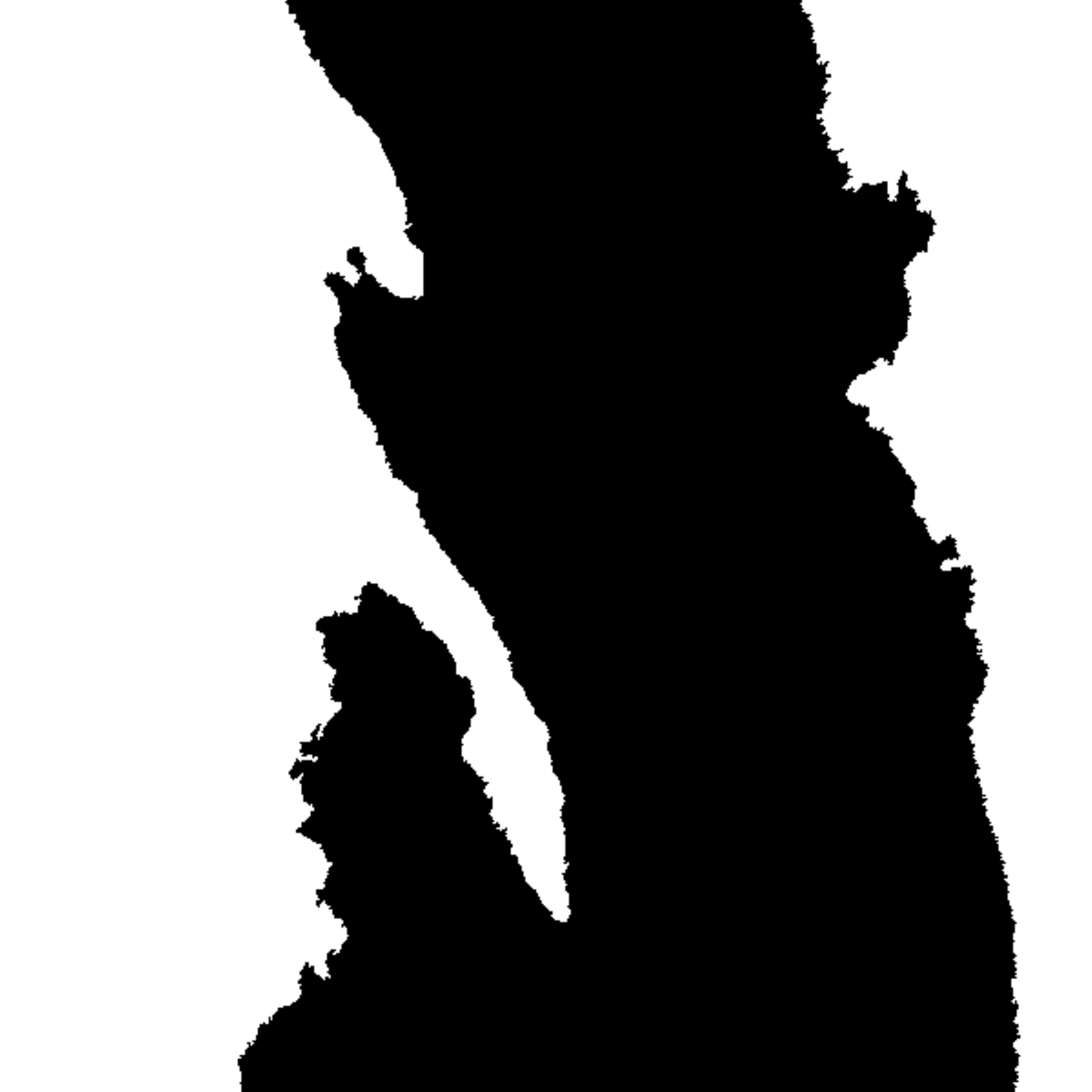}} &
\fbox{\includegraphics[height=0.08\textheight]{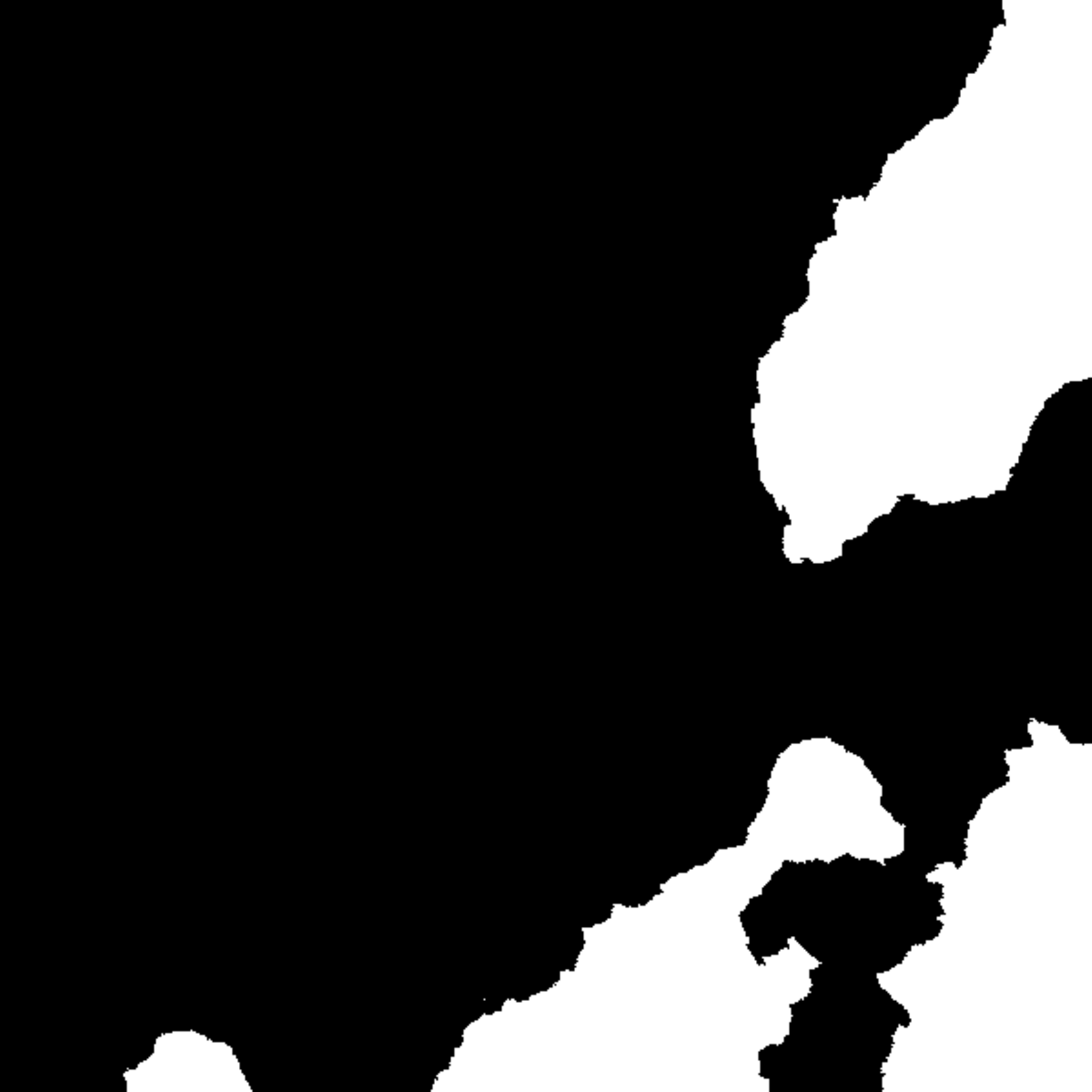}} \vspace{-0.38in}\\

{\fontsize{0.4cm}{1em}\selectfont Our \mbox{approach}} \vspace{1.5cm}& 
\includegraphics[height=0.08\textheight]{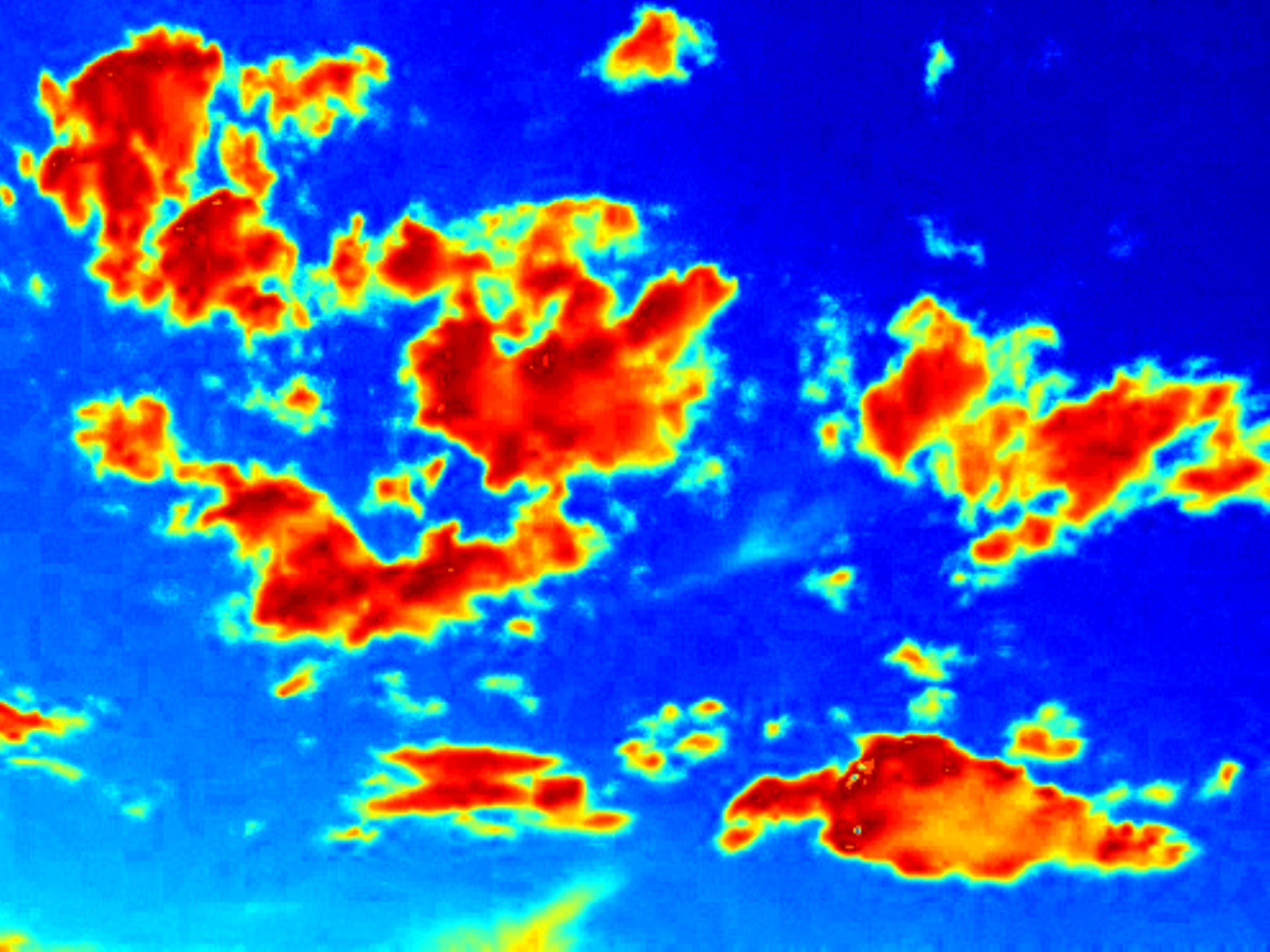} &
\includegraphics[height=0.08\textheight]{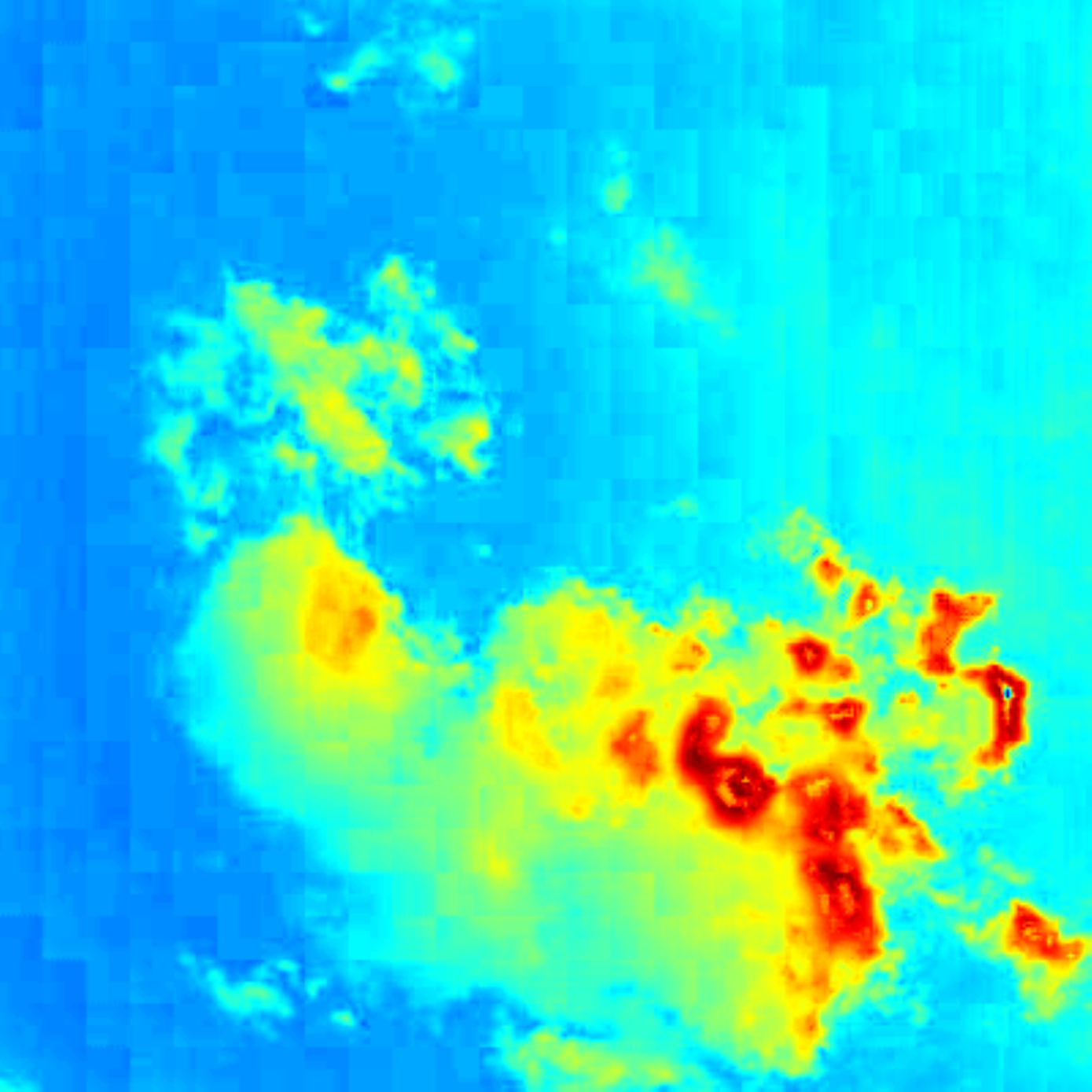} &
\includegraphics[height=0.08\textheight]{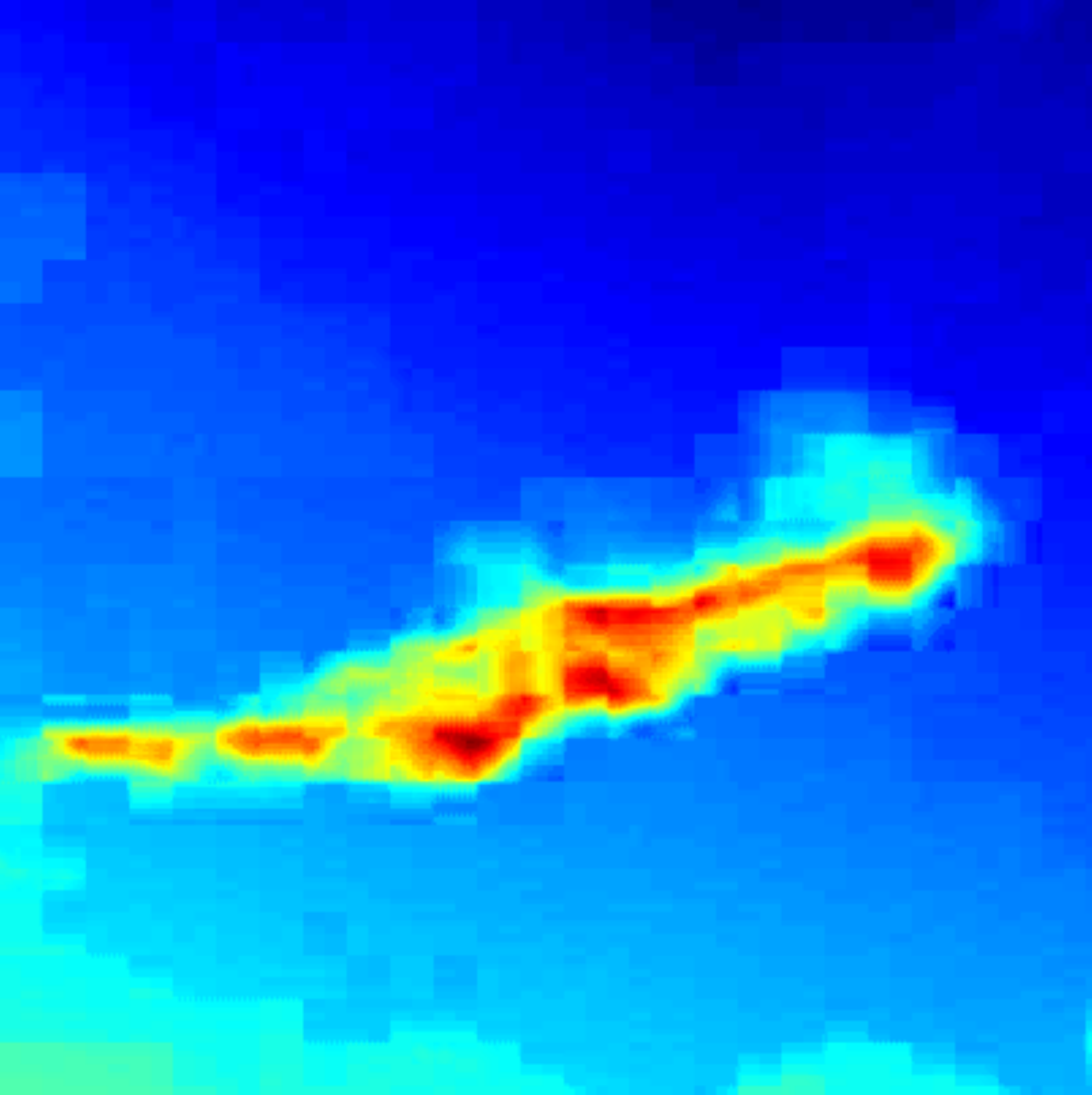} &
\includegraphics[height=0.08\textheight]{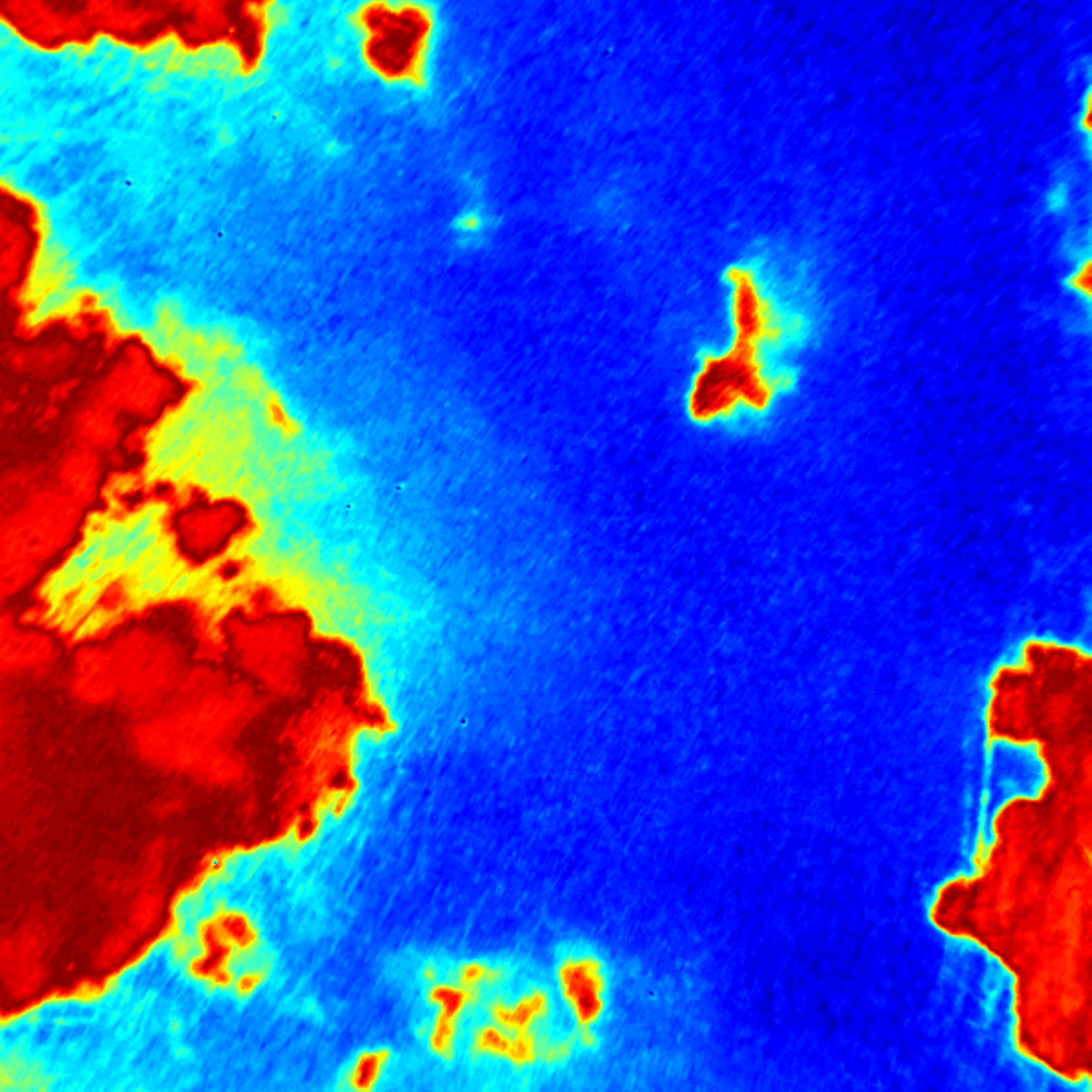} &
\includegraphics[height=0.08\textheight]{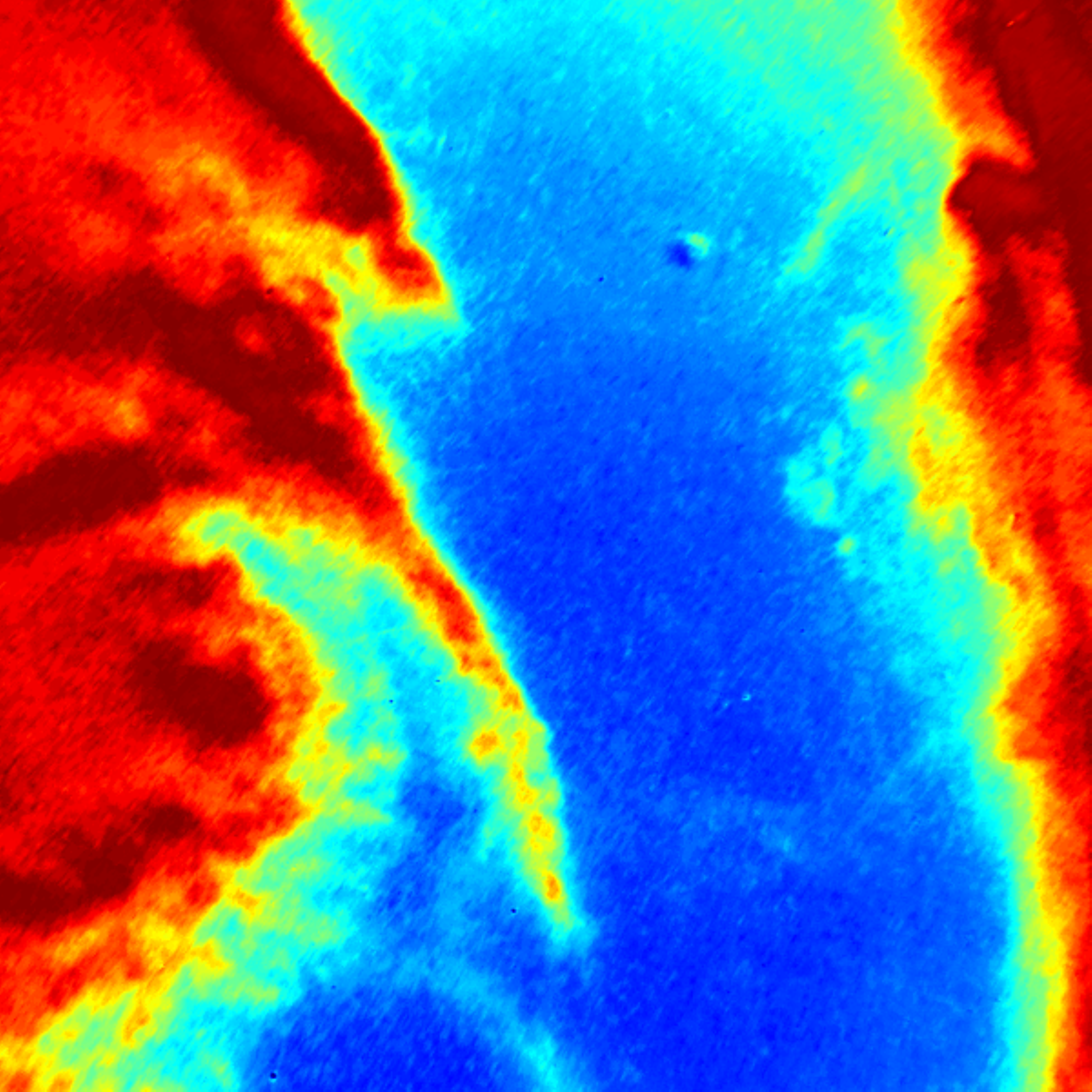} &
\includegraphics[height=0.08\textheight]{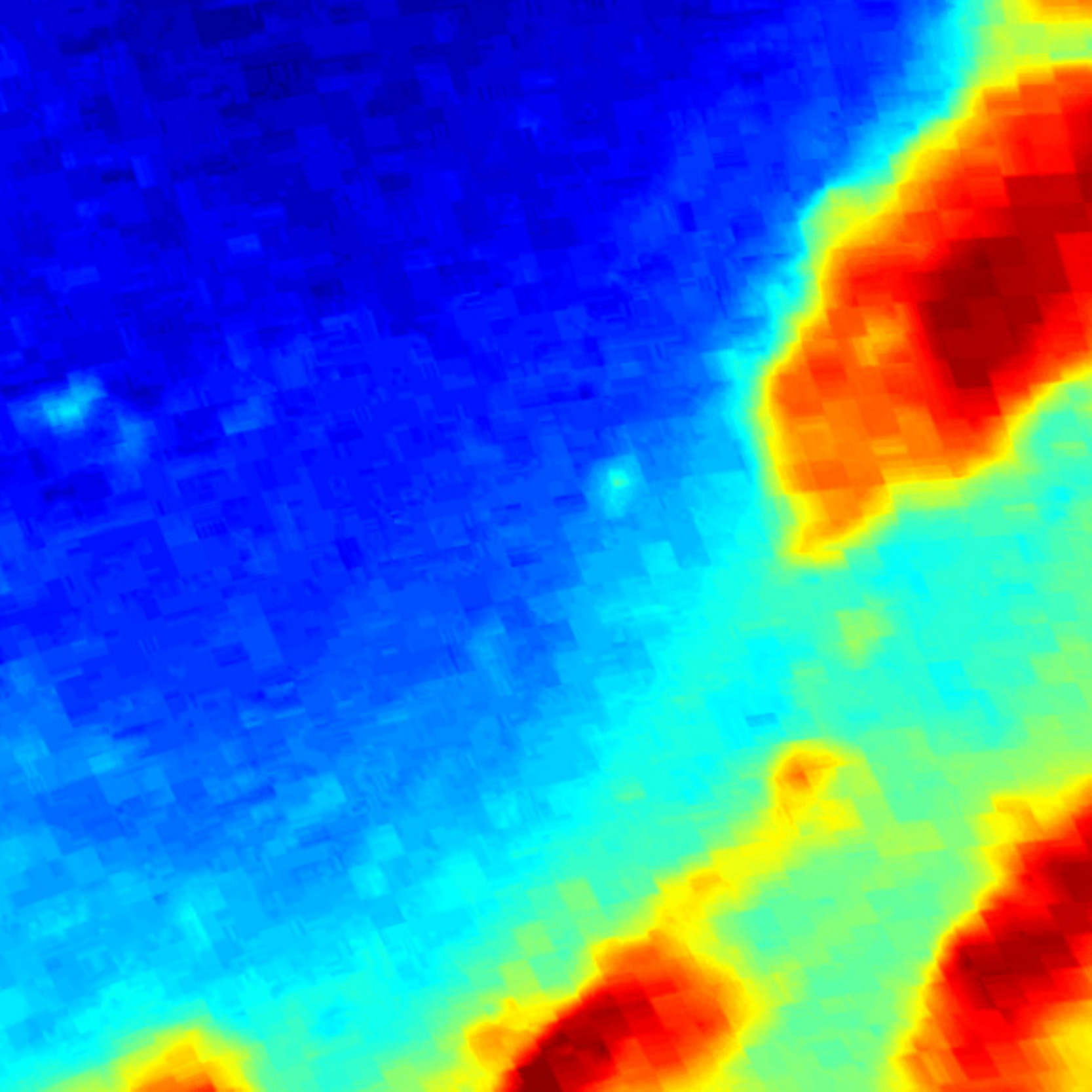} \vspace{-0.39in}\\

{\fontsize{0.4cm}{1em}\selectfont Our \mbox{approach} (binary)} \vspace{1.5cm}& 
\includegraphics[height=0.08\textheight]{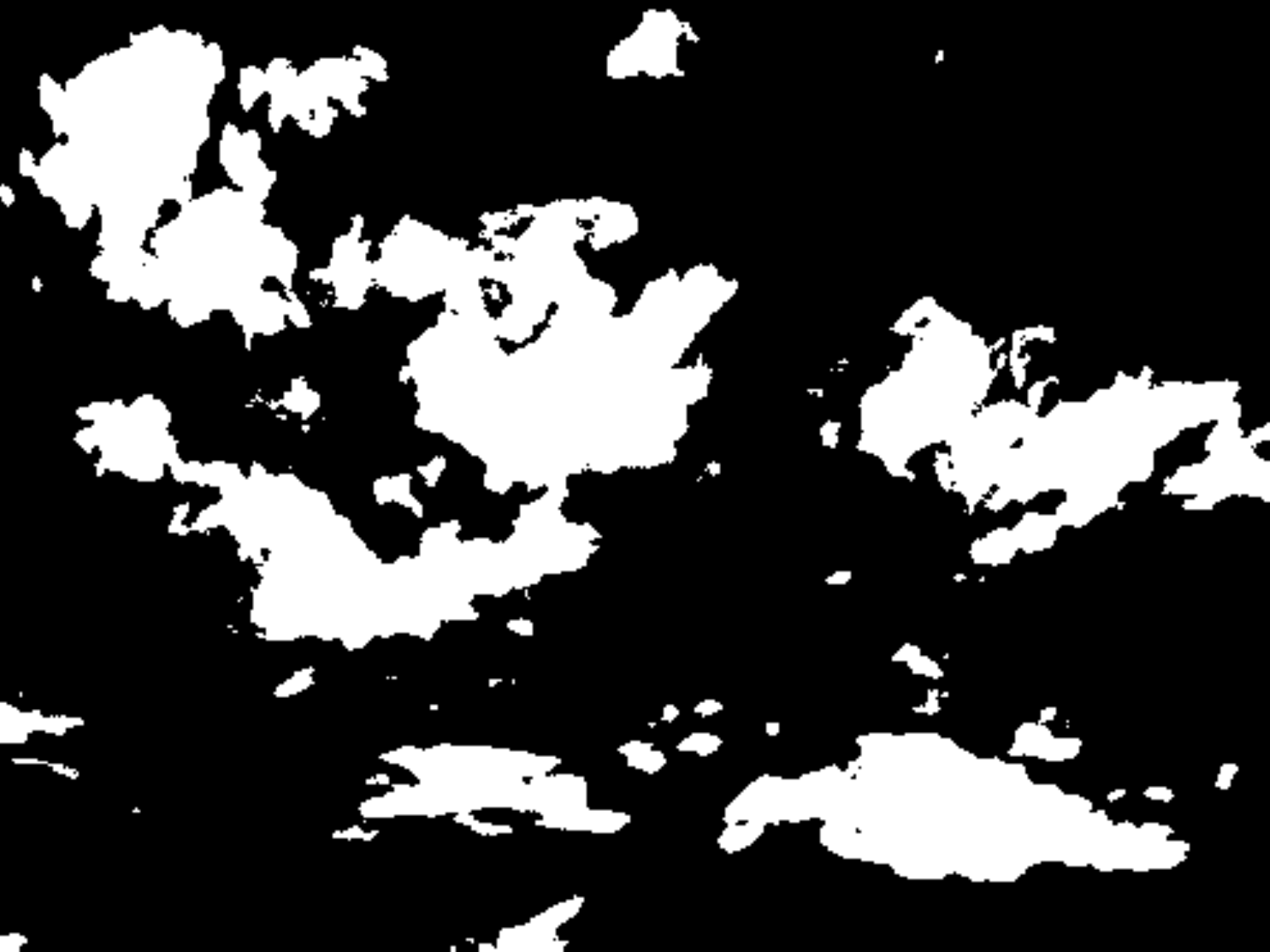} &
\includegraphics[height=0.08\textheight]{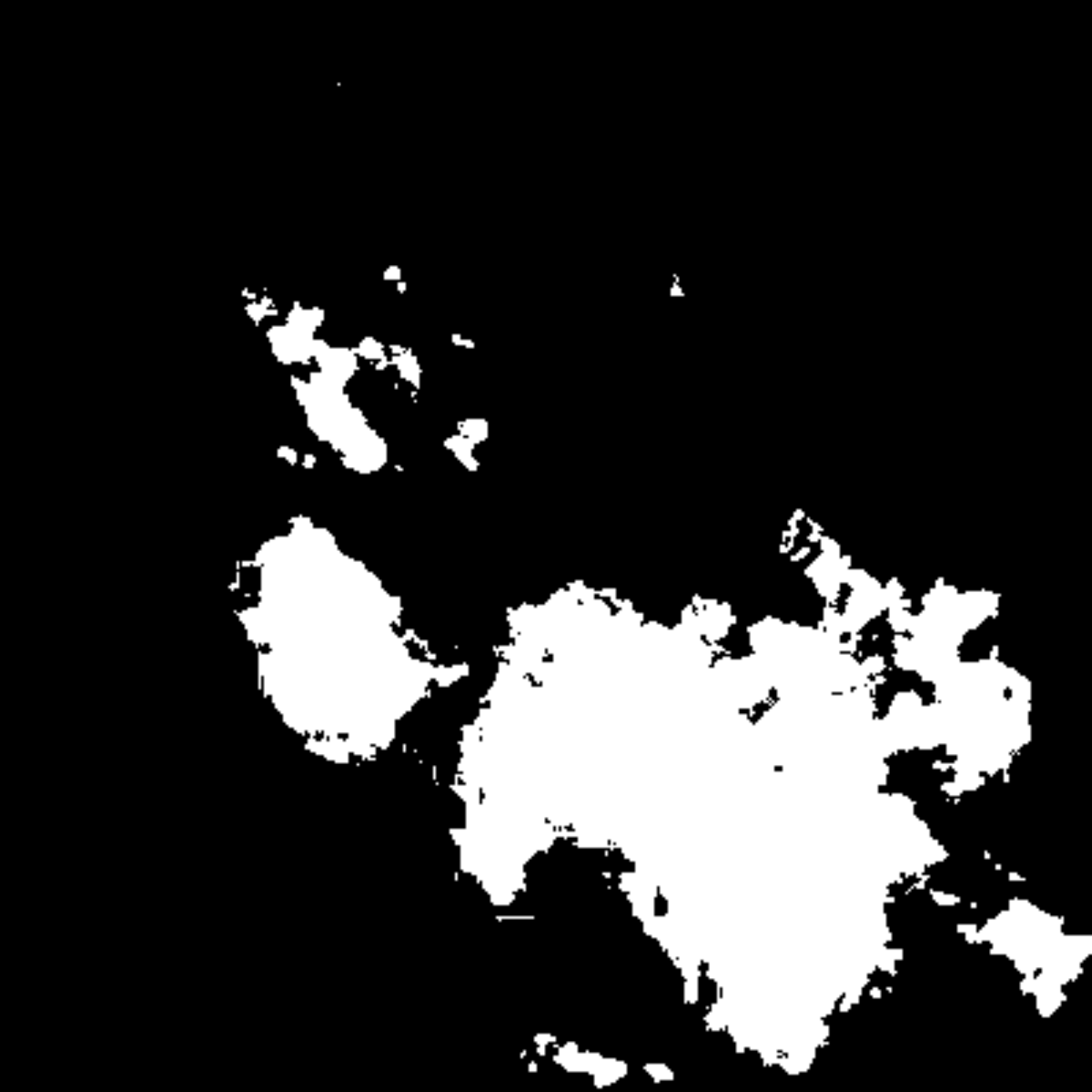} &
\includegraphics[height=0.08\textheight]{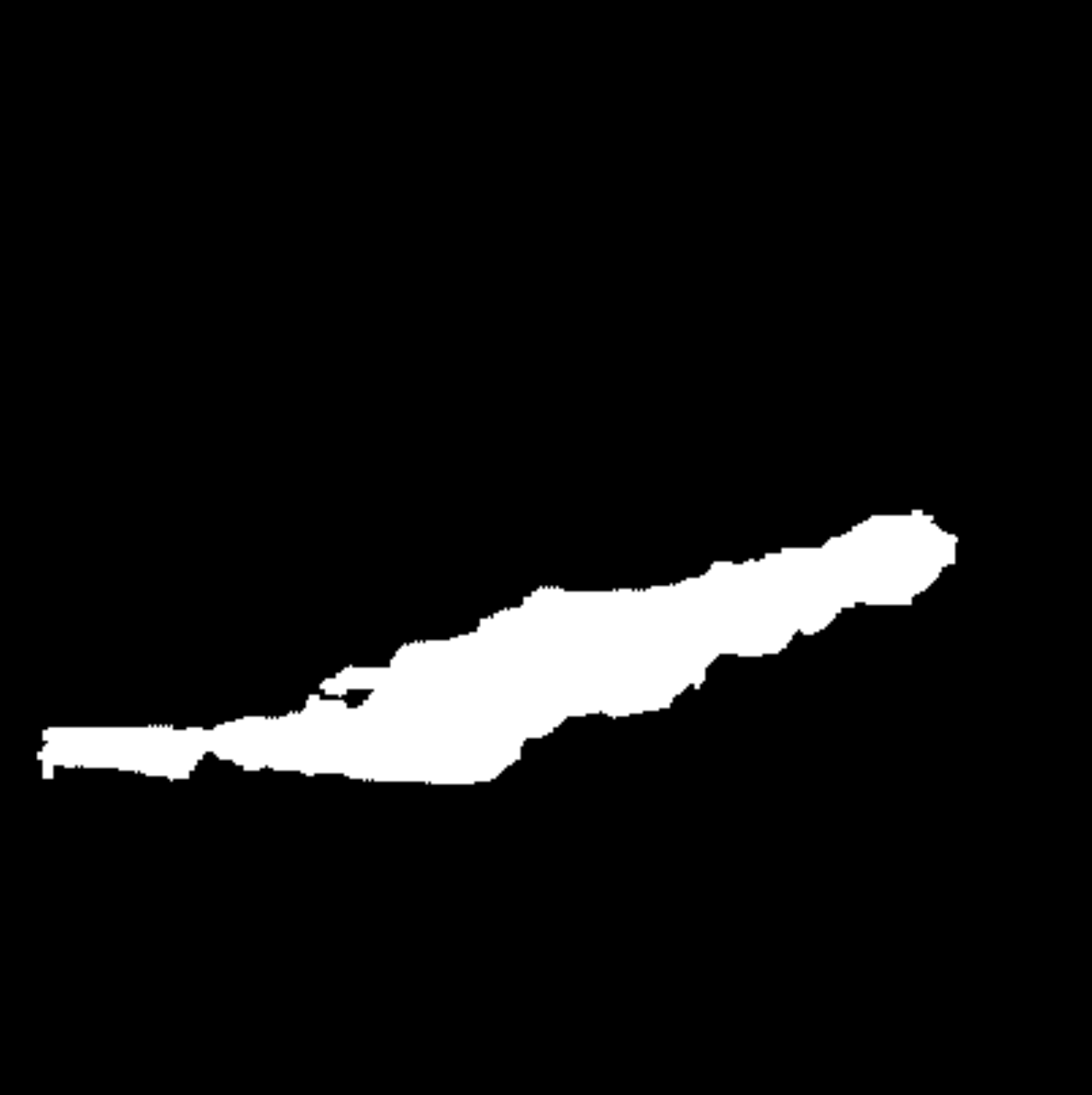} &
\fbox{\includegraphics[height=0.08\textheight]{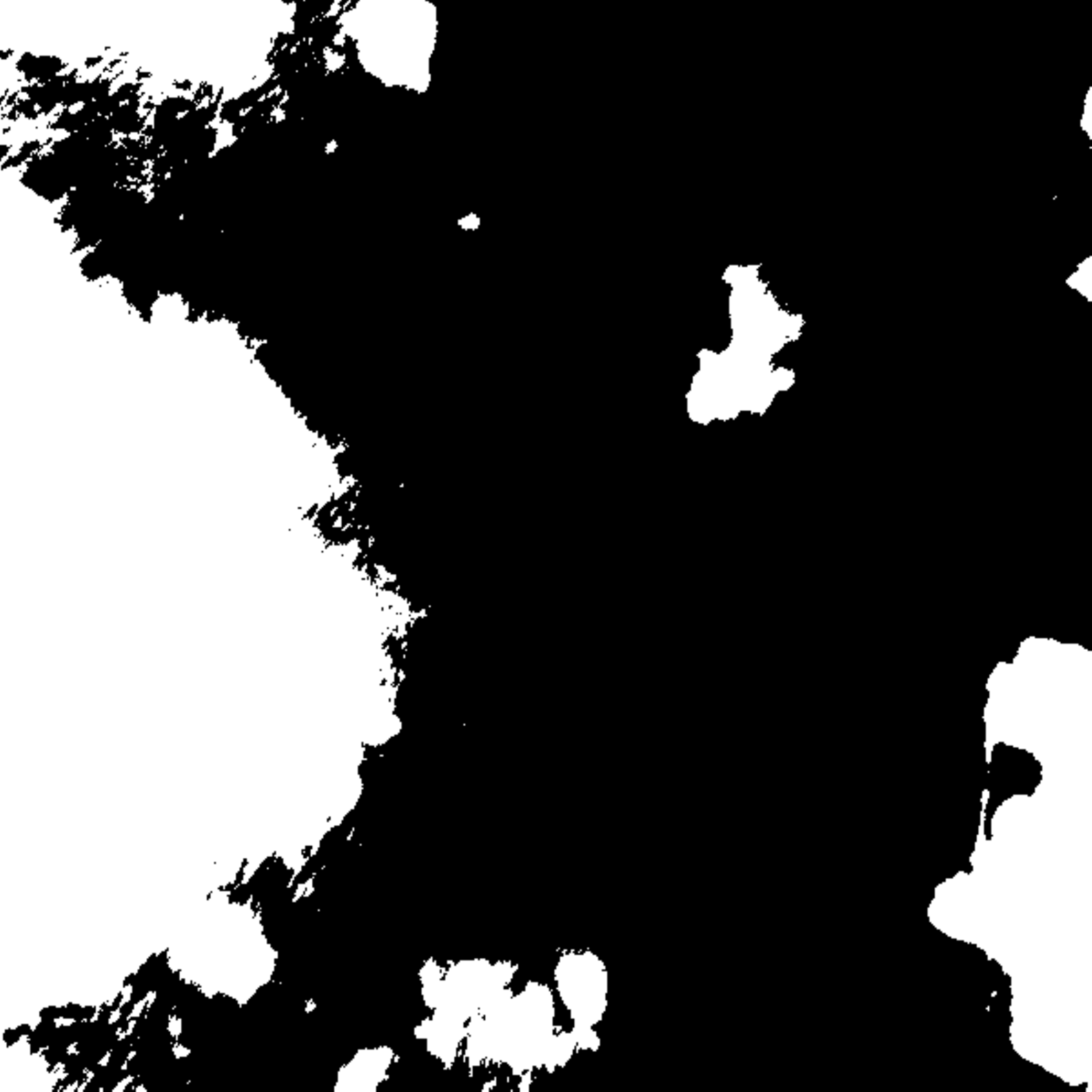}} &
\fbox{\includegraphics[height=0.08\textheight]{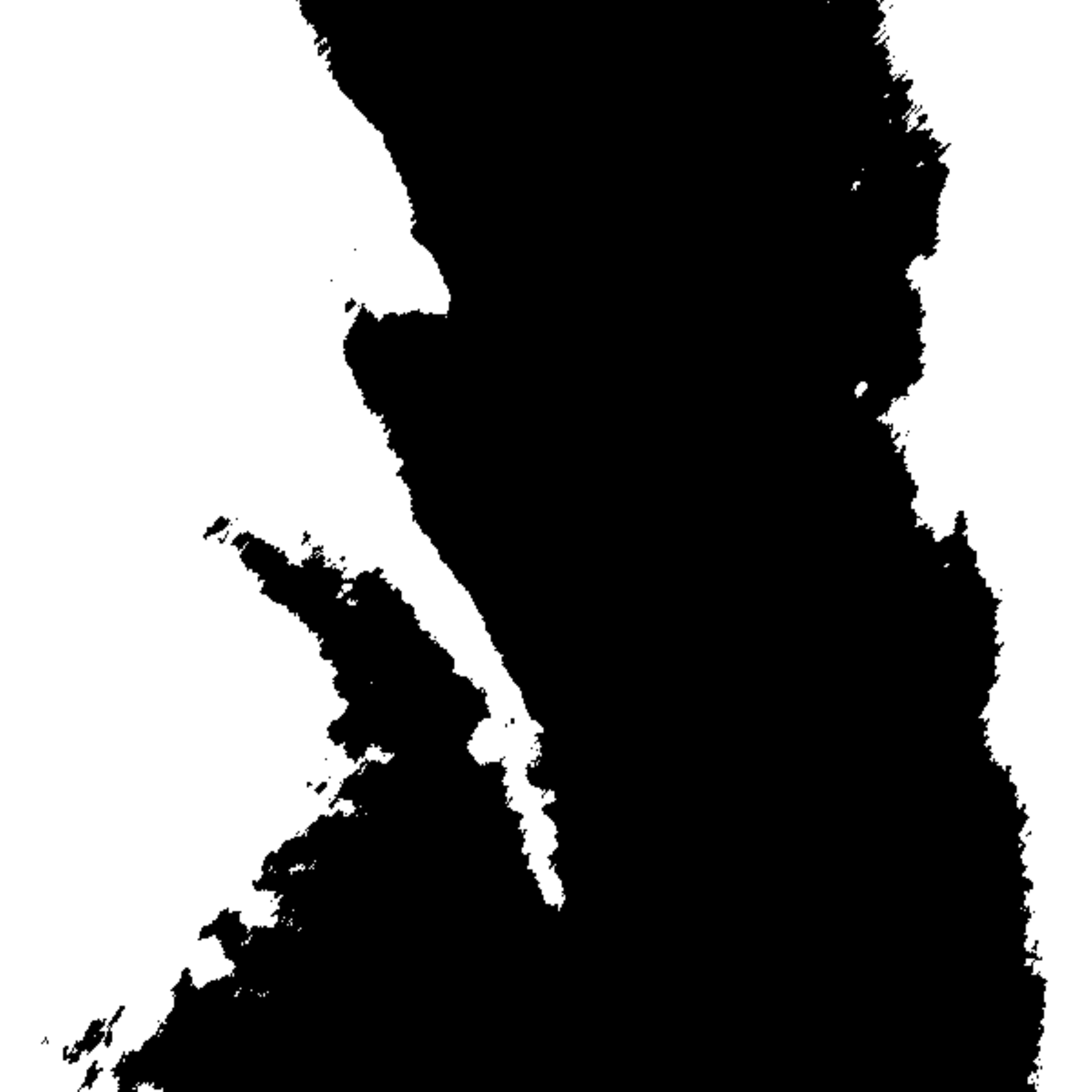}} &
\fbox{\includegraphics[height=0.08\textheight]{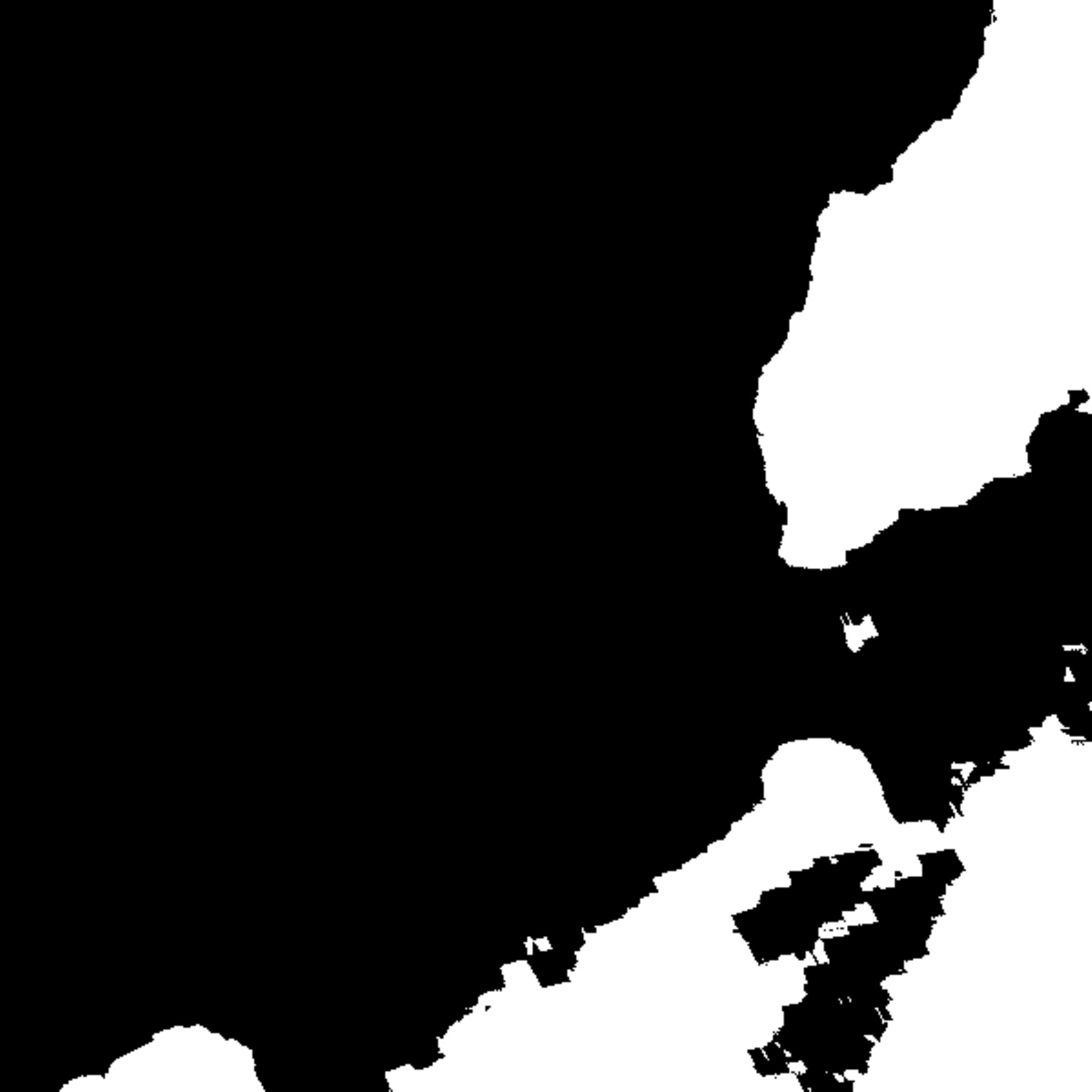}} \vspace{-0.39in}\\

{\fontsize{0.4cm}{1em}\selectfont Li et al.} \vspace{1.9cm}& 
\includegraphics[height=0.08\textheight]{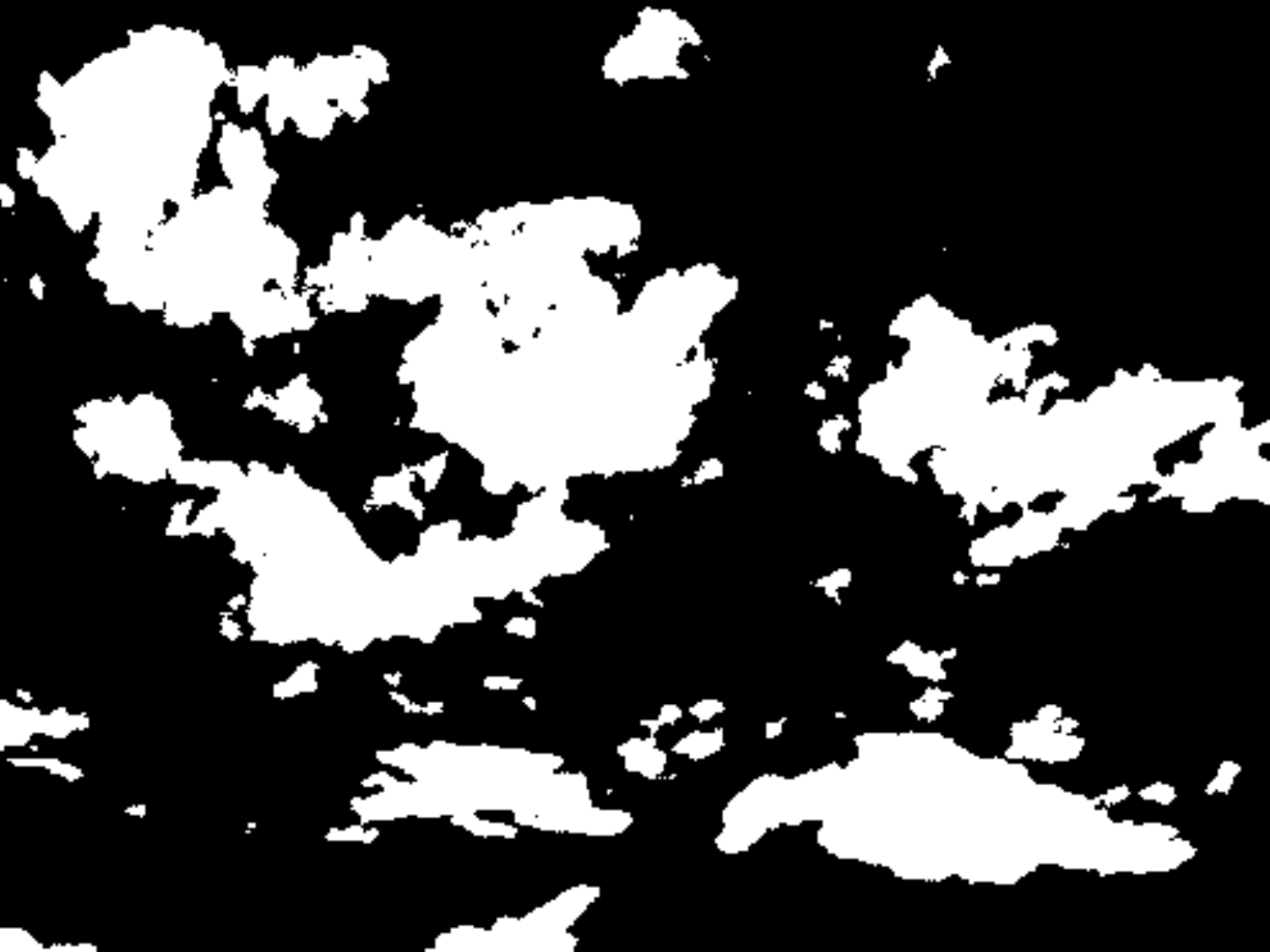} &
\includegraphics[height=0.08\textheight]{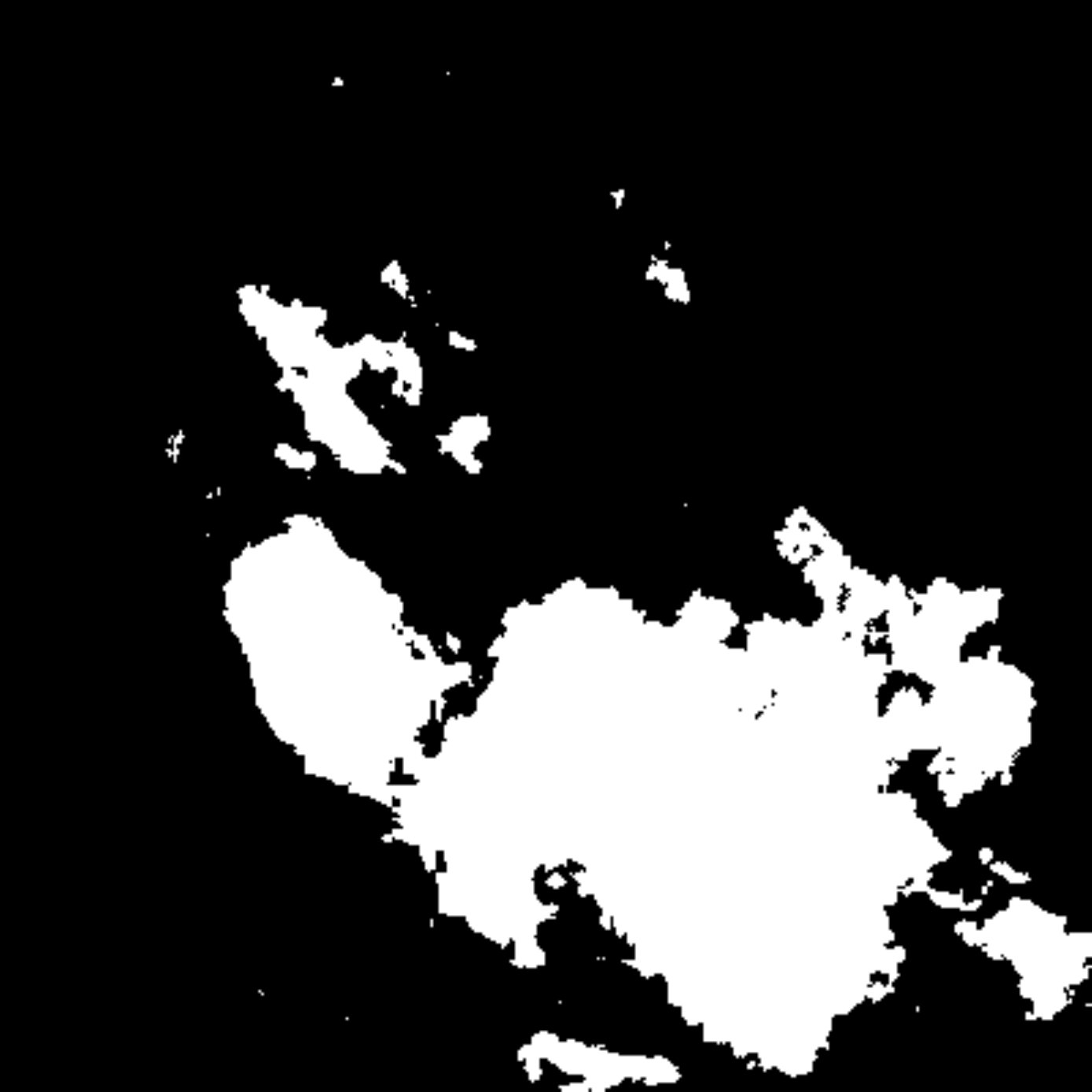} &
\includegraphics[height=0.08\textheight]{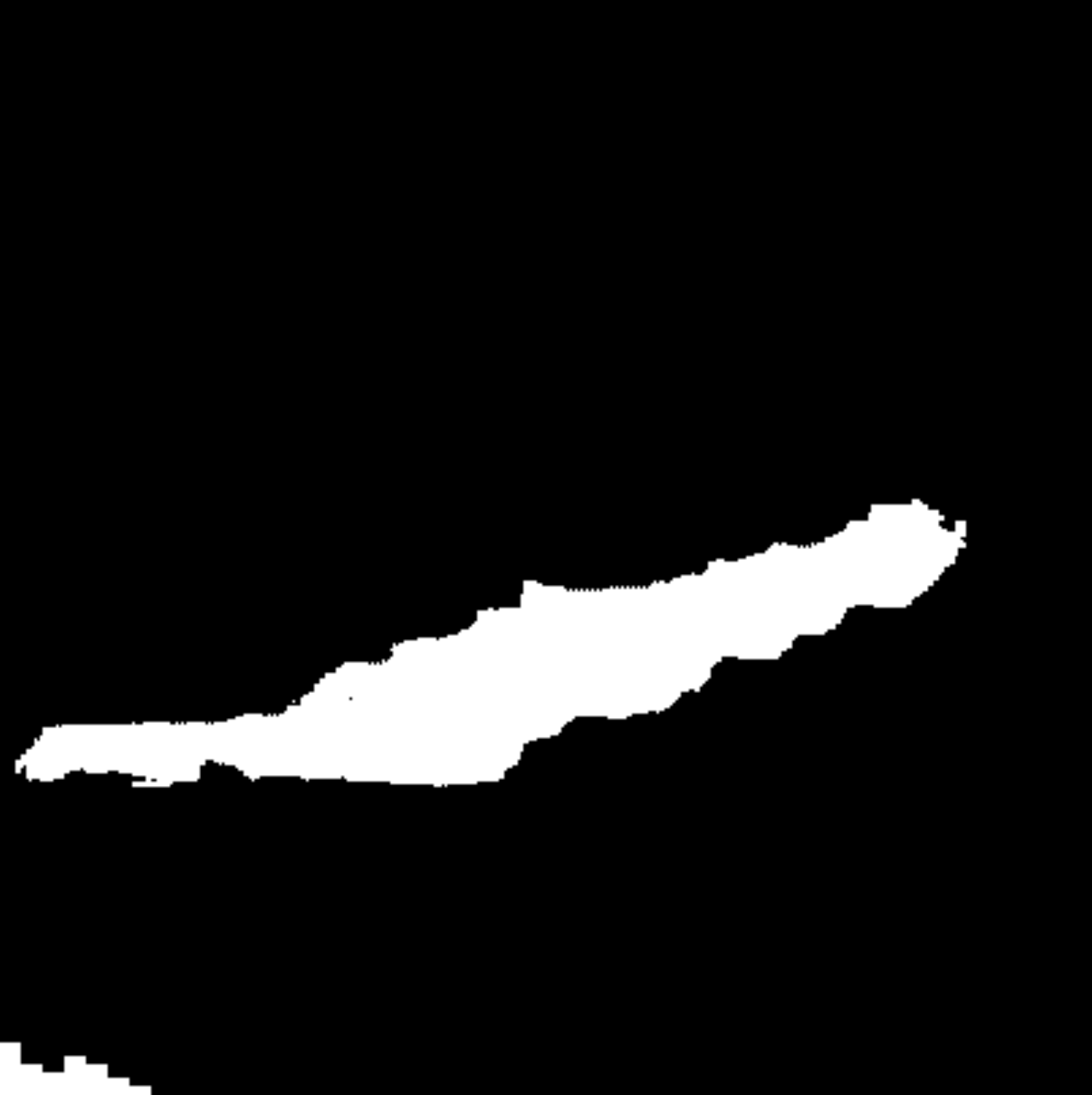} &
\fbox{\includegraphics[height=0.08\textheight]{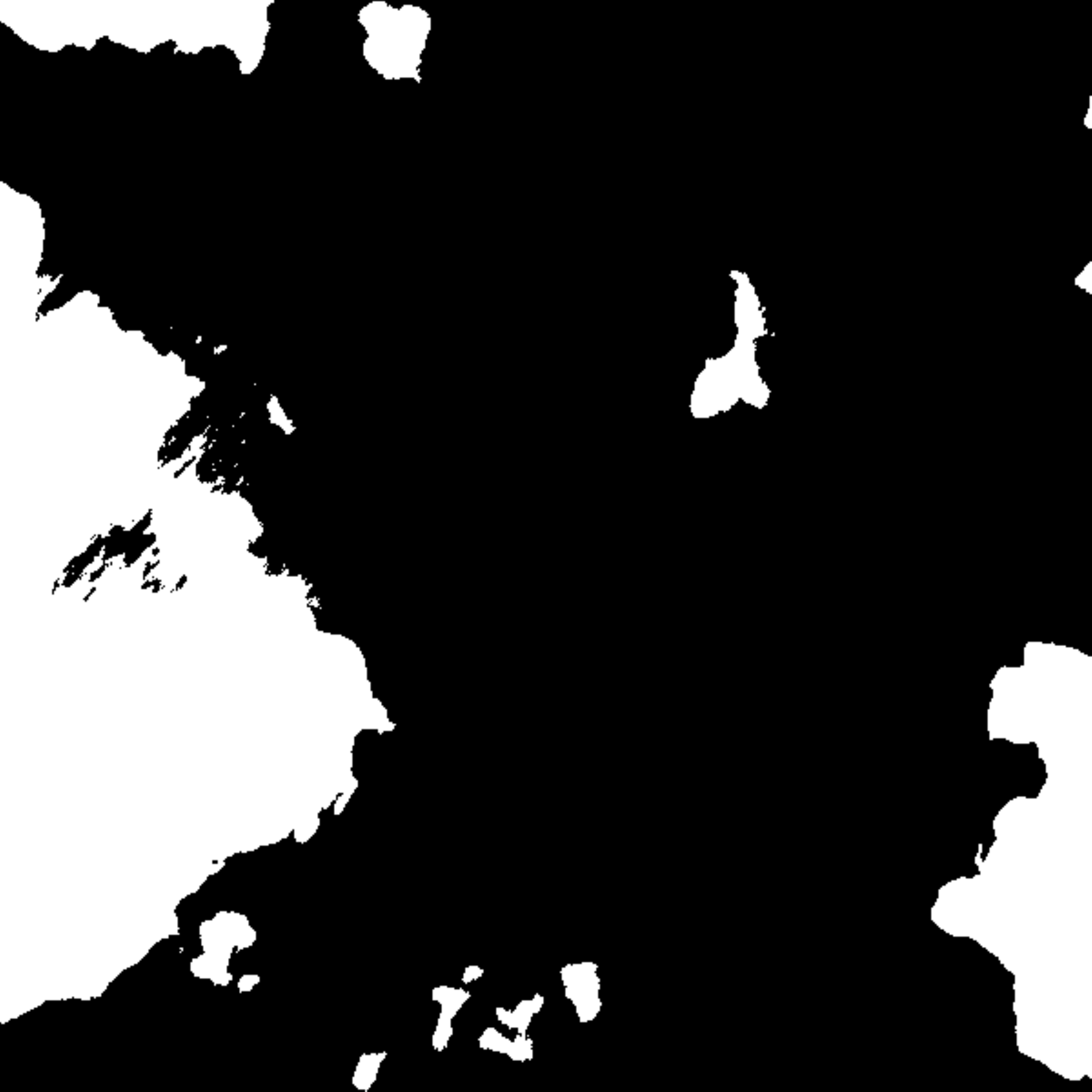}} &
\fbox{\includegraphics[height=0.08\textheight]{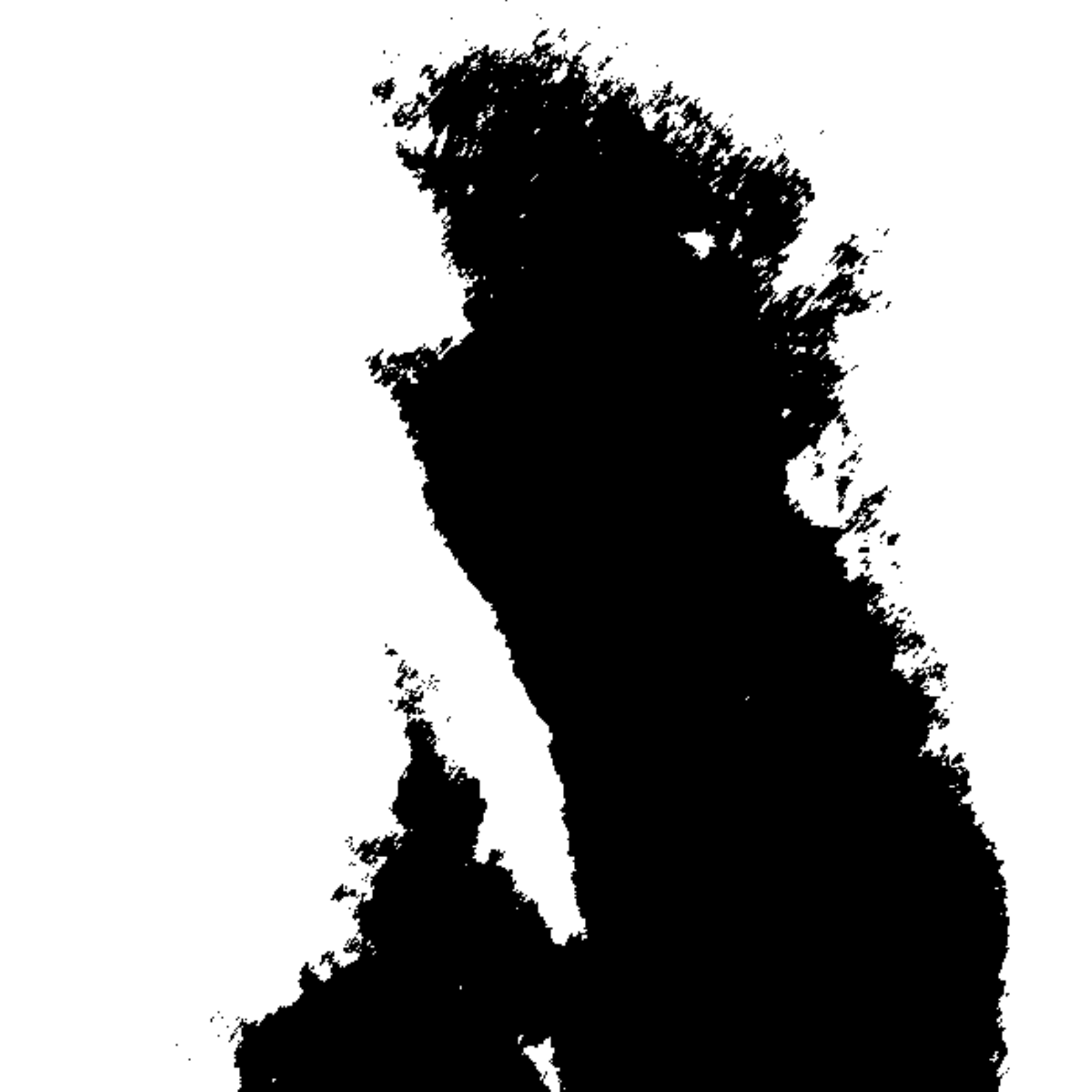}} &
\fbox{\includegraphics[height=0.08\textheight]{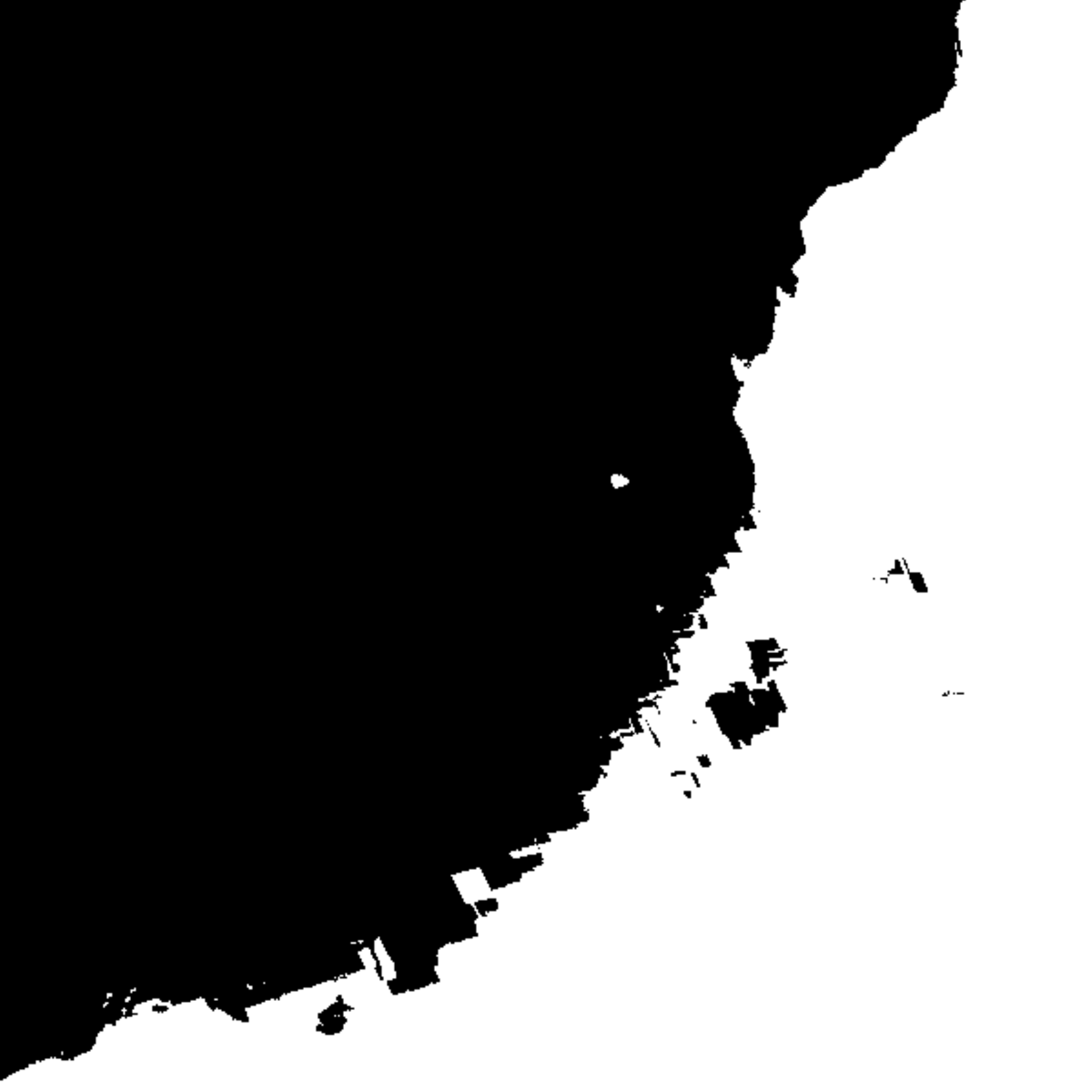}} \vspace{-0.35in}\\

{\fontsize{0.4cm}{1em}\selectfont Souza et al.} \vspace{1.5cm}& 
\includegraphics[height=0.08\textheight]{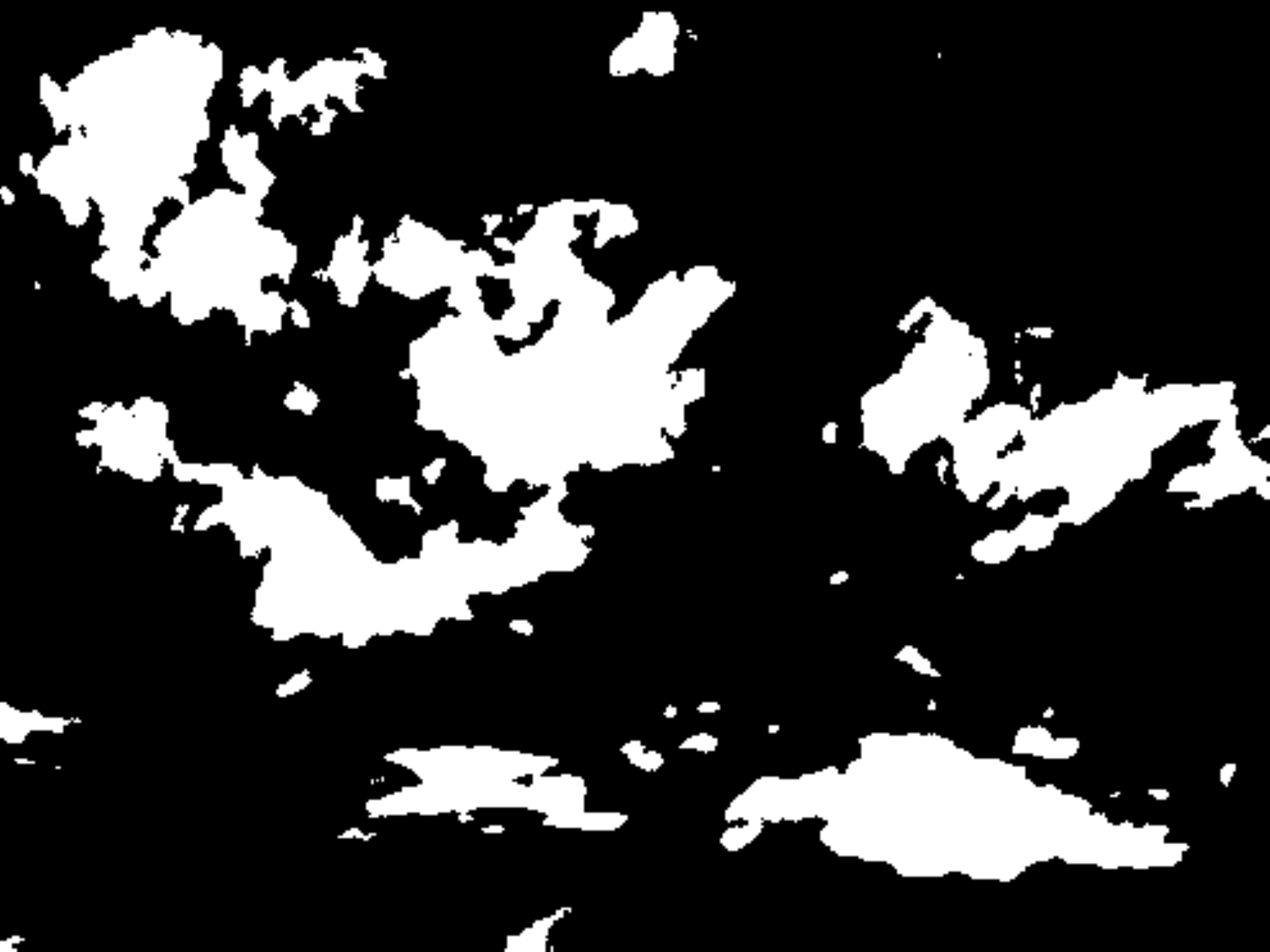} &
\includegraphics[height=0.08\textheight]{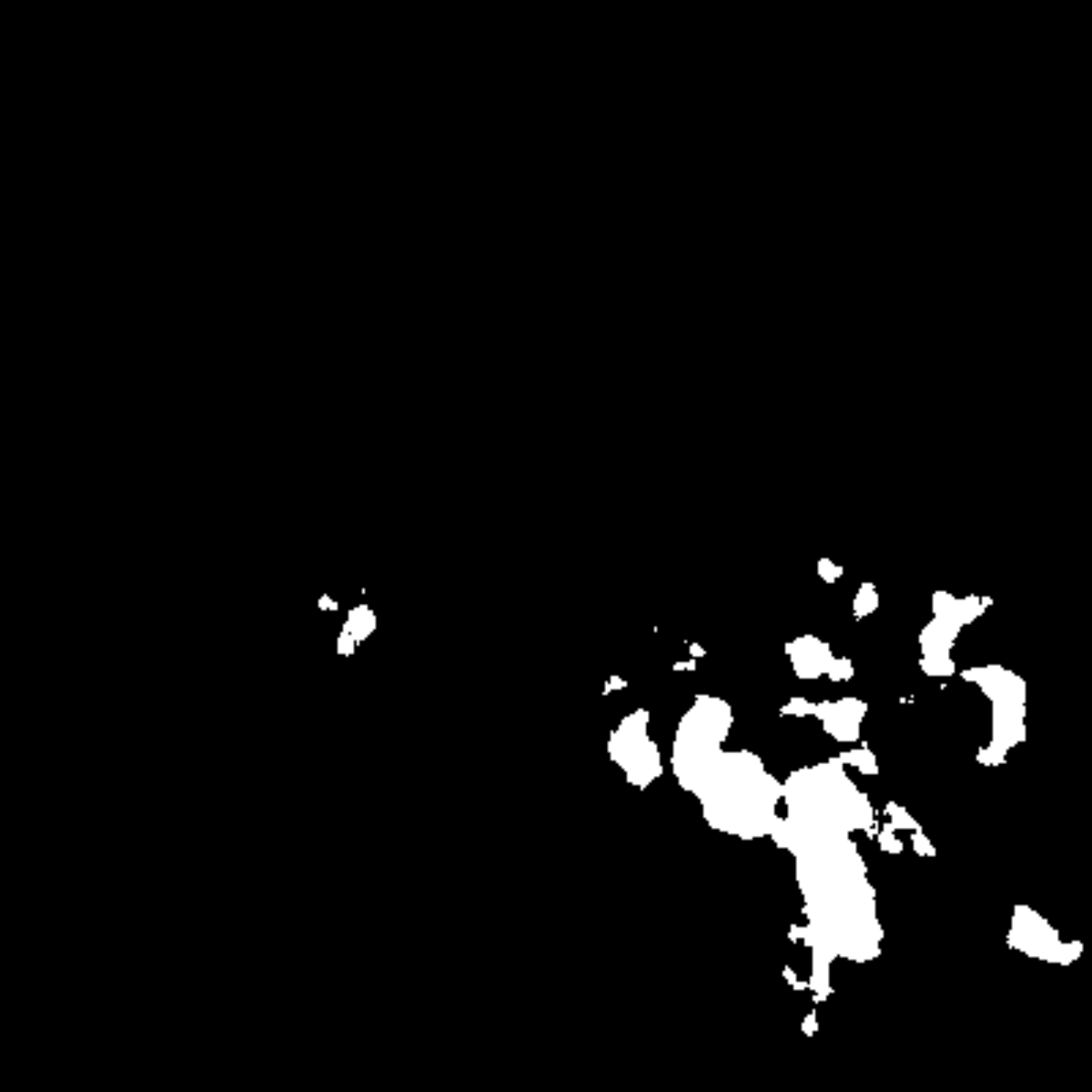} &
\includegraphics[height=0.08\textheight]{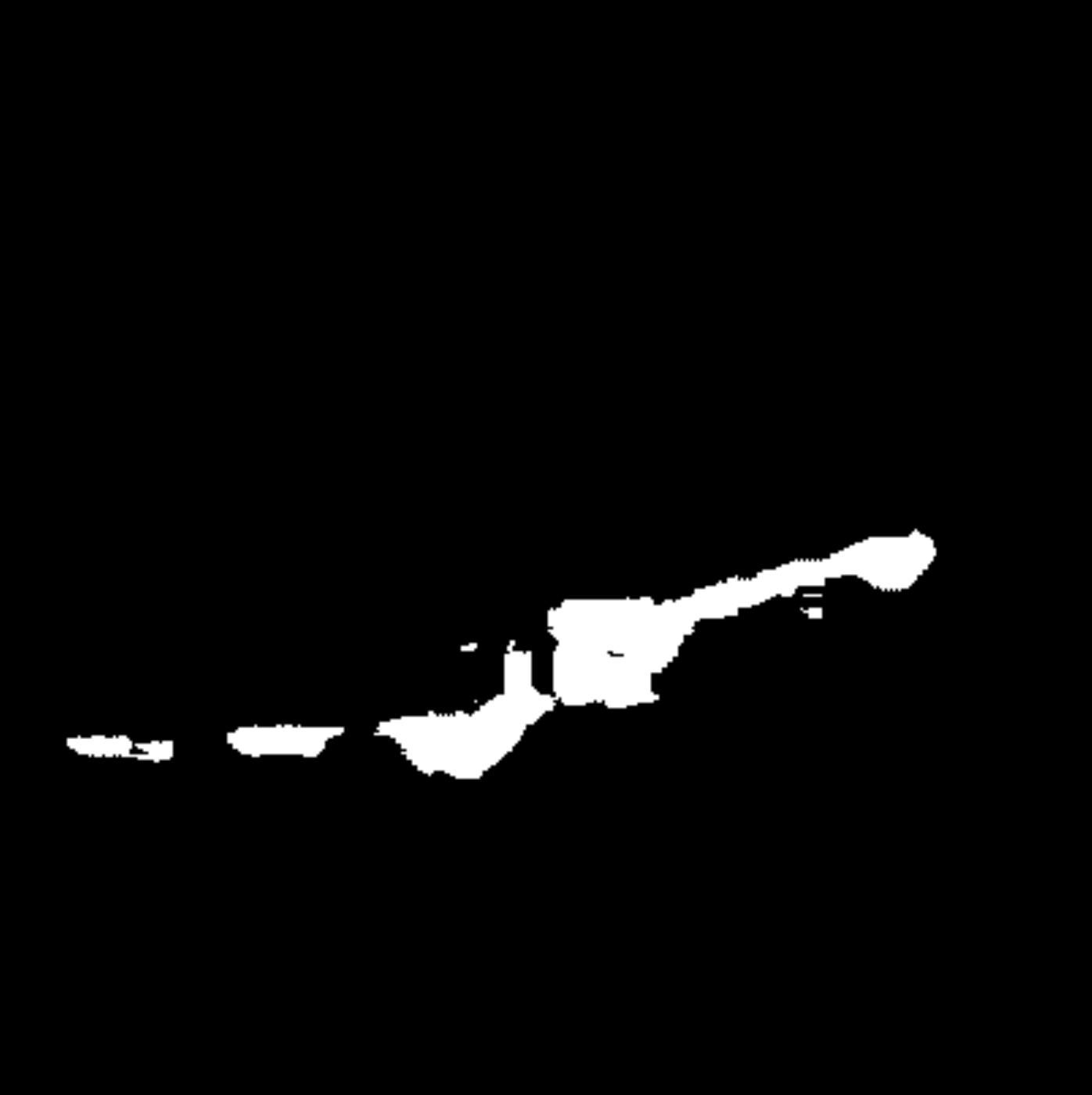} &
\fbox{\includegraphics[height=0.08\textheight]{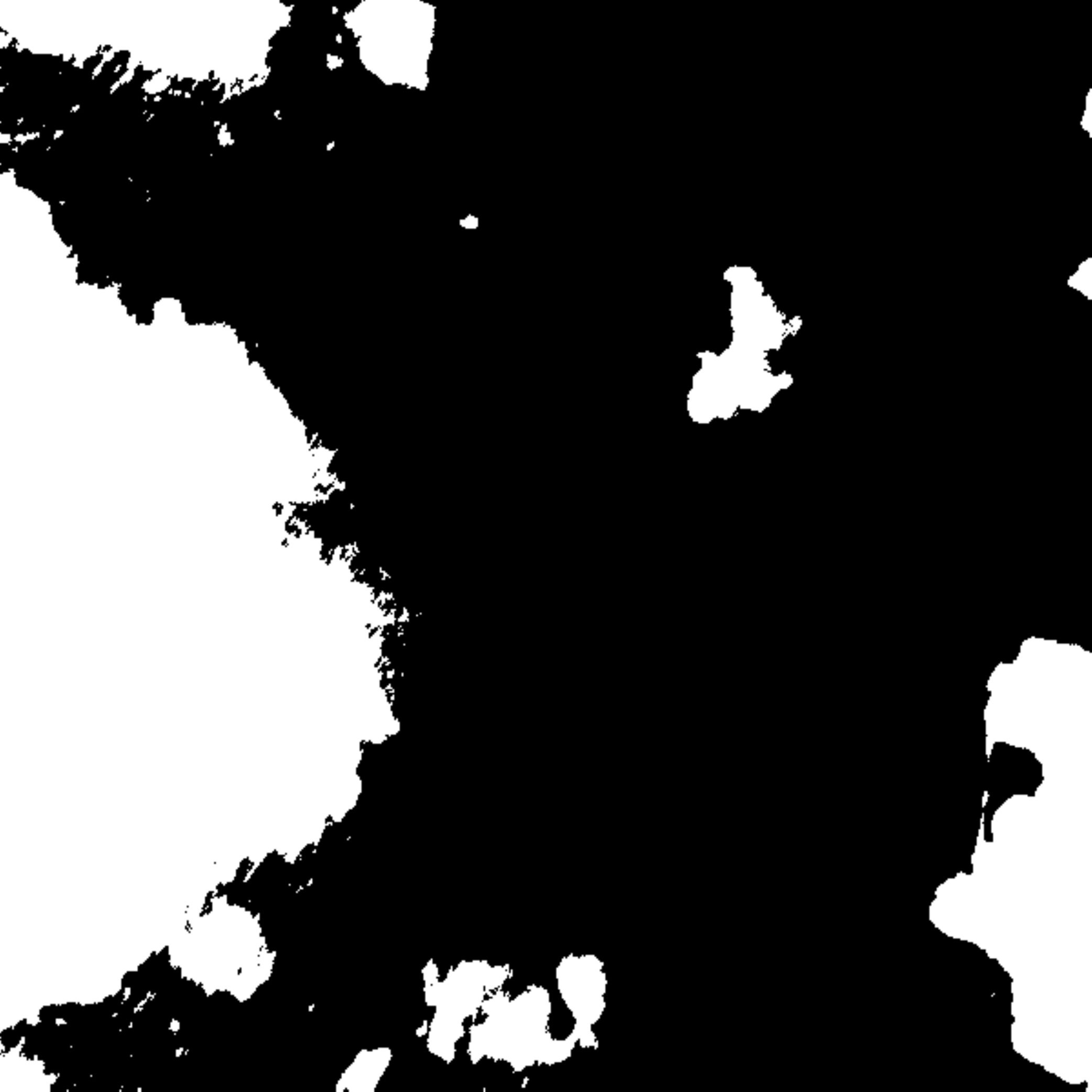}} &
\fbox{\includegraphics[height=0.08\textheight]{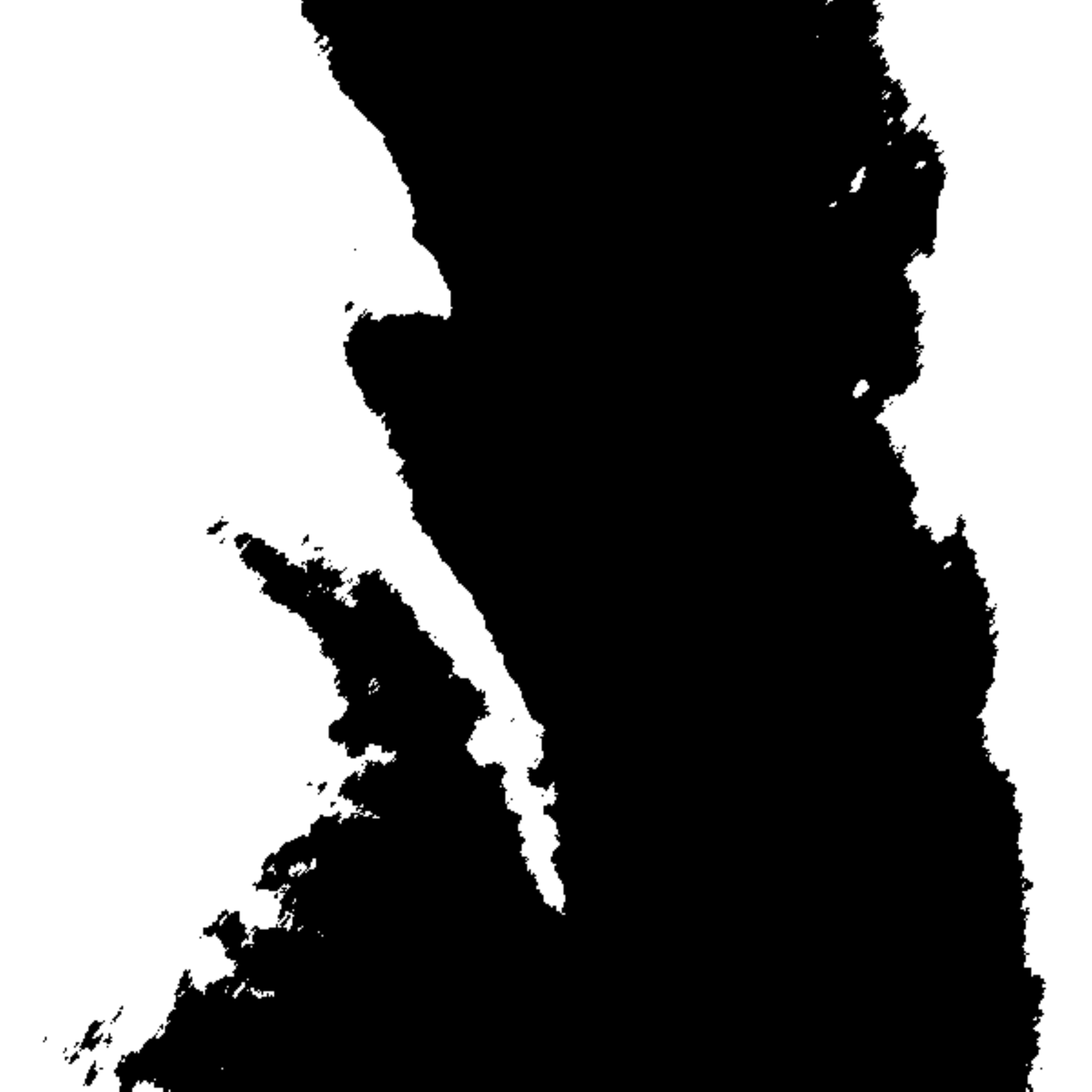}} &
\fbox{\includegraphics[height=0.08\textheight]{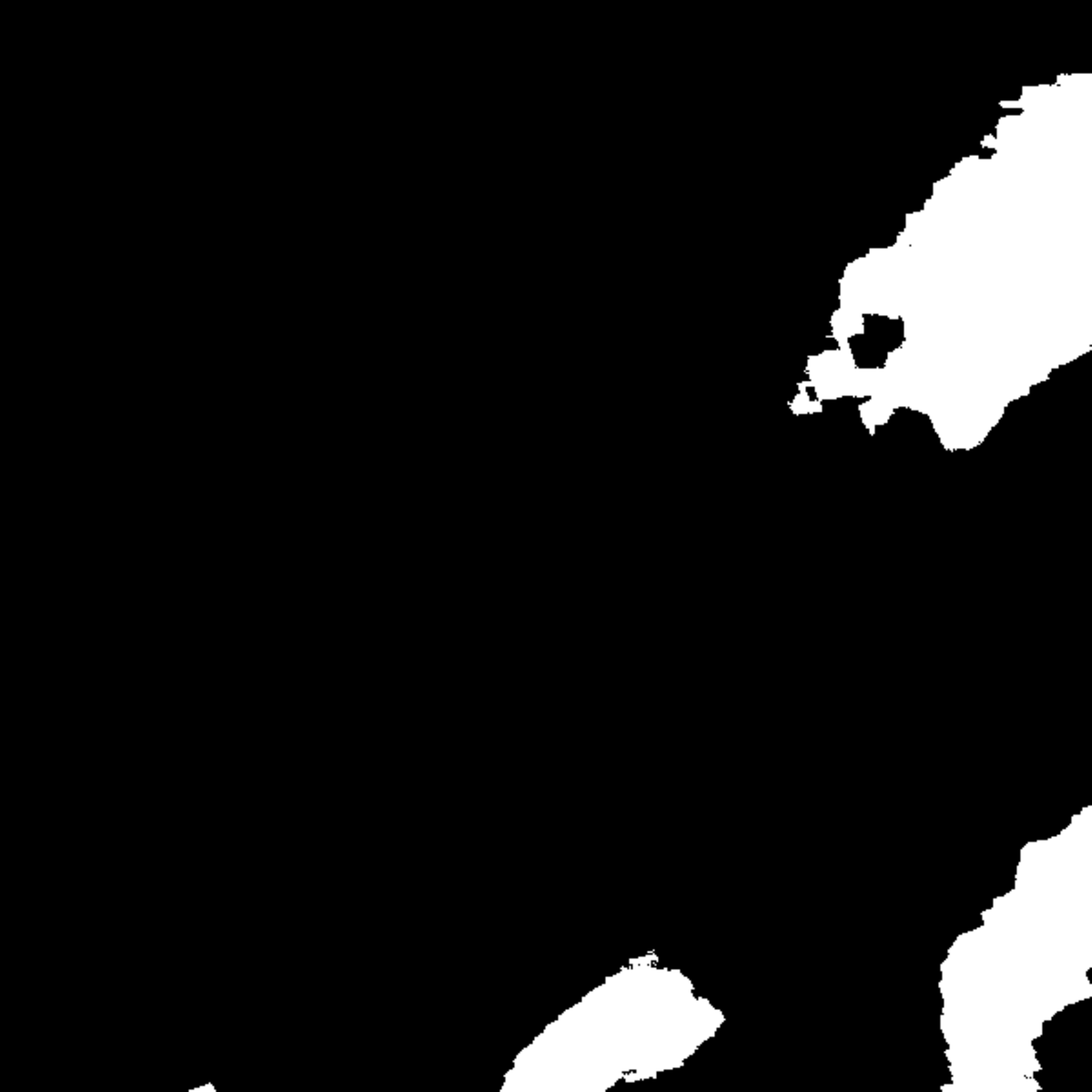}} \vspace{-0.37in}\\

{\fontsize{0.4cm}{1em}\selectfont Long et al.} \vspace{1.5cm}& 
\fbox{\includegraphics[height=0.08\textheight]{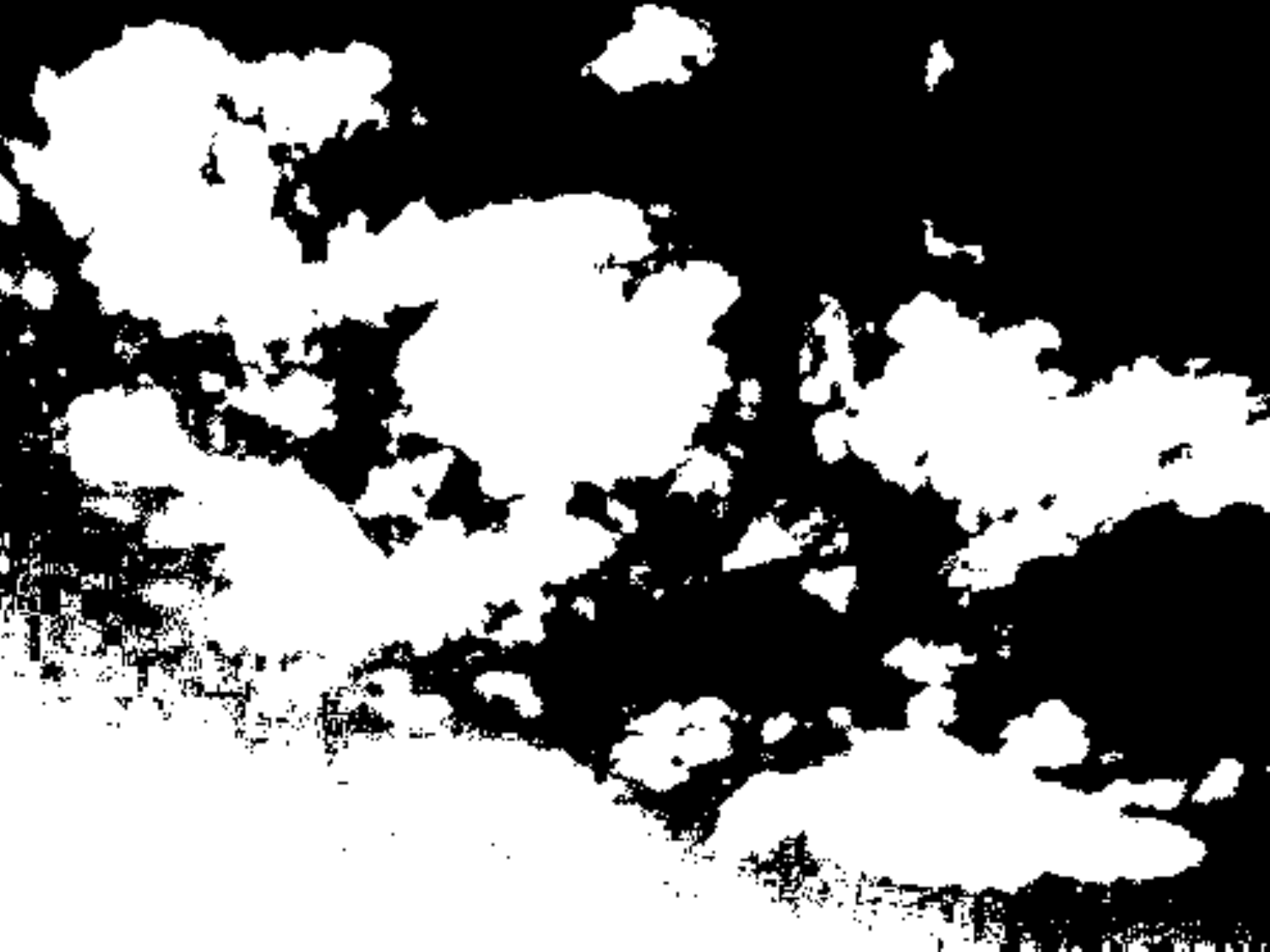}} &
\fbox{\includegraphics[height=0.08\textheight]{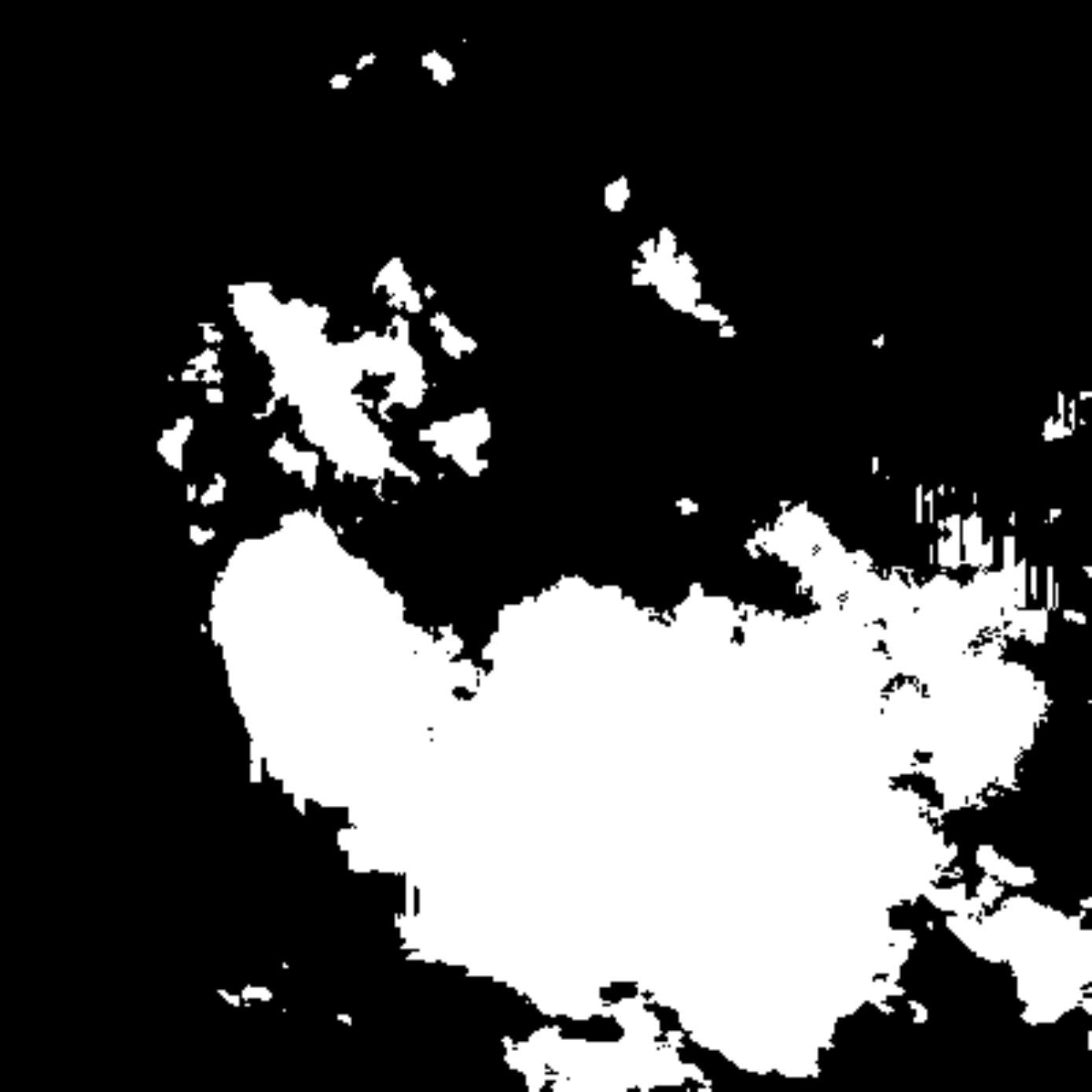}} &
\fbox{\includegraphics[height=0.08\textheight]{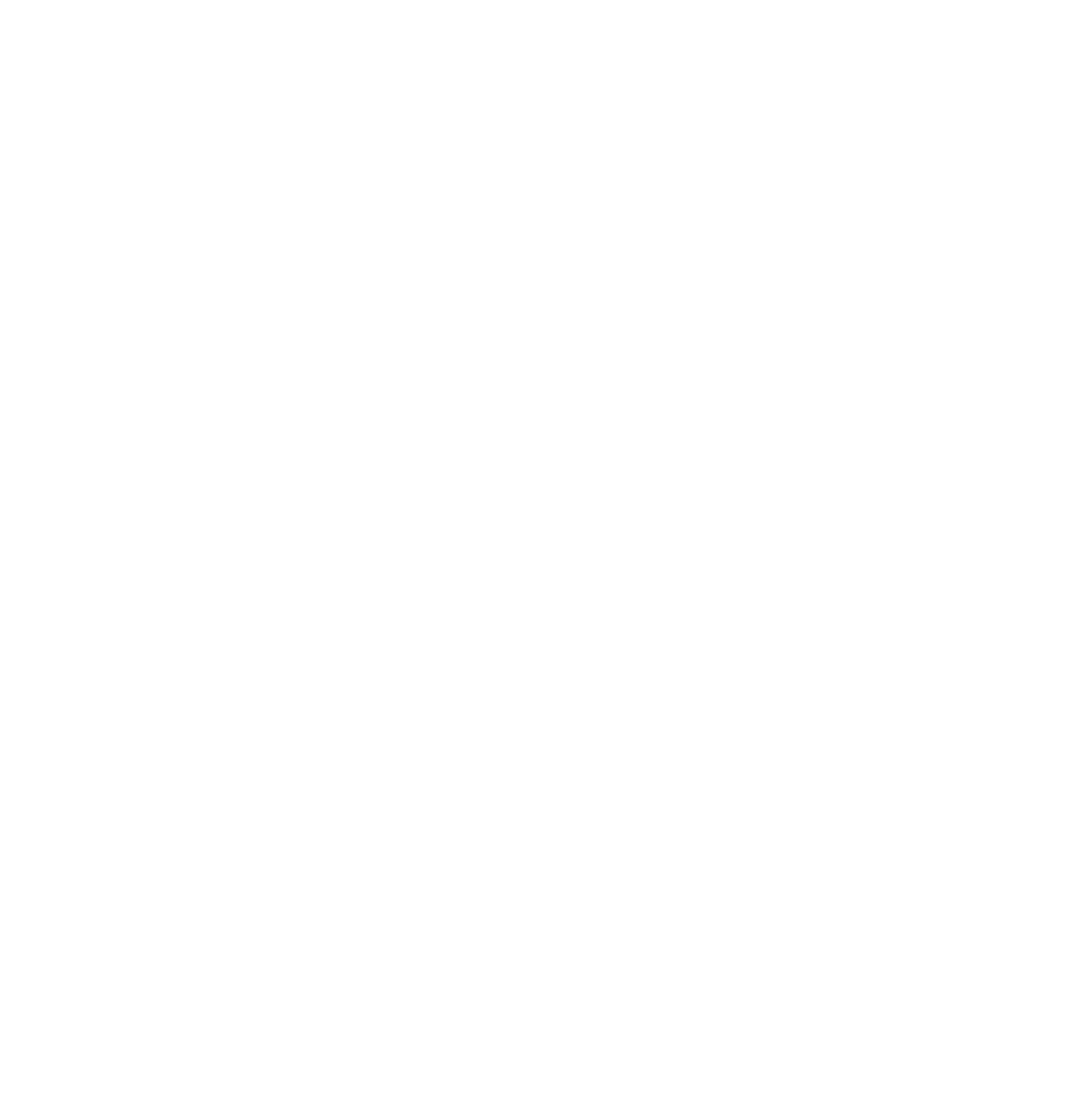}} &
\fbox{\includegraphics[height=0.08\textheight]{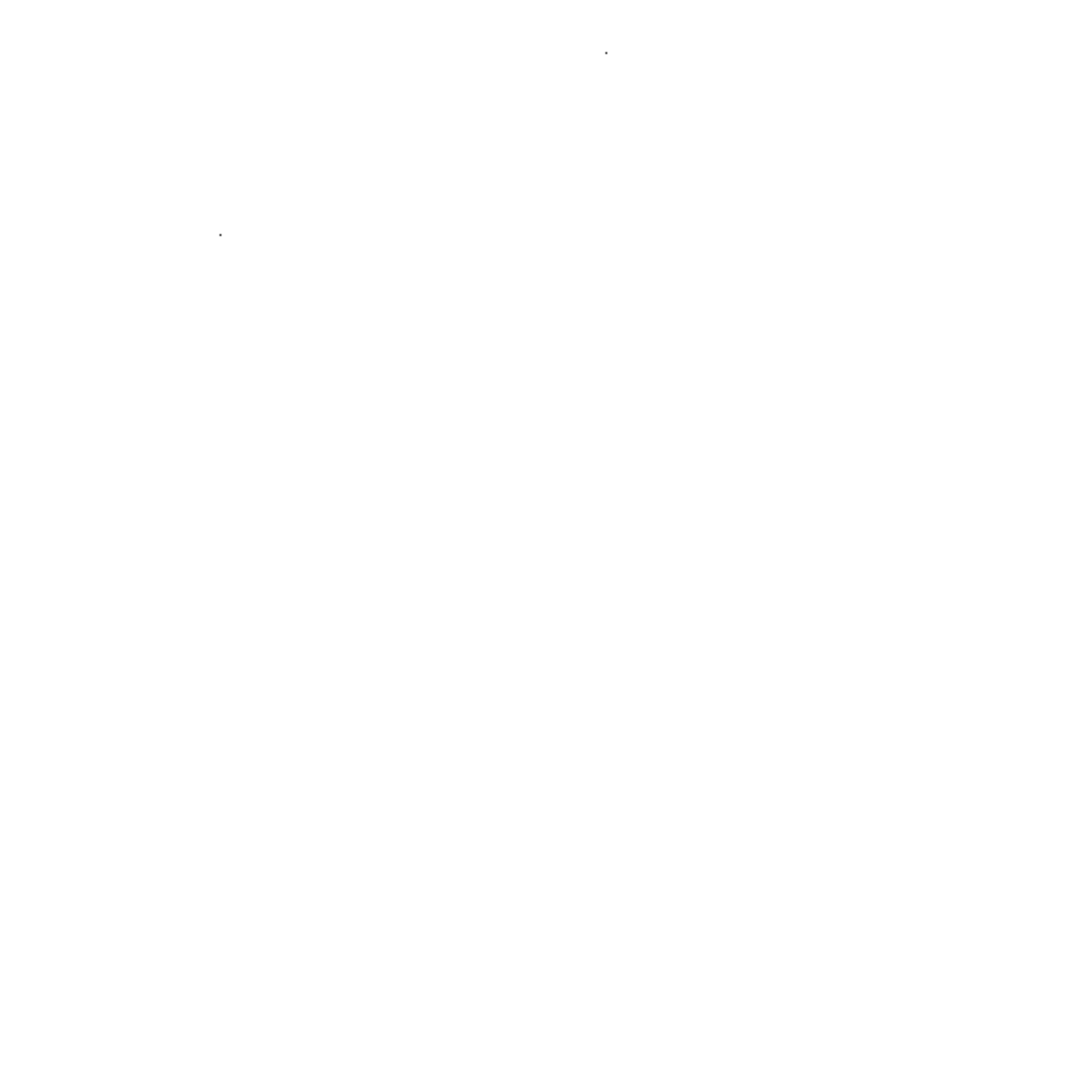}} &
\fbox{\includegraphics[height=0.08\textheight]{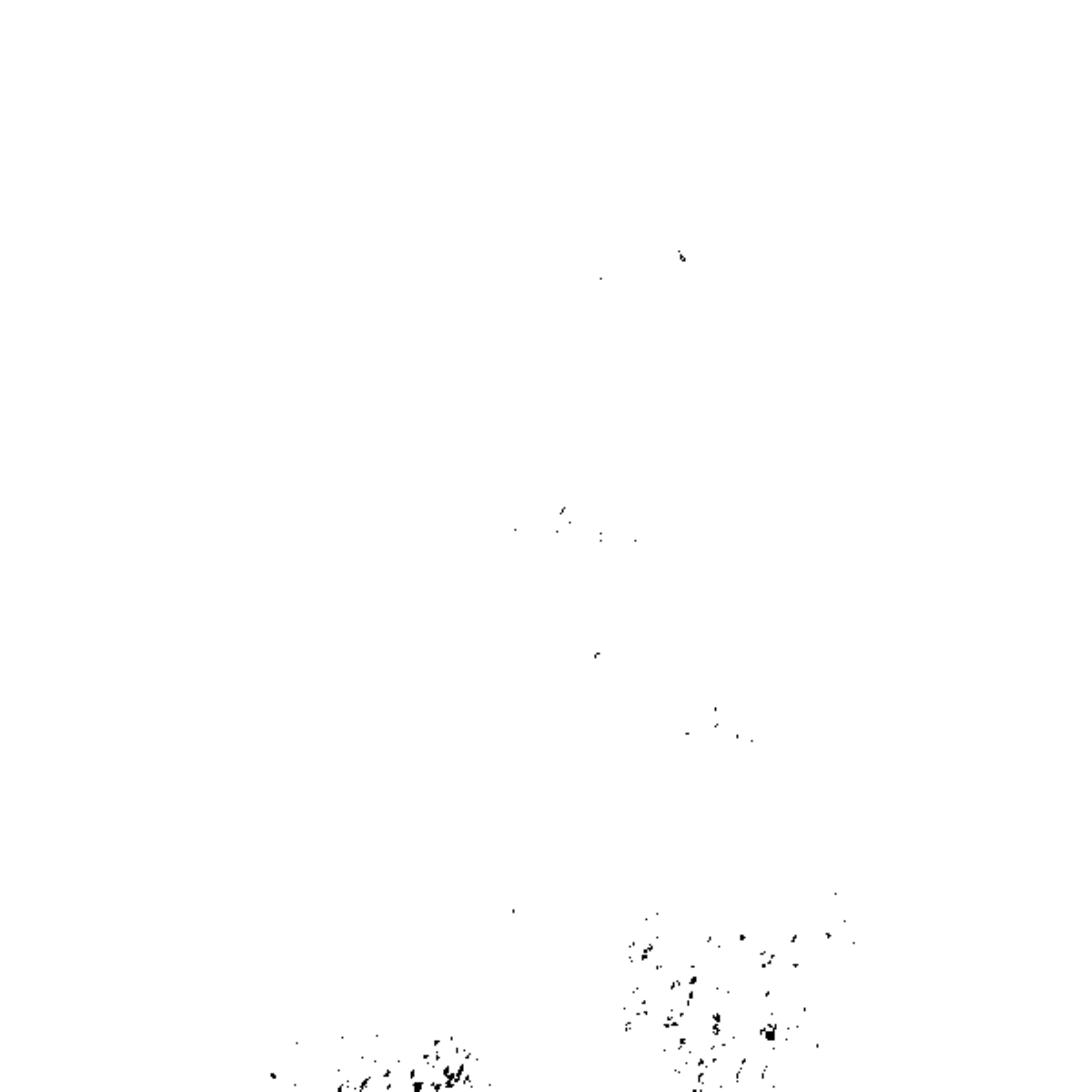}} &
\fbox{\includegraphics[height=0.08\textheight]{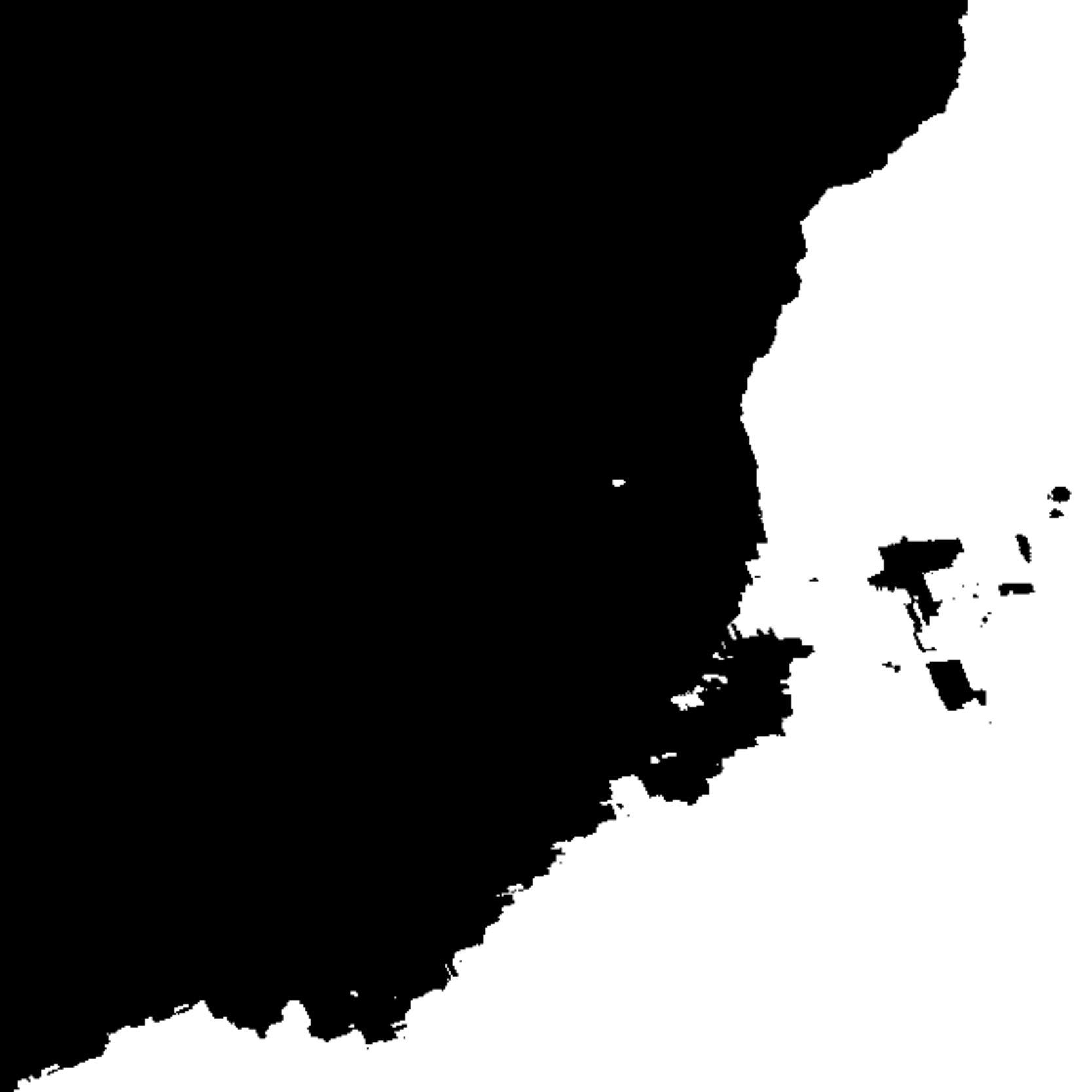}} \vspace{-0.39in}\\

{\fontsize{0.4cm}{1em}\selectfont Mantelli-Neto et al.} \vspace{1.5cm}& 
\fbox{\includegraphics[height=0.08\textheight]{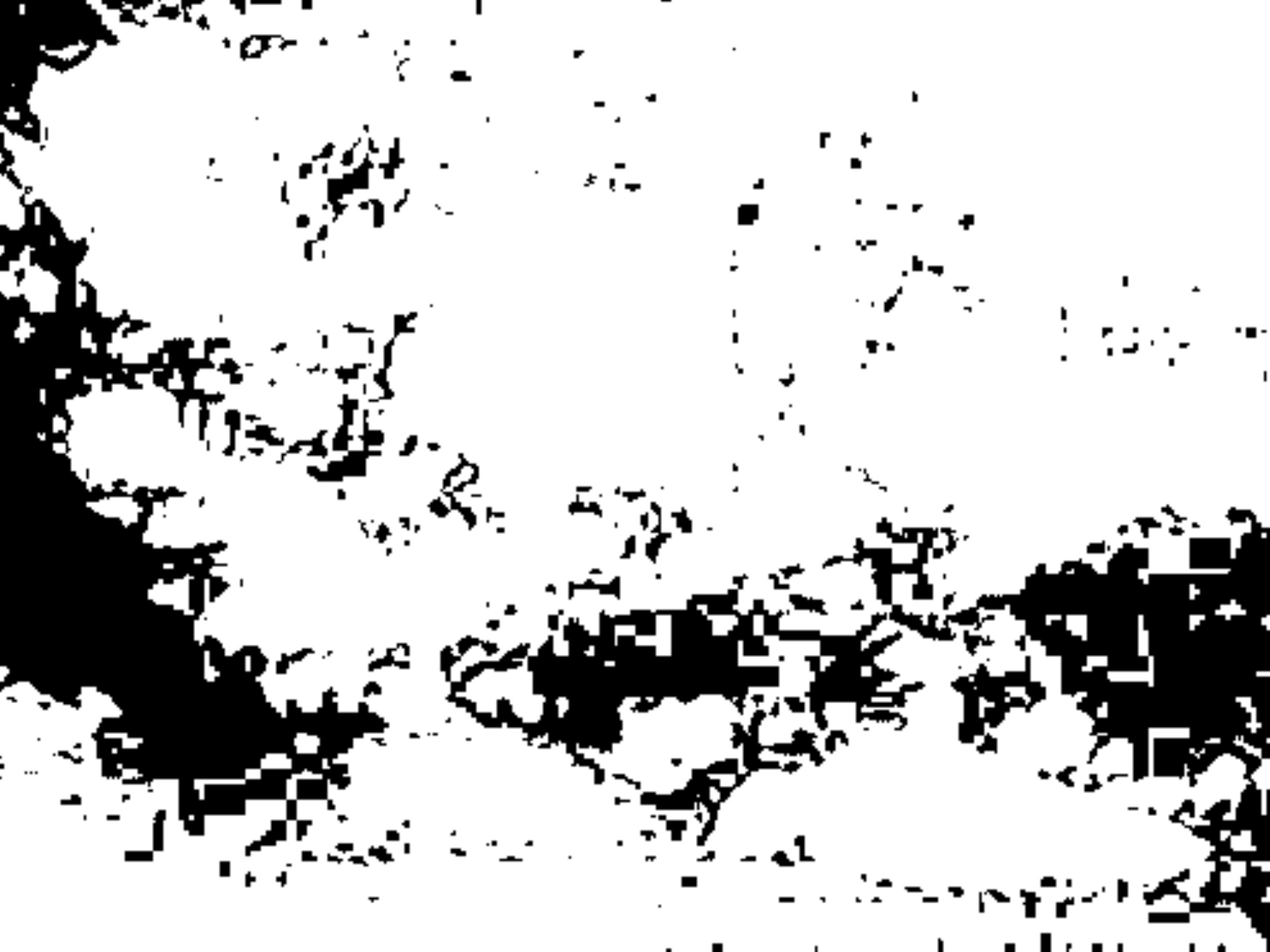}} &
\fbox{\includegraphics[height=0.08\textheight]{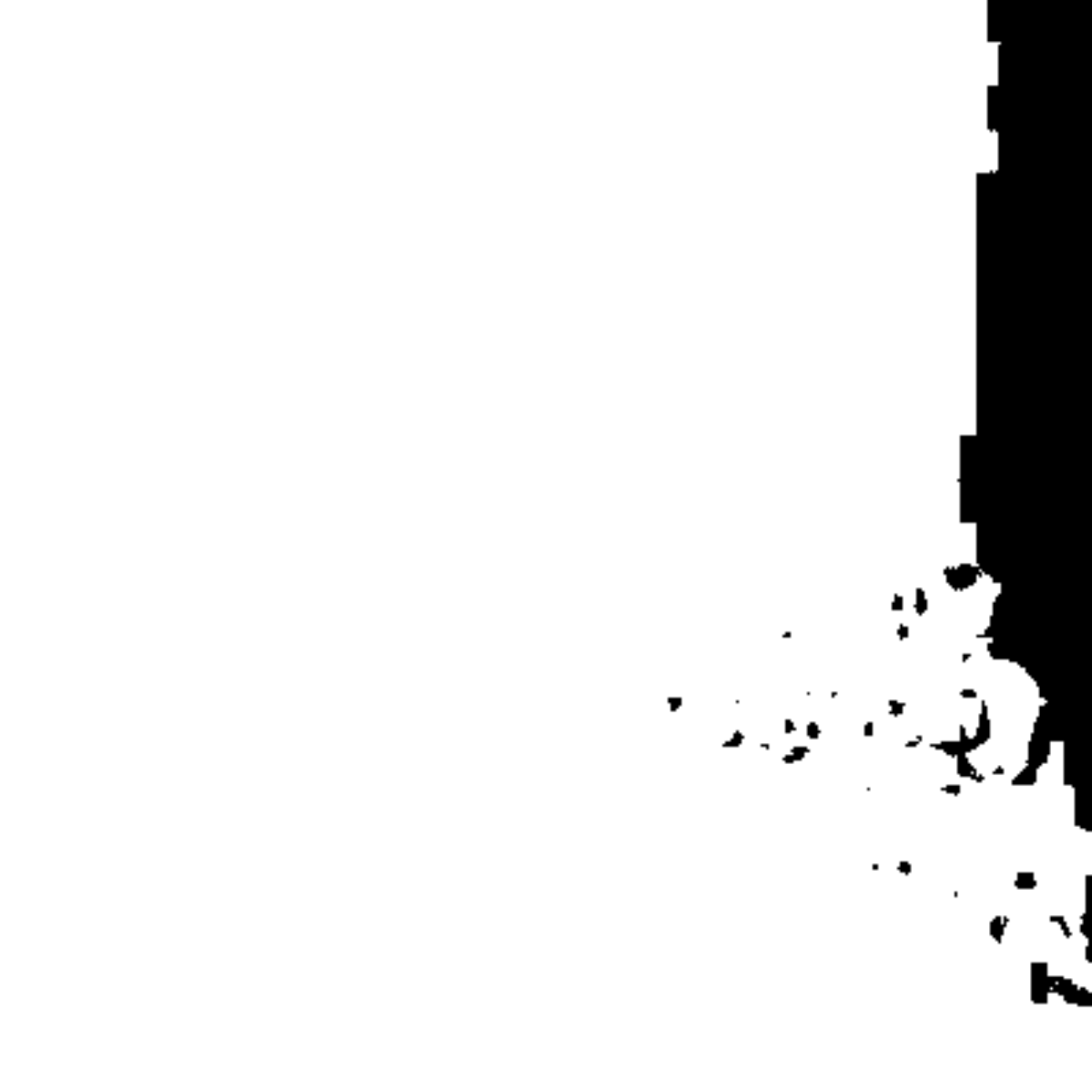}} &
\includegraphics[height=0.08\textheight]{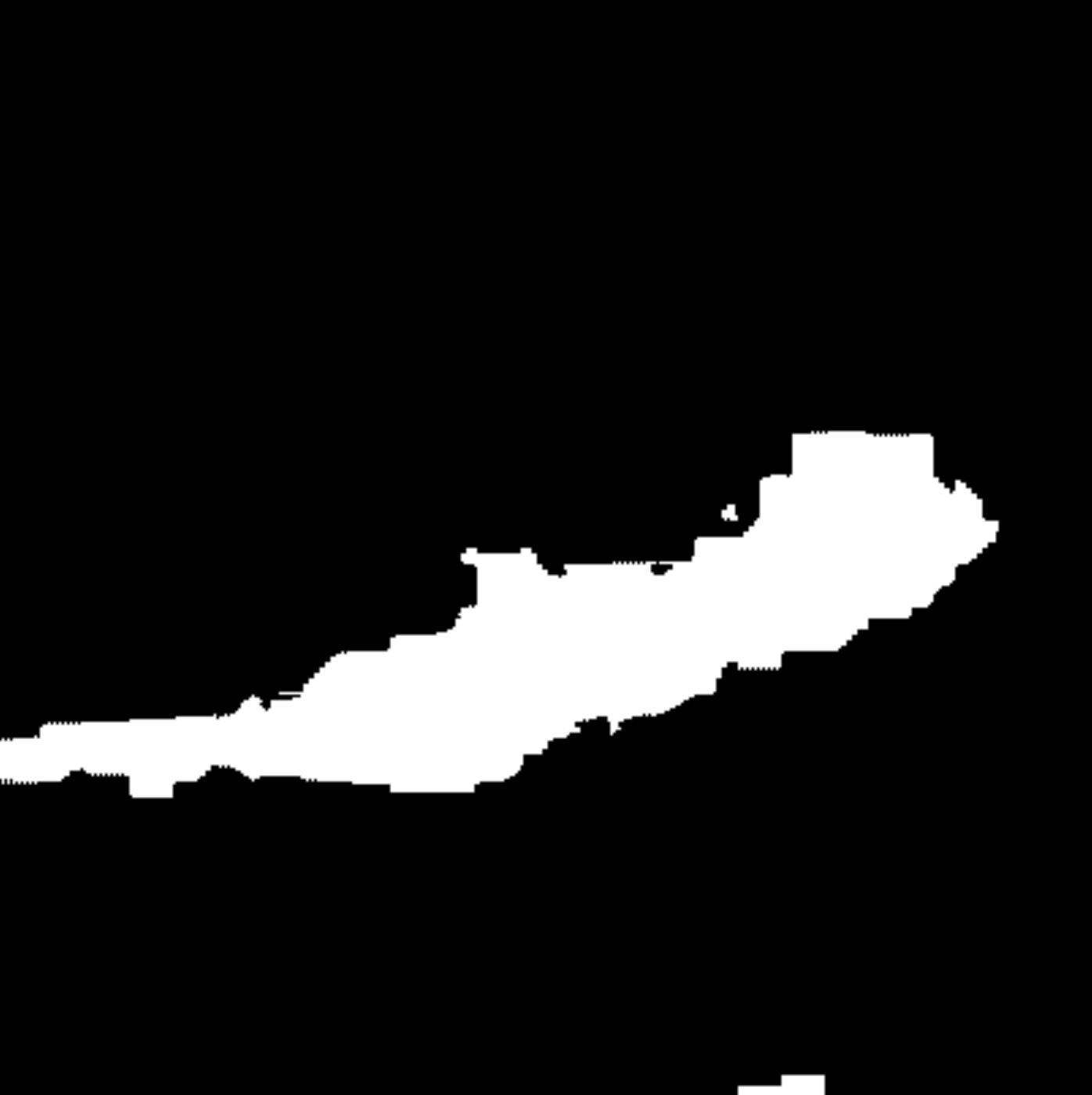} &
\fbox{\includegraphics[height=0.08\textheight]{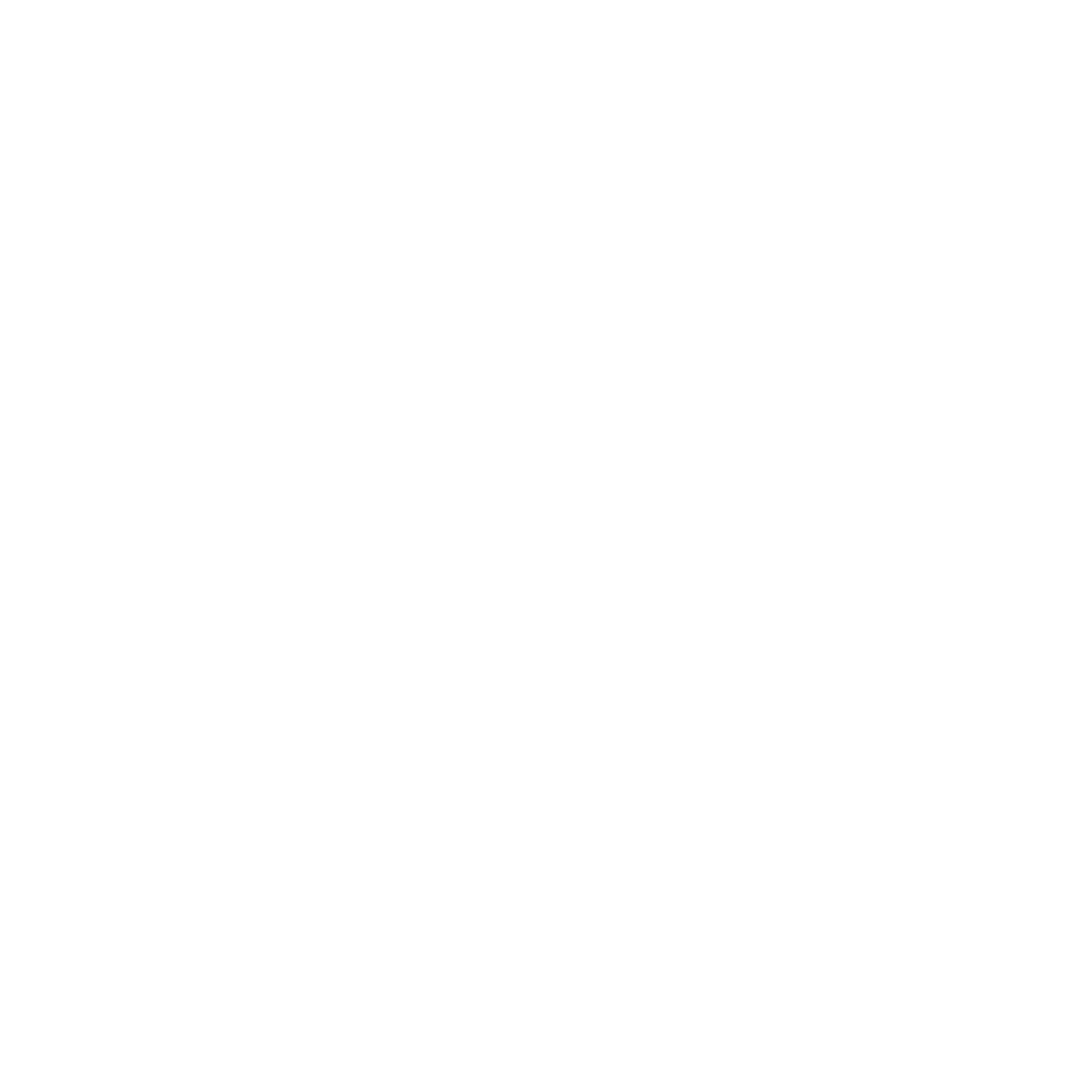}} &
\fbox{\includegraphics[height=0.08\textheight]{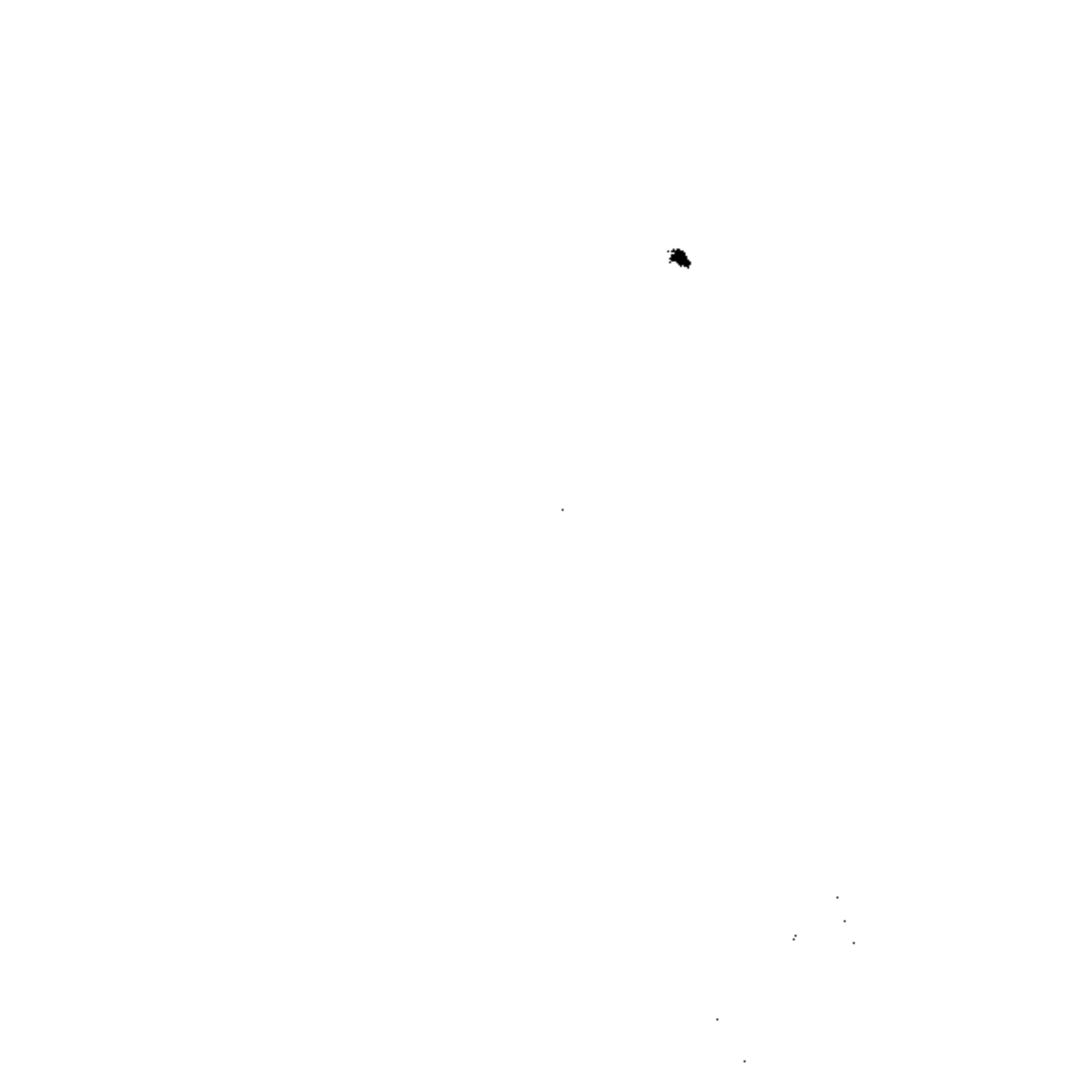}} &
\fbox{\includegraphics[height=0.08\textheight]{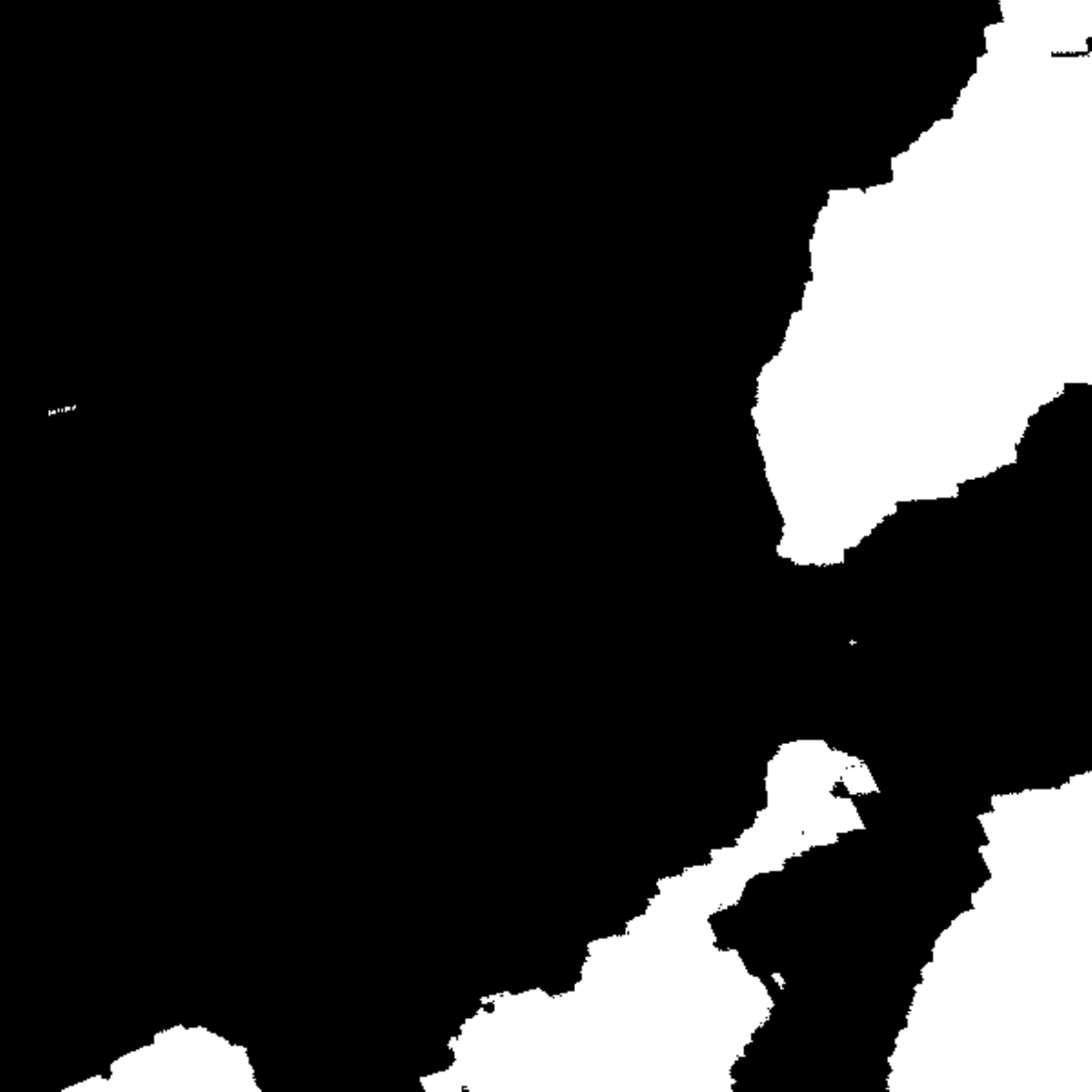}} \vspace{-0.39in}\\

{\fontsize{0.4cm}{1em}\selectfont SLIC + \mbox{DBSCAN}} \vspace{1.5cm}& 
\includegraphics[height=0.08\textheight]{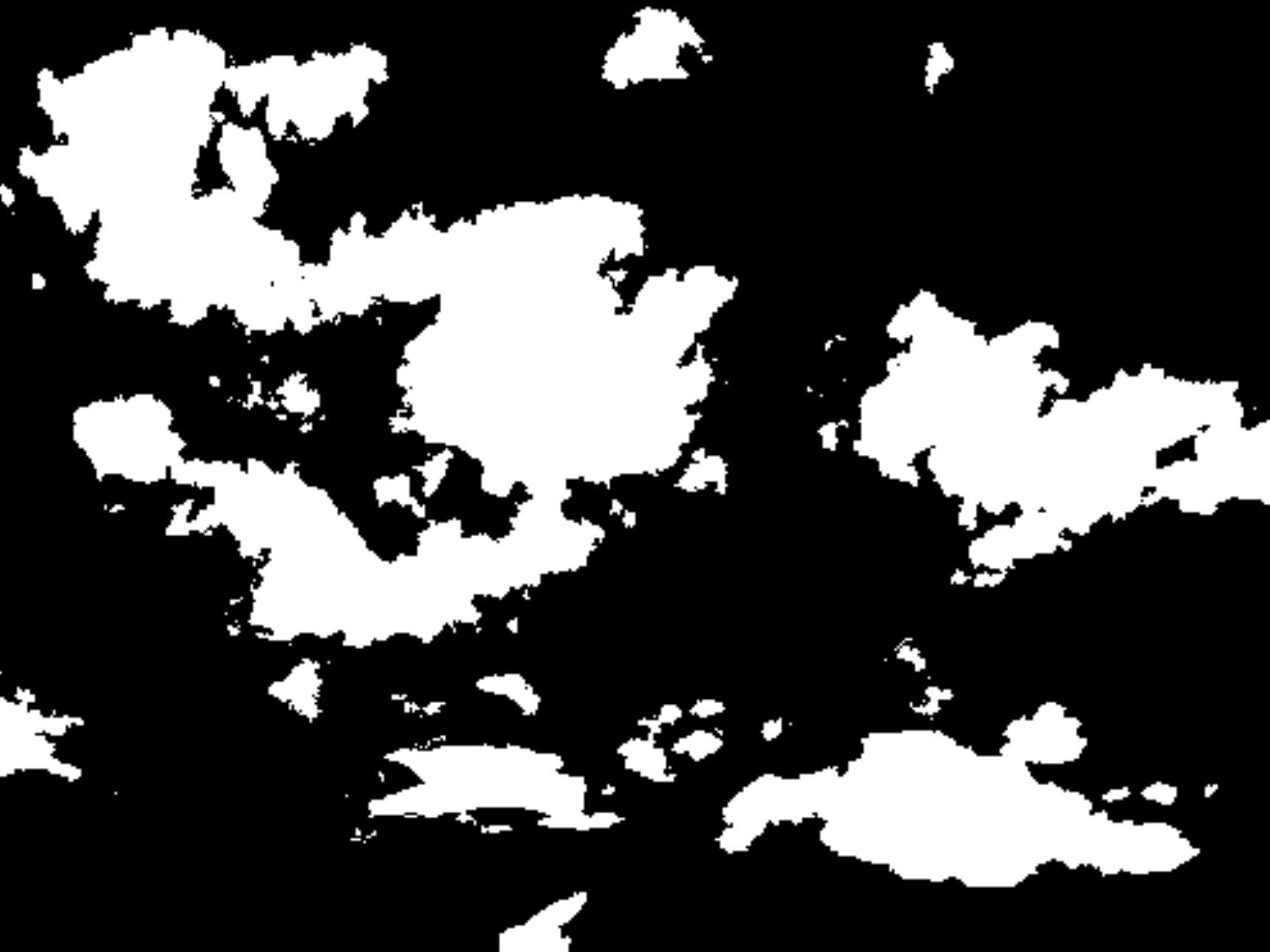} &
\includegraphics[height=0.08\textheight]{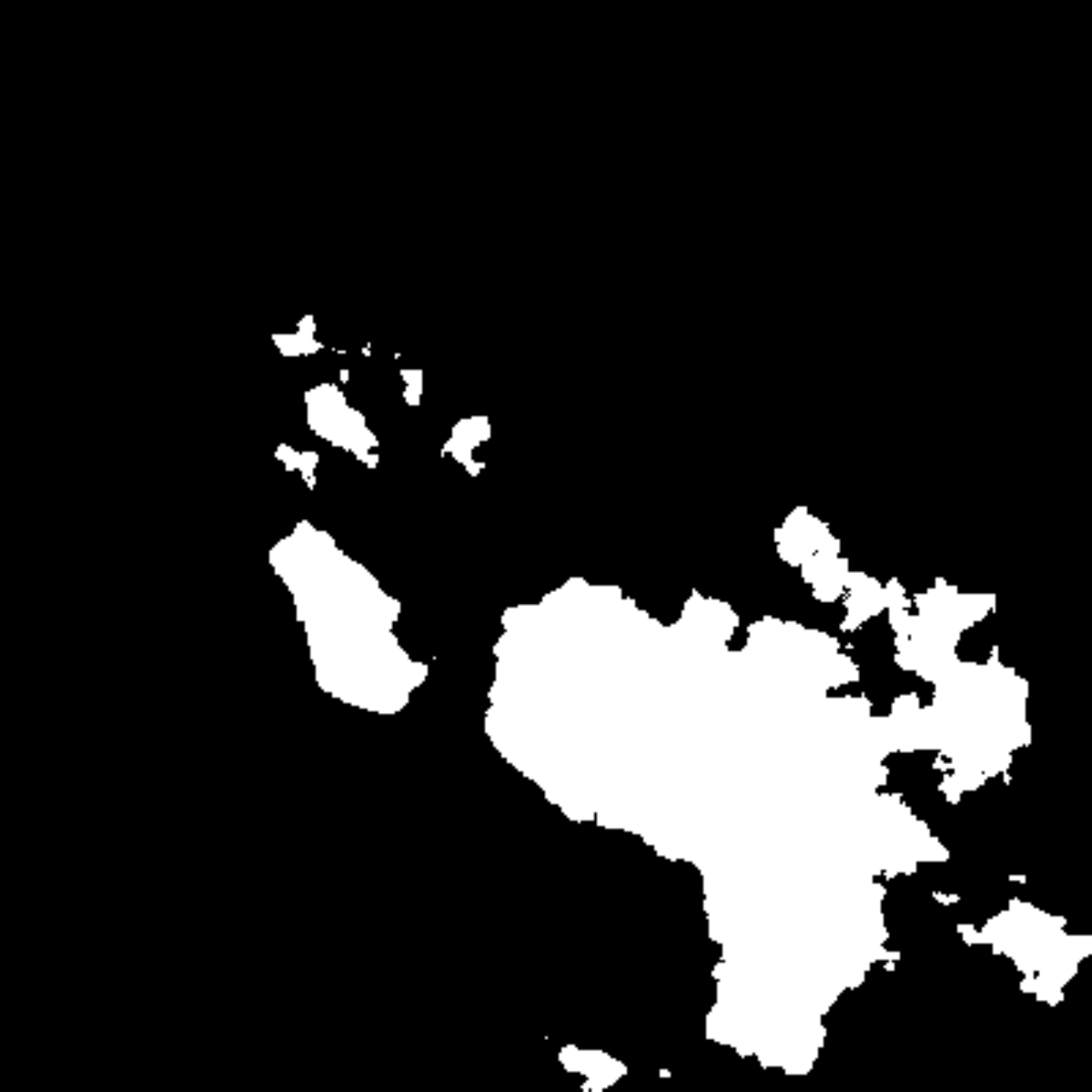} &
\fbox{\includegraphics[height=0.08\textheight]{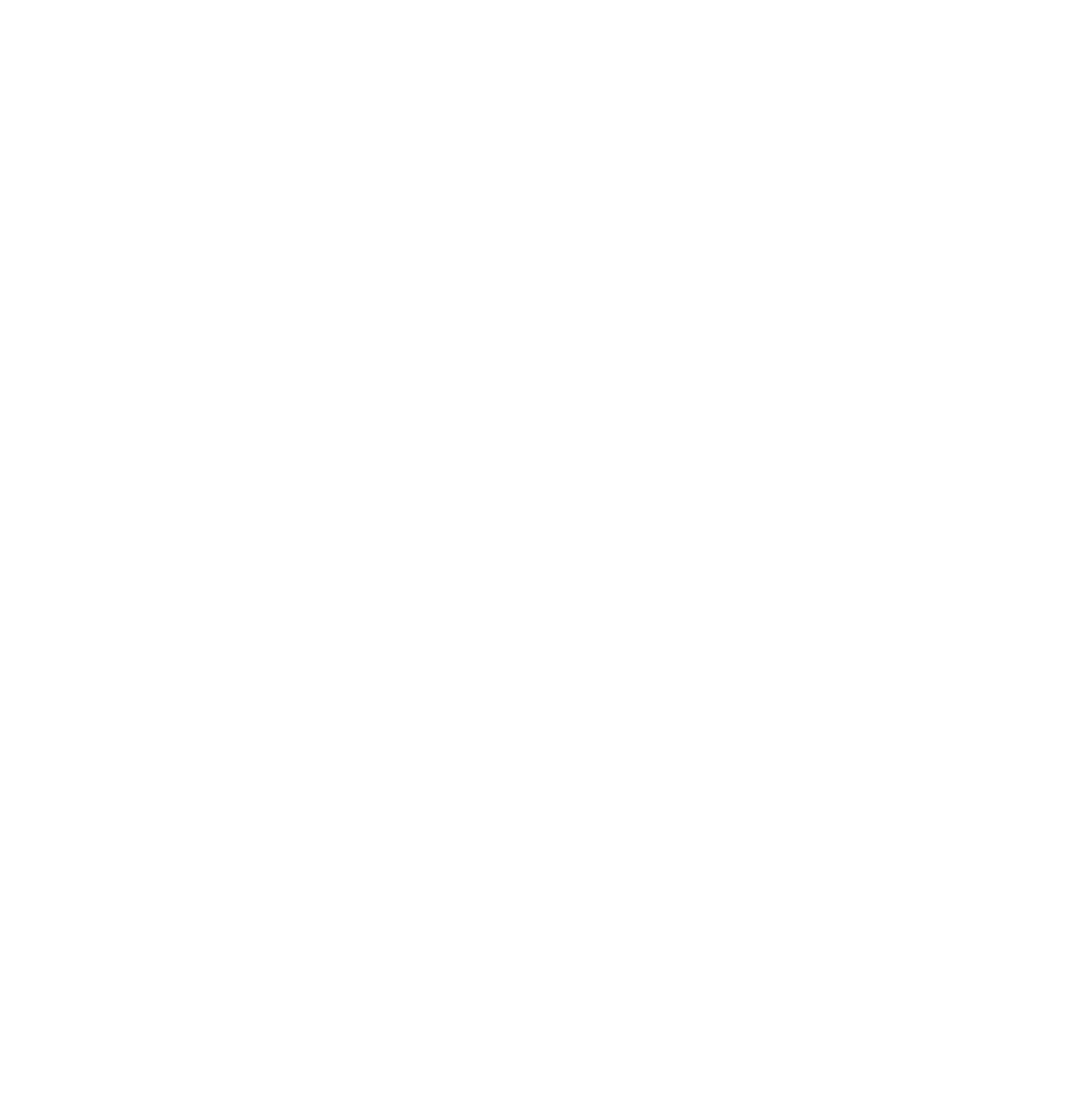}} &
\fbox{\includegraphics[height=0.08\textheight]{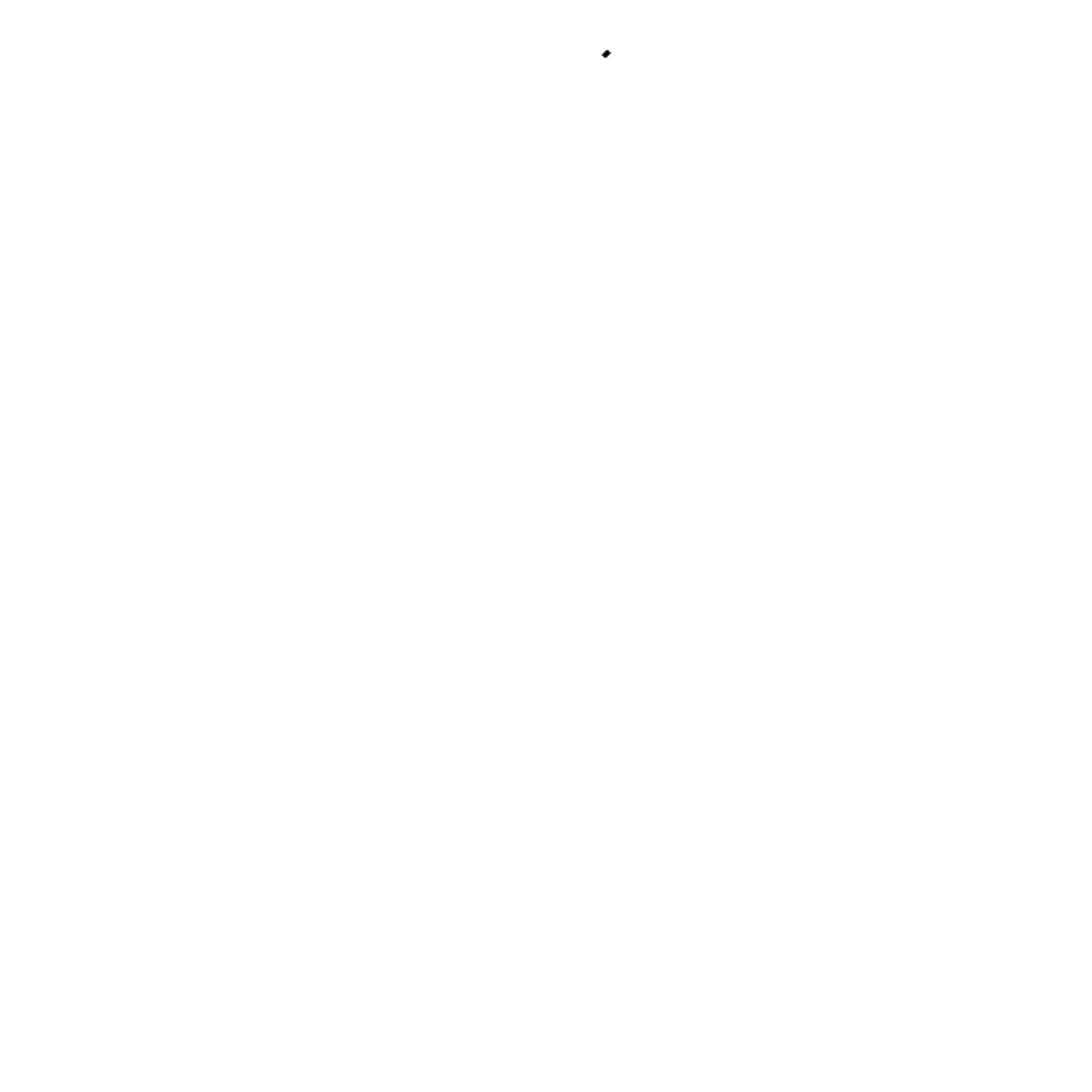}} &
\fbox{\includegraphics[height=0.08\textheight]{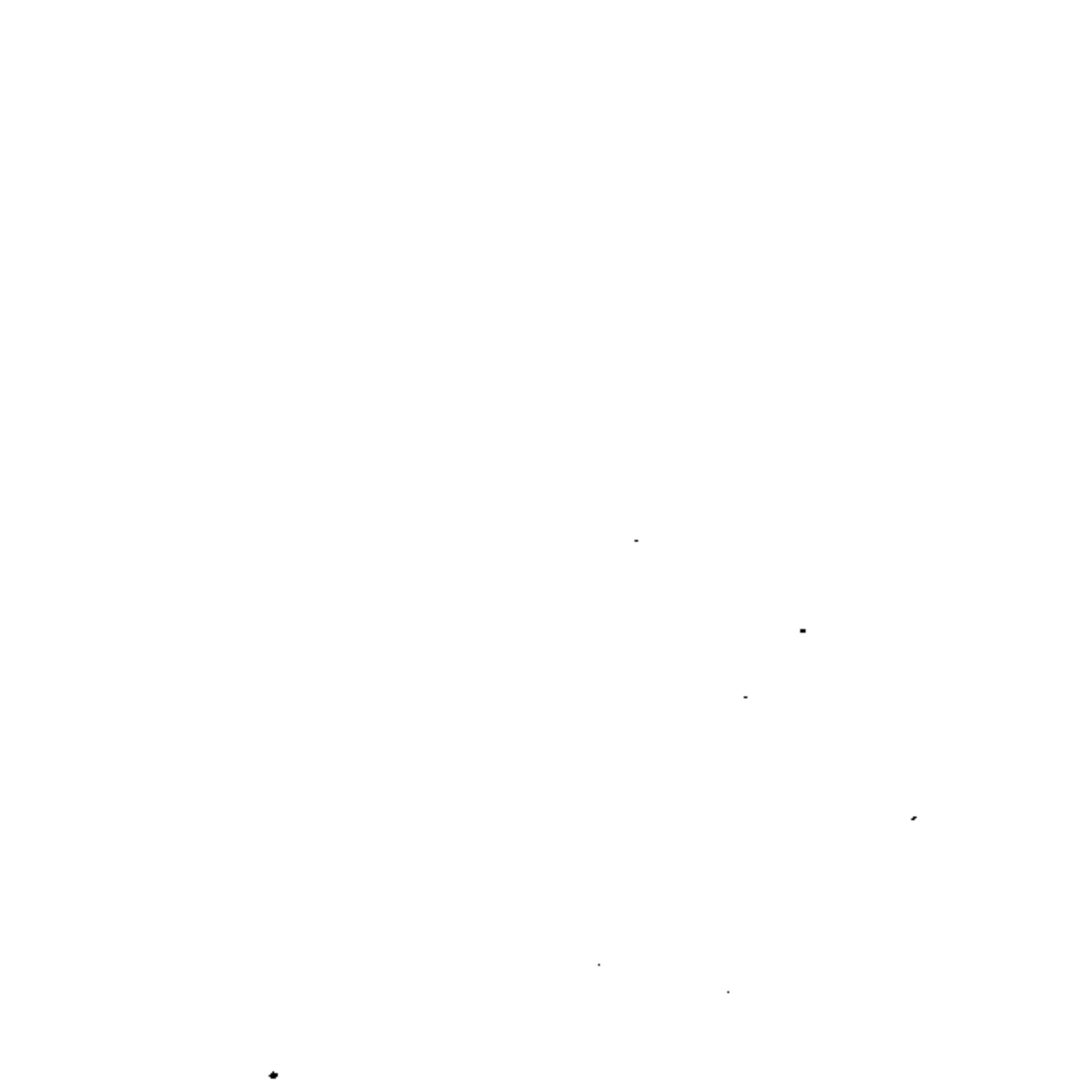}} &
\fbox{\includegraphics[height=0.08\textheight]{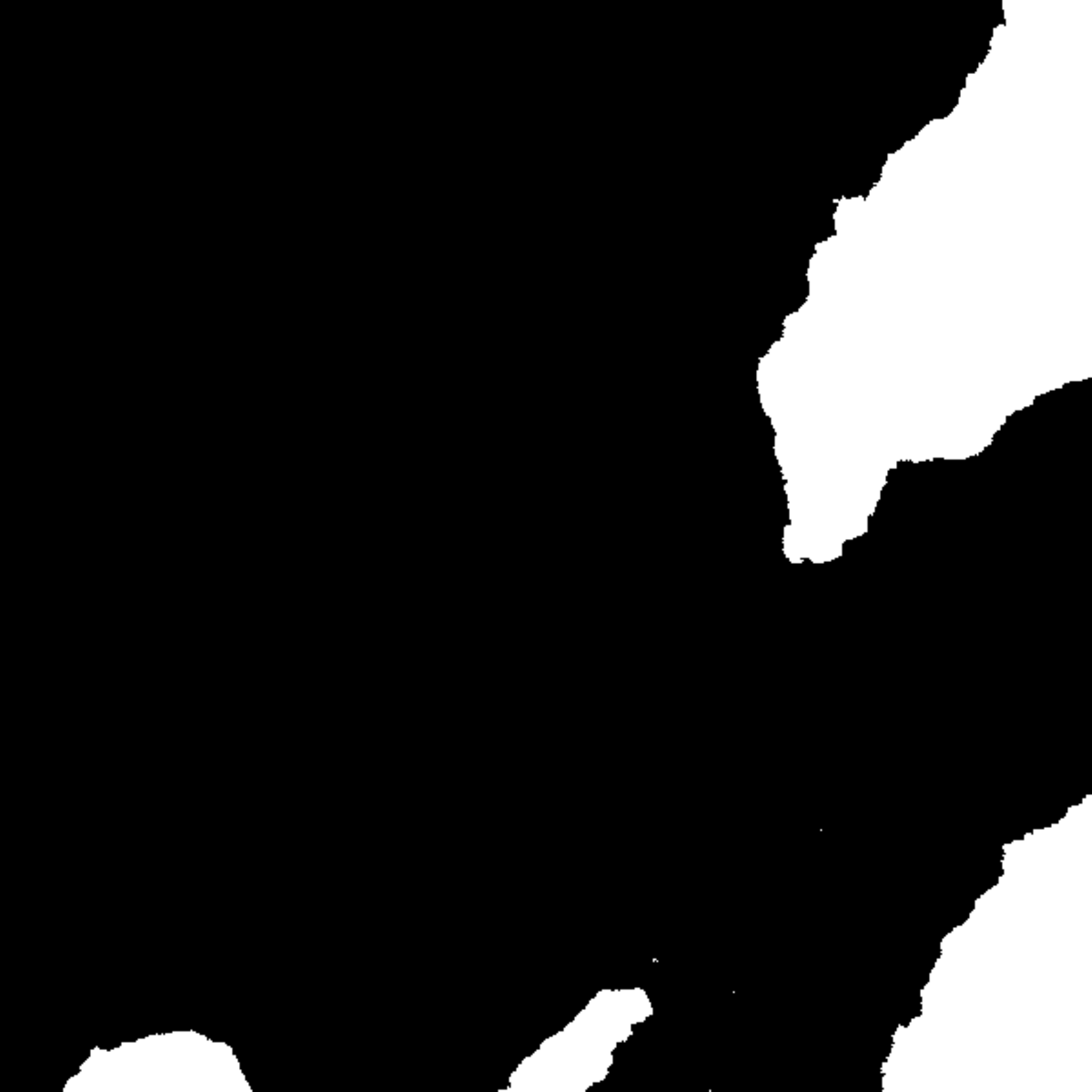}} \vspace{-0.39in}\\

{\fontsize{0.4cm}{1em}\selectfont GRAY + SVM} \vspace{1.2cm}& 
\fbox{\includegraphics[height=0.08\textheight]{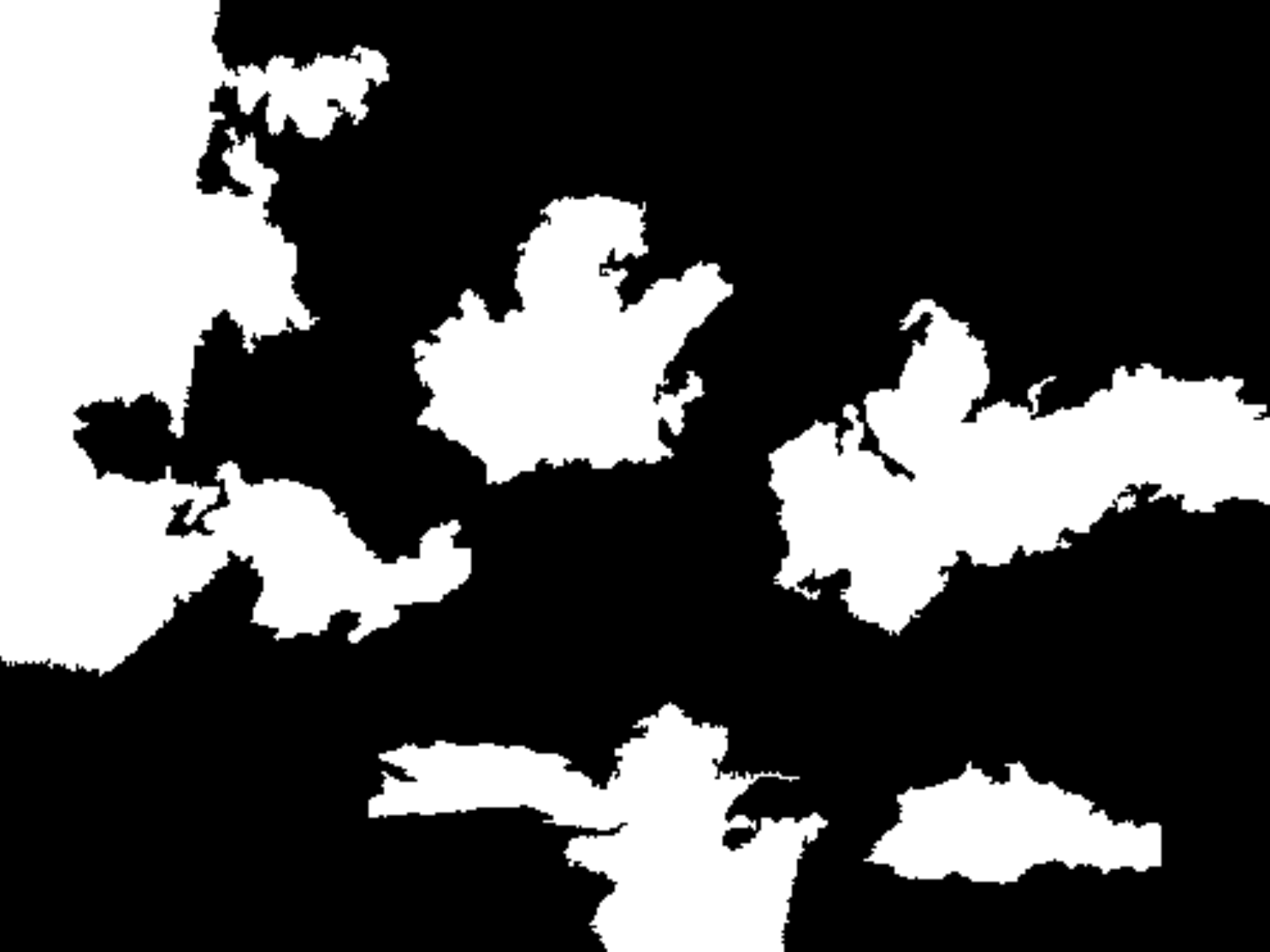}} &
\includegraphics[height=0.08\textheight]{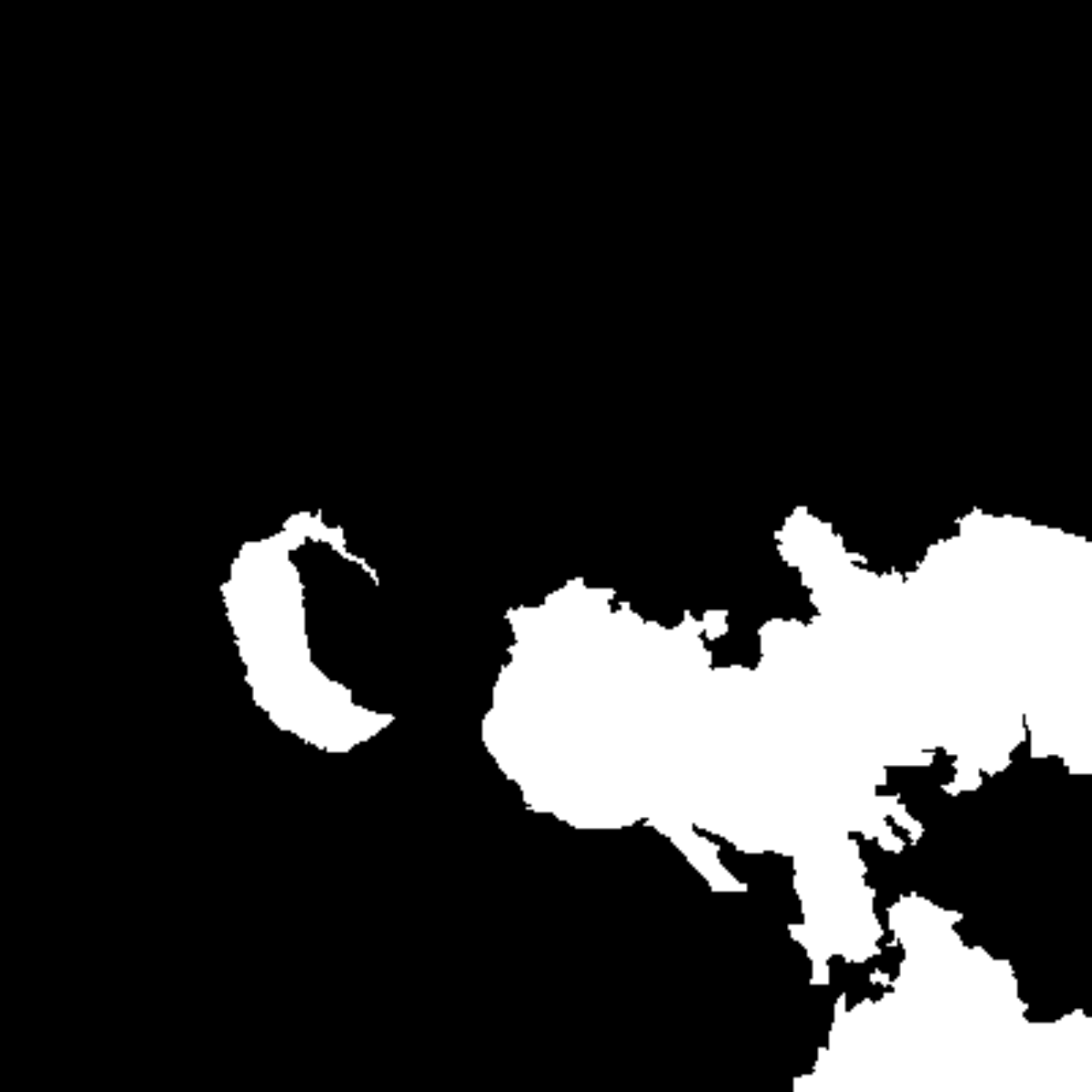} &
\fbox{\includegraphics[height=0.08\textheight]{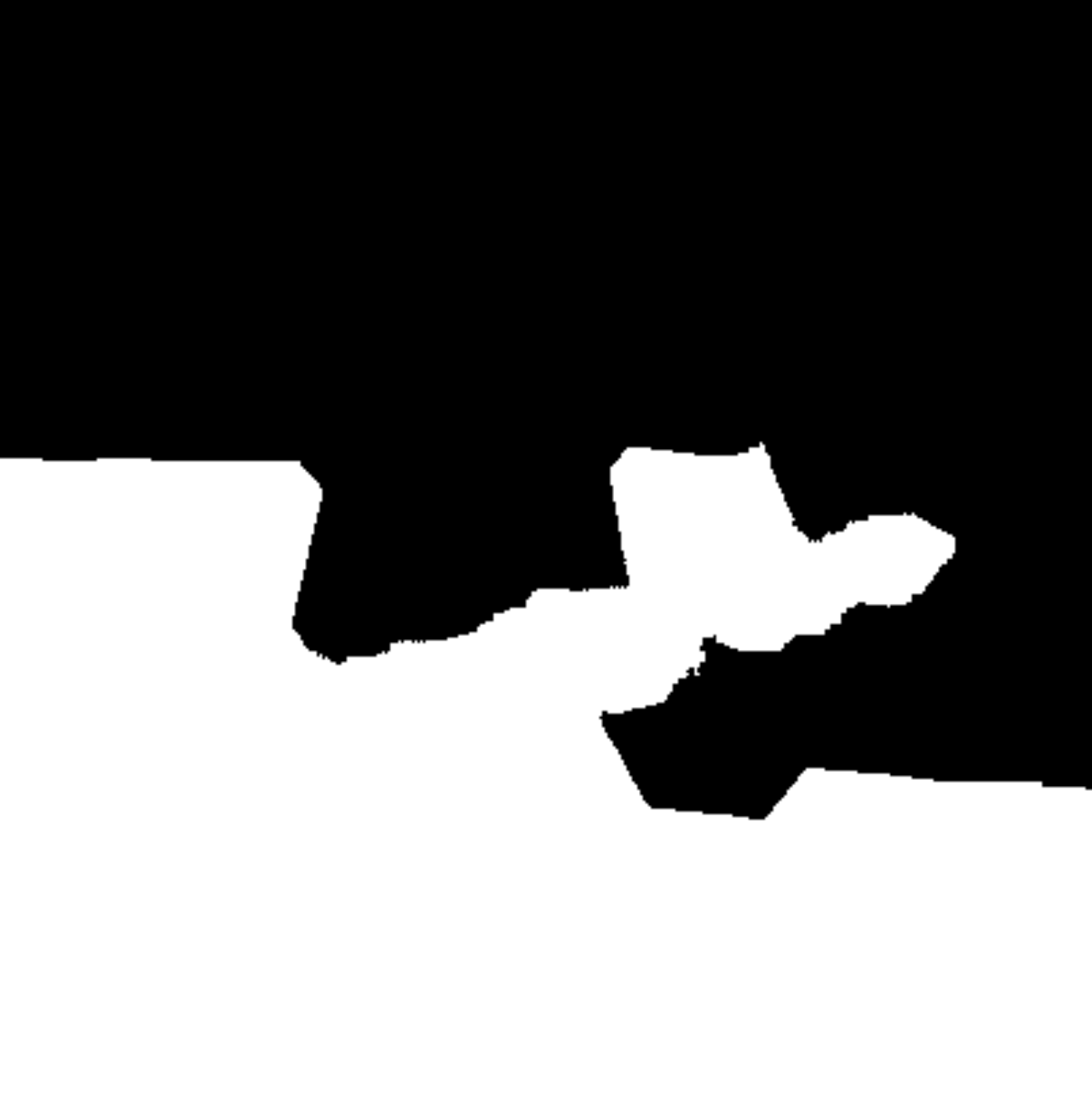}} &
\fbox{\includegraphics[height=0.08\textheight]{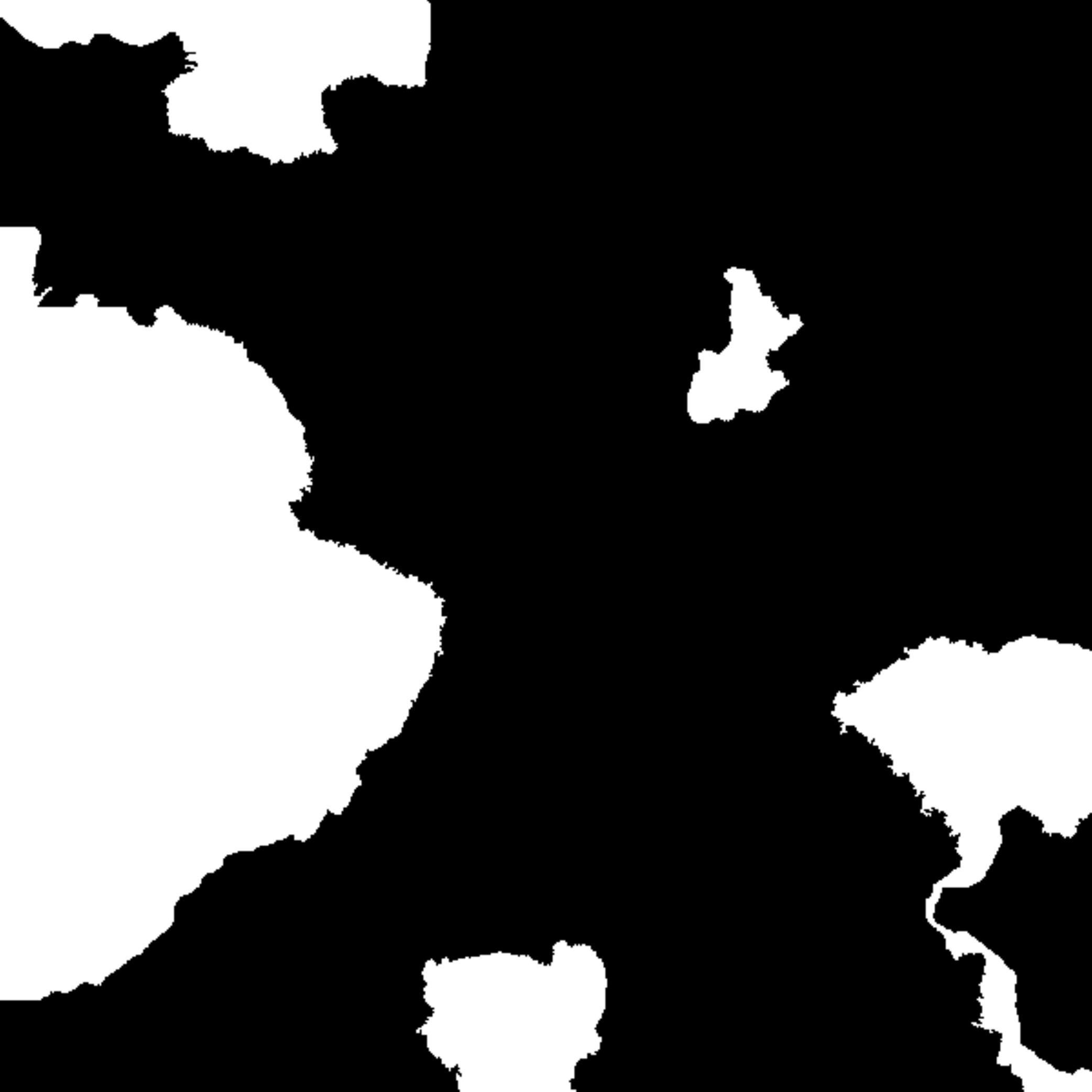}} &
\fbox{\includegraphics[height=0.08\textheight]{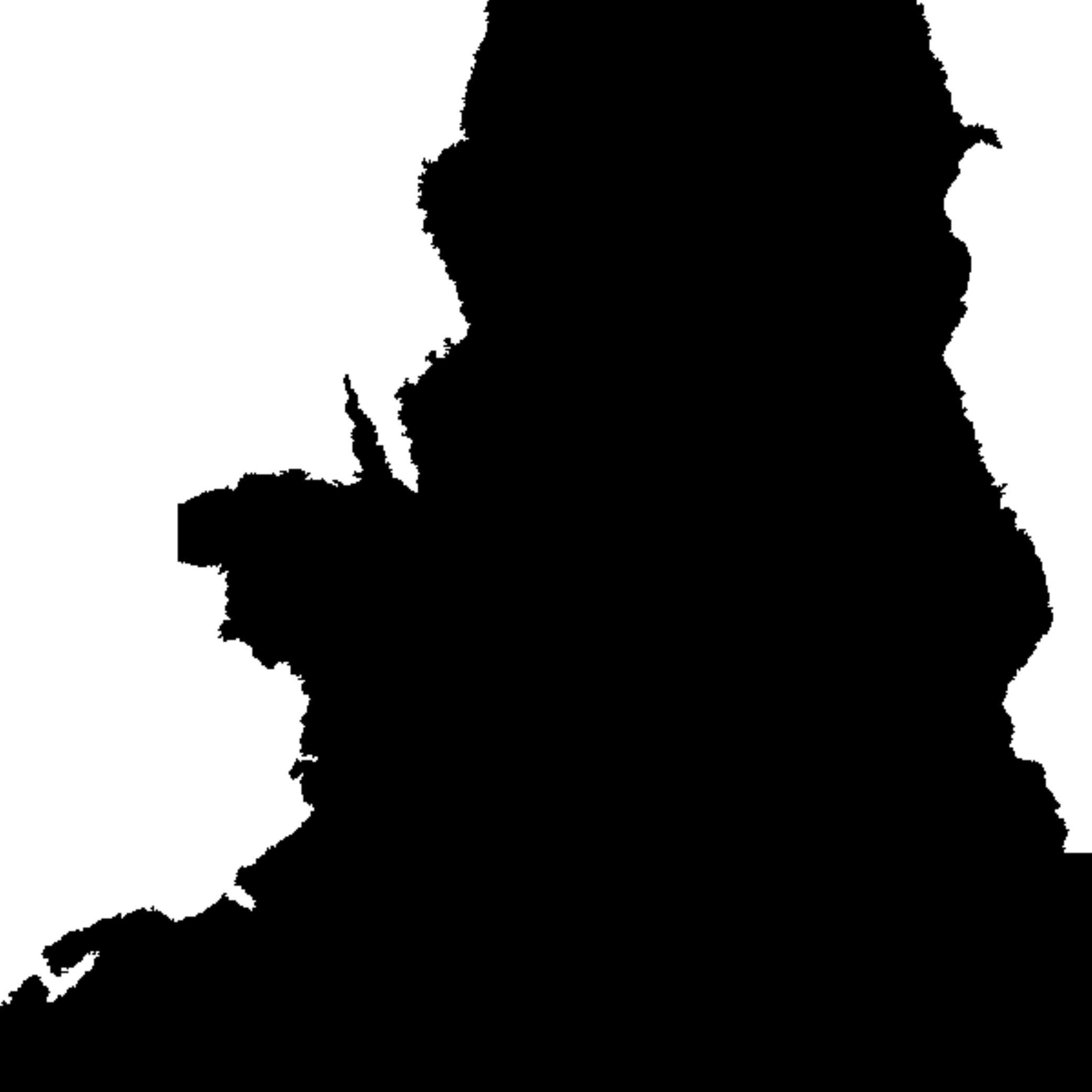}} &
\fbox{\includegraphics[height=0.08\textheight]{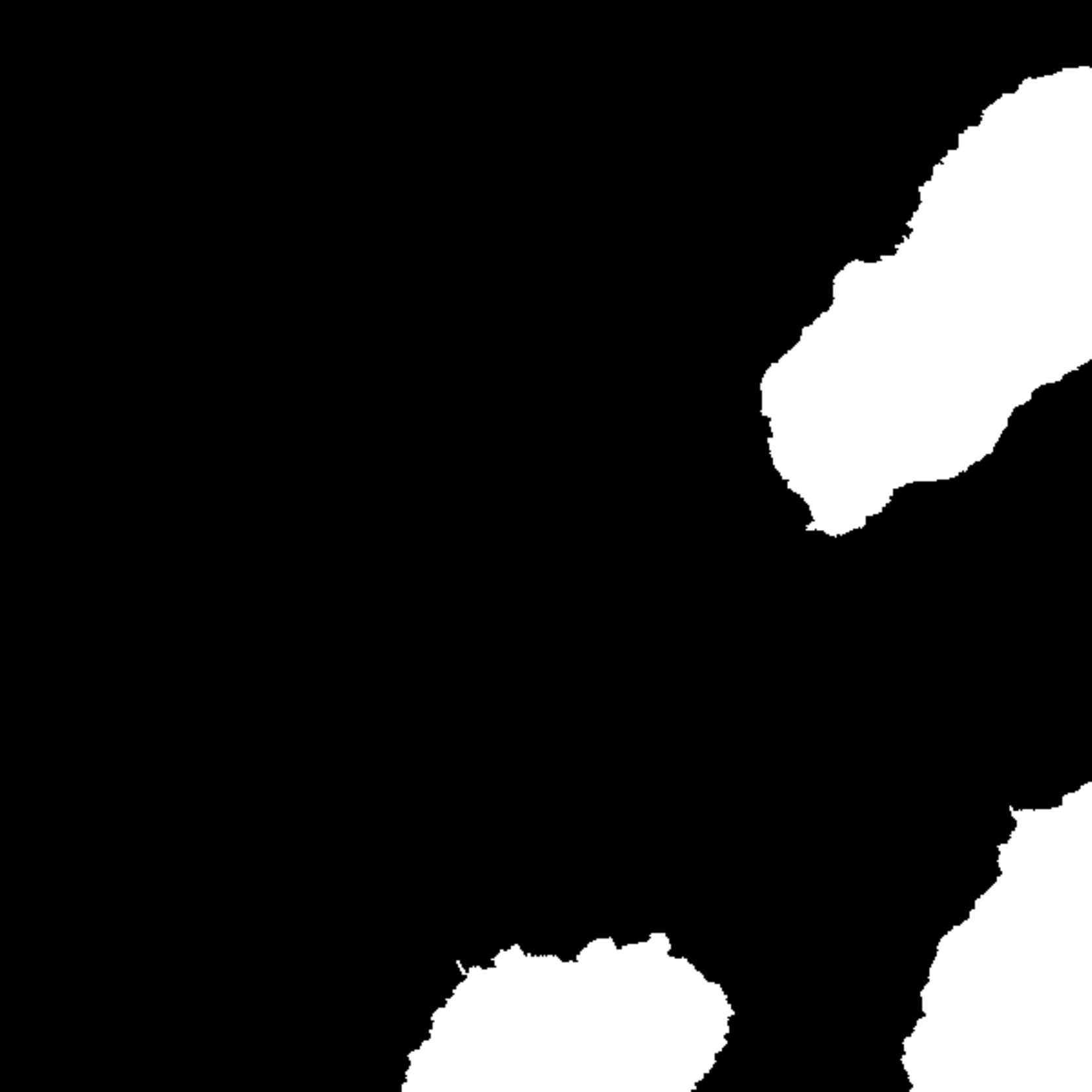}} \vspace{-0.39in}\\

{\fontsize{0.4cm}{1em}\selectfont dSIFT + BOW + SVM} \vspace{1.5cm}& 
\fbox{\includegraphics[height=0.08\textheight]{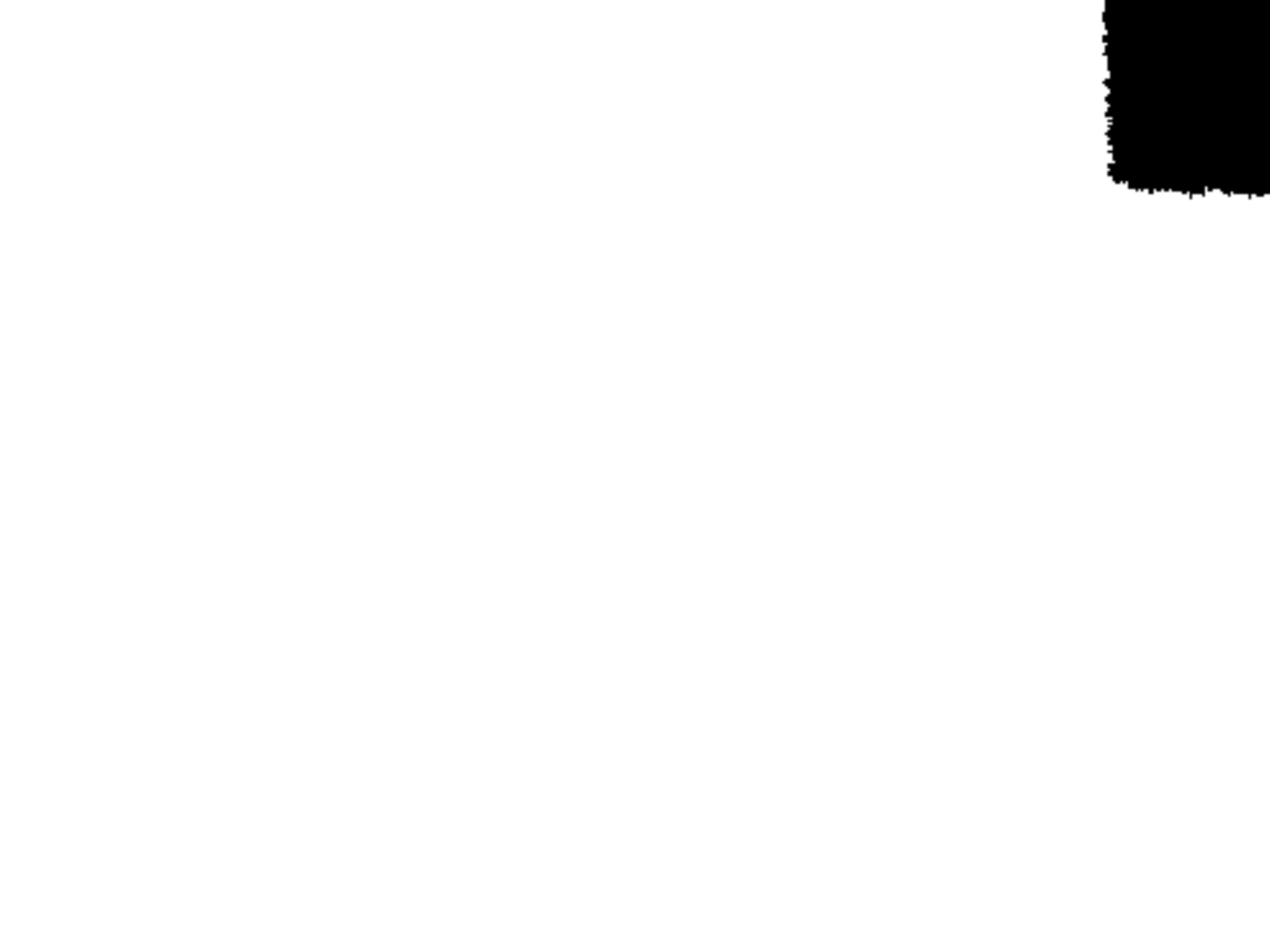}} &
\includegraphics[height=0.08\textheight]{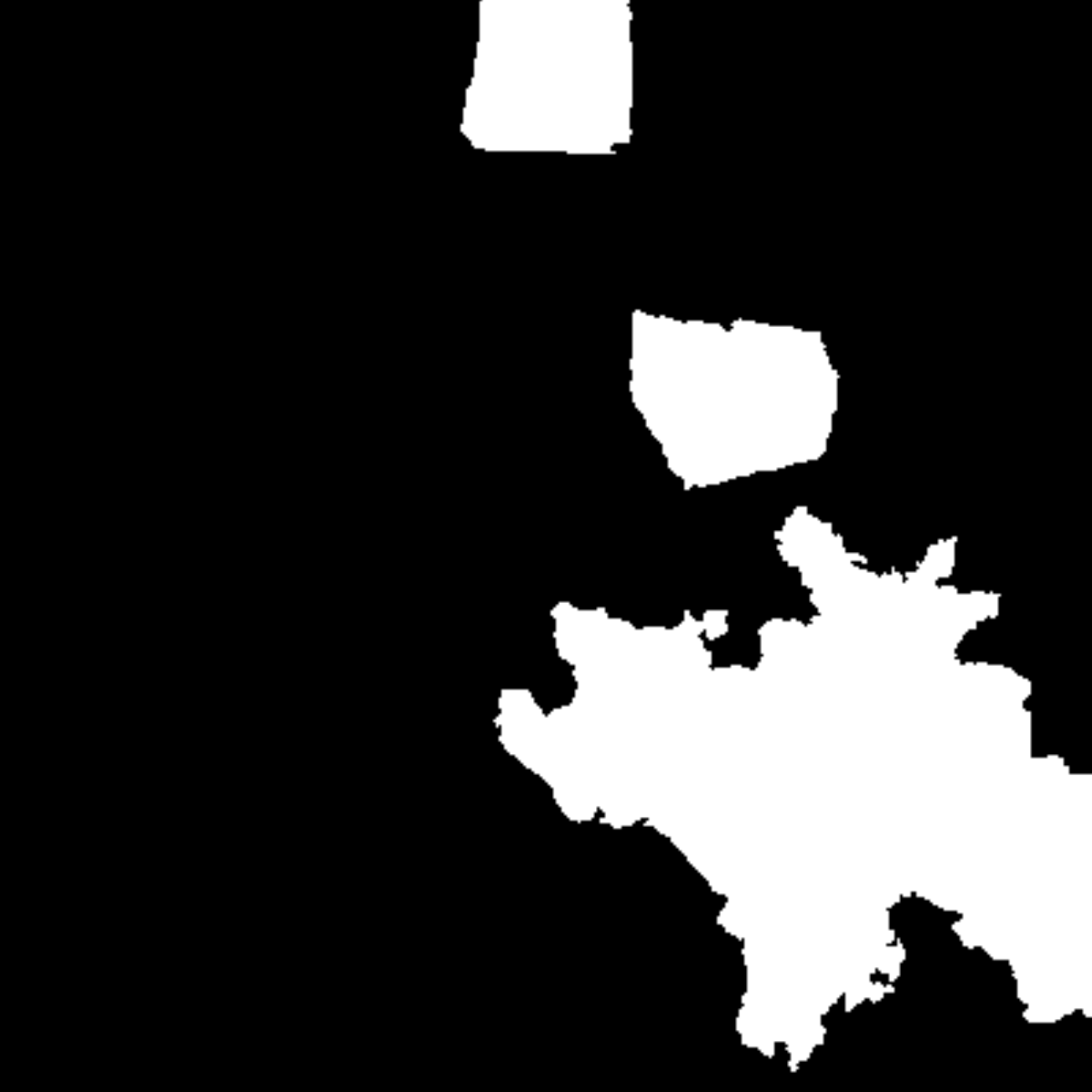} &
\fbox{\includegraphics[height=0.08\textheight]{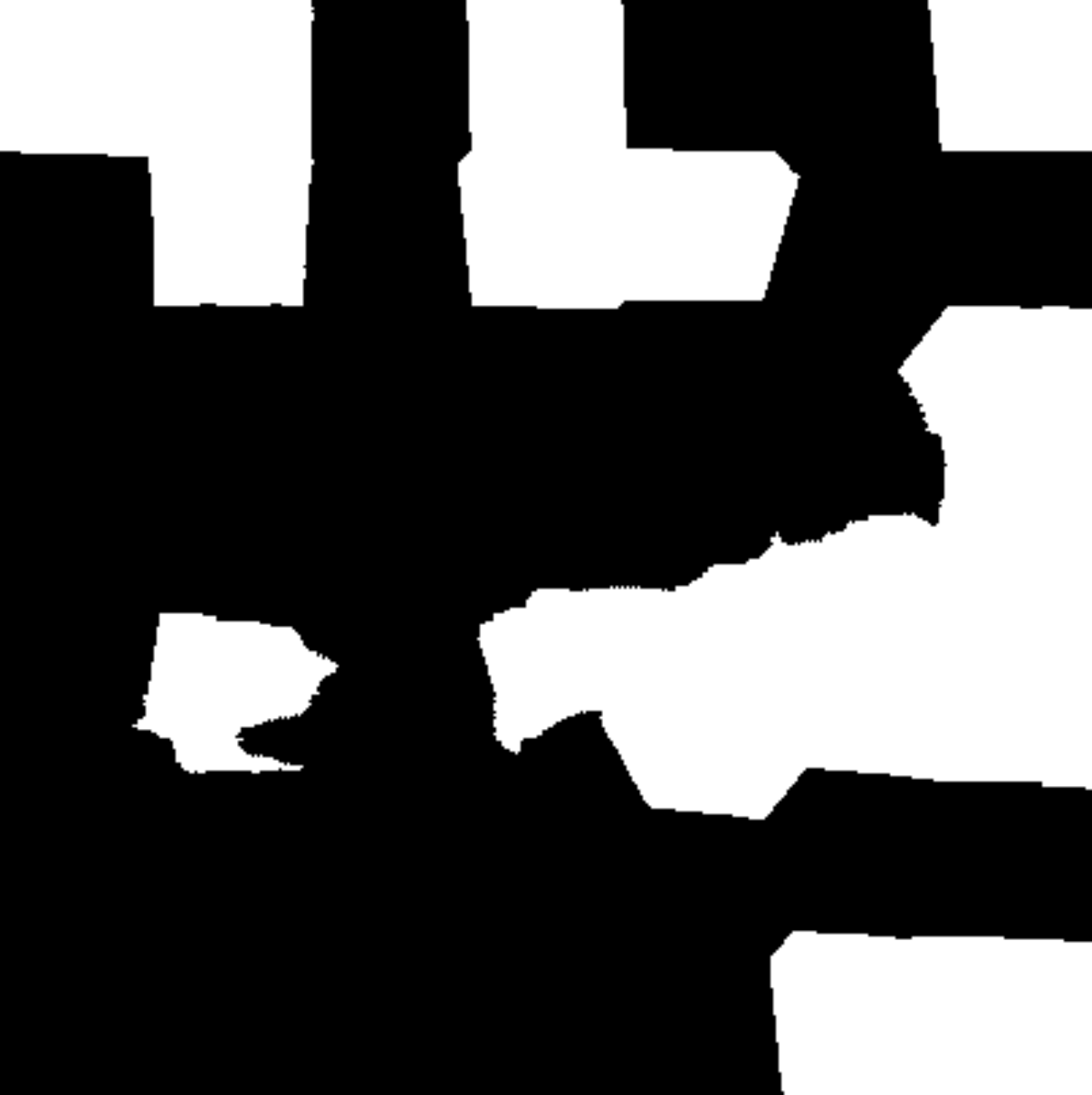}} &
\fbox{\includegraphics[height=0.08\textheight]{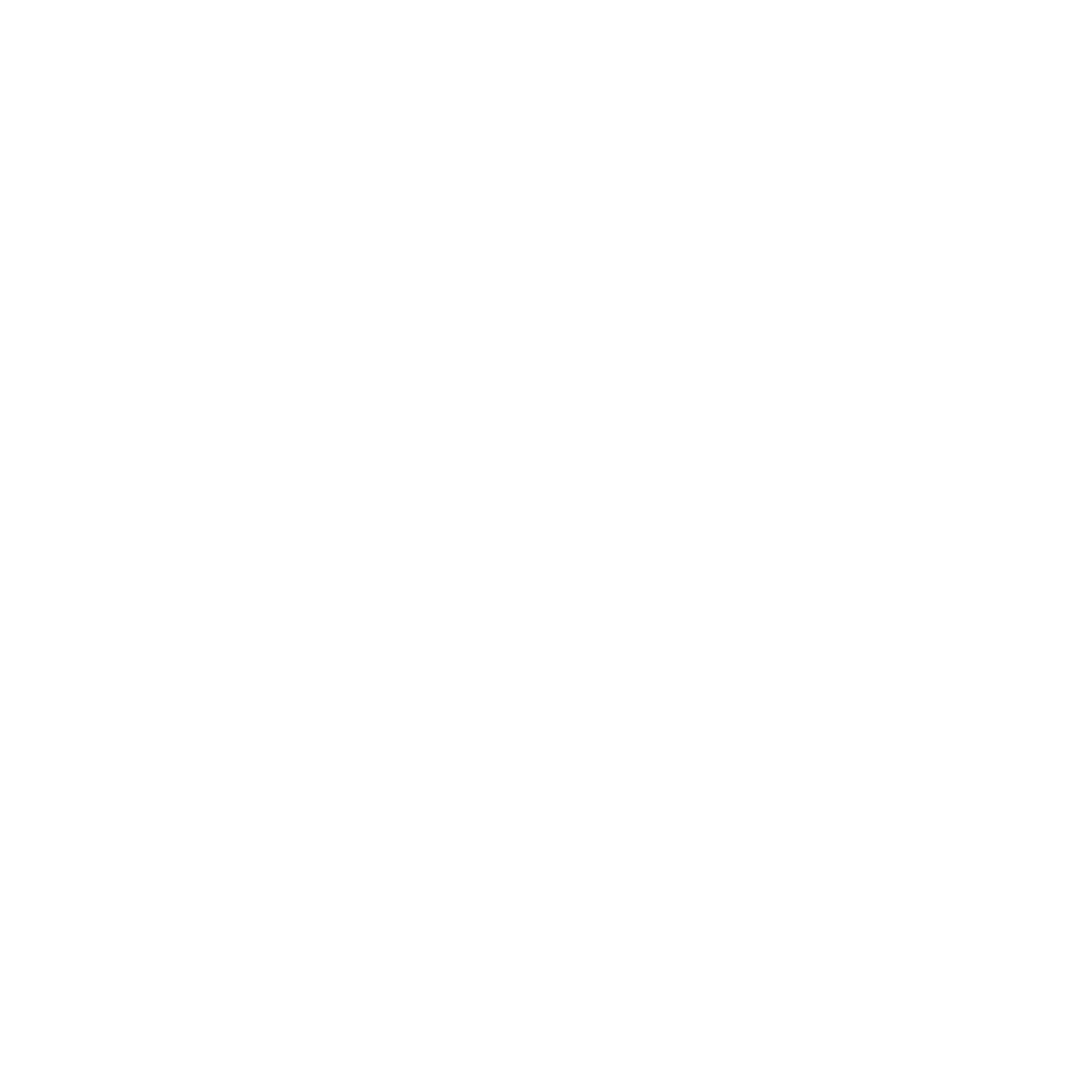}} &
\fbox{\includegraphics[height=0.08\textheight]{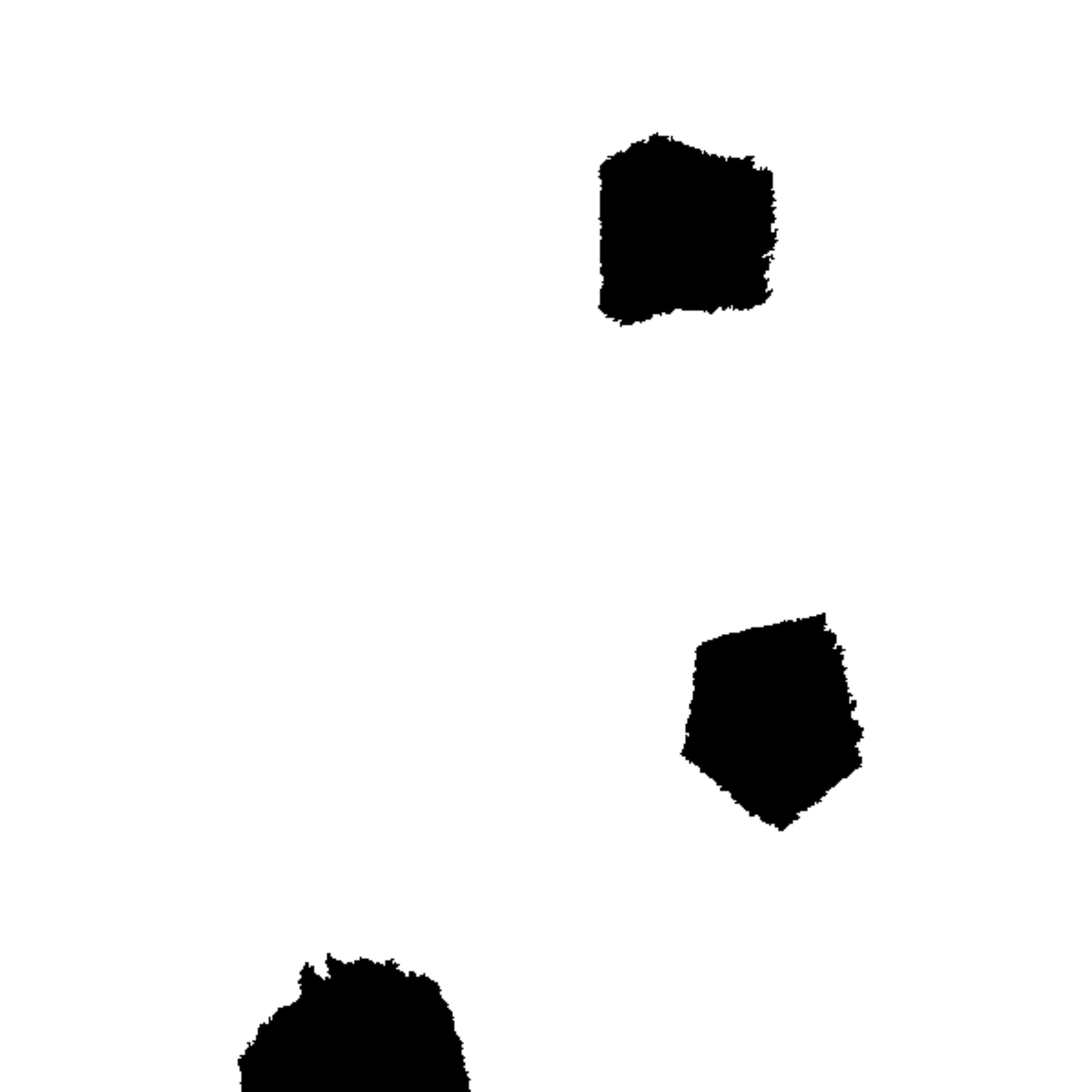}} &
\fbox{\includegraphics[height=0.08\textheight]{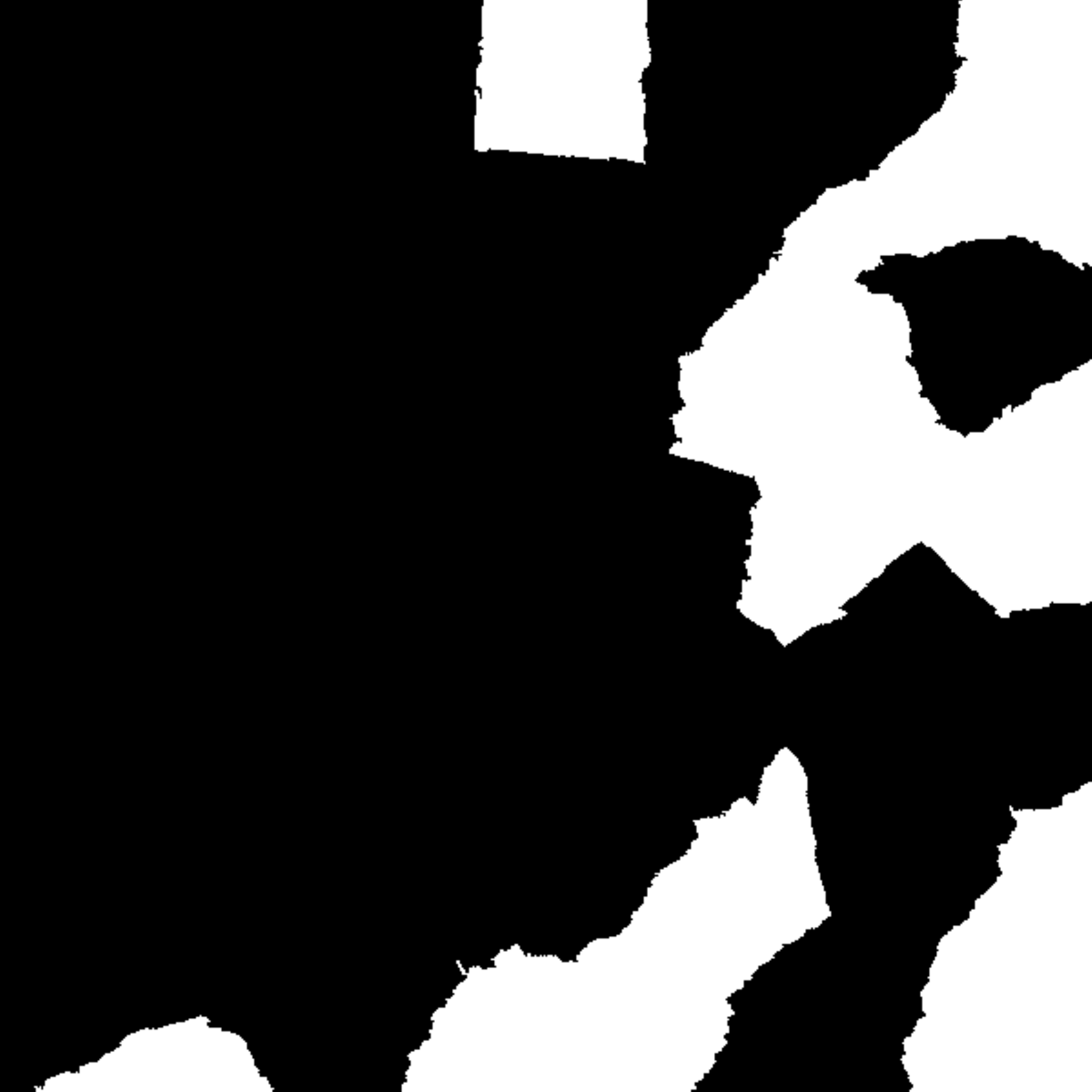}} \vspace{-0.39in}\\

\end{tabular}
\captionof{figure}[foo]{Sample images from HYTA (left 3 columns) and SWIMSEG (right 3 columns) databases, along with the corresponding ground truth and the sky/cloud segmentation obtained using various methods.}
\label{fig:sample-results}
\end{table*}

The simulations were conducted in Matlab on a 64-bit Ubuntu 14.04 LTS workstation, Intel i5 CPU $@$2.67GHz.  In terms of computation time, our proposed method takes an average of  $1.31$s and $1.89$s for a single image from the HYTA and SWIMSEG databases, respectively. In the training stage, HYTA requires $21.5$s and SWIMSEG required $1018.6$s for computing the regression coefficient matrix $\mathbf{B}$. The computation times of other benchmarking algorithm are provided in Table~\ref{scoretable}.

Table~\ref{scoretable} provides an objective evaluation of our proposed approach with other state-of-the-art algorithms for all images in the testing set of the two databases.  Existing algorithms require extensive tuning using various manually-defined thresholds, and achieve either high precision or high recall in our evaluation, but not both. Long et al.\ and Mantelli-Neto et al.\ have high recall values. However, they fare poorly in the corresponding precision values. Souza et al.\ on the other hand has a high precision value, but low recall. These existing methods often suffer from under- or over-identification of cloud pixels as they are threshold-based, and a single threshold may not work well for all types of sky/cloud images. Our proposed approach performs best for both HYTA and SWIMSEG databases on the basis of F-scores, highlighting the effectiveness of PLS-based learning method. 

\begin{table*}[htb!]
\small
\centering
\begin{tabular}{ |lr||r|r|r|c|c| }
\hline
& Methods & Precision & Recall & F-score & Misclassification rate & Time [s] \\ 
\hline\hline
\parbox[t]{3mm}{\multirow{12}{*}{\rotatebox[origin=c]{90}{\small \textbf{HYTA}}}} 
& Li et al.\ & 0.89 & 0.82 & 0.81 & 0.12 & 1.34 \\ 
& Souza et al.\ & 0.88 &  0.67 &  0.65 & 0.15 & 1.33 \\ 
& Long et al.\ & 0.64 & \textbf{0.97} & 0.71  & 0.26 & 1.22 \\ 
& Mantelli-Neto et al.\ & 0.54 & \textbf{0.97} & 0.63 & 0.44 & 1.43 \\ 
& SLIC + DBSCAN & 0.65 & 0.83 & 0.60 & 0.32 & 5.10 \\ 
& GRAY + SVM & 0.89 & 0.58 & 0.67 & 0.31 & 2.44  \\
& LBP + SVM & 0.80 & 0.65 & 0.72 & 0.22 & 4.29 \\
& ColorHIST + SVM & 0.75 & 0.74 & 0.75 & 0.21 & 2.47  \\
& dSIFT + BOW + SVM & 0.61 & 0.66 & 0.62 & 0.38 & 4.91 \\
& Texture + BOW + SVM & 0.82 & 0.62 & 0.69 & 0.25 & 2.60  \\
& Proposed Method & \textbf{0.94} & 0.80 & \textbf{0.85} & \textbf{0.10} & 1.31  \\ 
\hline\hline
\parbox[t]{3mm}{\multirow{12}{*}{\rotatebox[origin=c]{90}{\small \textbf{SWIMSEG}}}} 
& Li et al.\ & 0.90 & 0.86 & 0.88 & 0.11 & 2.06  \\ 
& Souza et al.\ & 0.95 &  0.76 &  0.81 & 0.14 & 2.04 \\ 
& Long et al.\ & 0.71 & \textbf{0.98} & 0.80 & 0.23 & 1.83  \\ 
& Mantelli-Neto et al.\ & 0.70 & 0.97 & 0.79 & 0.24 & 2.16 \\ 
& SLIC + DBSCAN & 0.72 & 0.79 & 0.65 & 0.33 & 5.39 \\ 
& GRAY + SVM & 0.87 & 0.56 & 0.64 & 0.33 & 2.61 \\
& LBP + SVM & 0.62 & 0.65 &  0.63 & 0.36 & 4.73 \\
& ColorHIST + SVM & 0.81 & 0.64 & 0.66 & 0.31 & 2.63  \\
& dSIFT + BOW + SVM & 0.65 & 0.88 & 0.72 & 0.28 & 5.04 \\
& Texture + BOW + SVM & 0.82 & 0.71 & 0.70 & 0.31 & 2.74  \\
& Proposed Method  & \textbf{0.92} & 0.90 & \textbf{0.90} & \textbf{0.09} & 1.89 \\ 
\hline
\end{tabular}
\caption{Performance evaluation using binary ground truth of HYTA and SWIMSEG databases. The best performance according to each criterion is indicated in bold. All experiments are evaluated on the same set of testing images from the HYTA and SWIMSEG databases. We also list the computation time for a single image for all algorithms.}
\label{scoretable}
\end{table*}

\section{Discussion}
\label{sec:Discussion}

There are three primary advantages of our proposed approach compared to other cloud detection algorithms.

First, our cloud segmentation framework is not based on any pre-defined assumptions about color spaces and does not place any restrictions on the type of input images.  We systematically compare different color channels and identify the most suitable ones. We also explain the reason for their better performance based on rigorous statistical evaluations in two datasets. 

Second, many existing cloud segmentation algorithms rely on a set of thresholds, conditions, and/or parameters that are manually defined for a particular sky/cloud image database. Our proposed cloud segmentation approach is entirely learning-based and thus provides a systematic solution to training for a given database. 

Third, conventional algorithms provide a binary output image from the input sky/cloud image. Although these binary images are informative in most cases, they lack flexibility and robustness. We have no indication of the effectiveness of the thresholding of the input image. In reality, because of the nature of clouds and cloud images, it is undoubtedly better to employ a \emph{soft} thresholding approach. Our proposed approach achieves a probabilistic classification of cloud pixels. This is very informative as it provides a general sense of \emph{belongingness} of pixels to the cloud class. However, as only binary ground-truth images are available, we convert these probability maps into binary images for performing a quantitative evaluation of our algorithm. In our future work, we plan to extend this analysis by creating \emph{probabilistic ground truth} images, where the ground truth of the input images is generated by aggregating  annotations from multiple experts.

Finally, from the extensive experiments on the two datasets (cf.\ Table~\ref{scoretable}), we observe that the performance with images from the SWIMSEG database is generally better than for HYTA, even though the behavior of color channels is similar in both (i.e.\ the same color channels do well for both SWIMSEG and HYTA).  We believe this is because the images in SWIMSEG were captured with a camera that has been calibrated for color, illumination, and geometry~\cite{WAHRSIS}. 

Naturally, many challenges remain, for example: How do different weather conditions affect the classification performance?  The weather in Singapore is relatively constant in terms of temperature and humidity, with little variation throughout the year.  Usually it is either partly cloudy, or rainy (in which case the sky will be completely overcast, making segmentation unnecessary). As a result, our database is not suitable for investigating this question. Completely overcast conditions can be dealt with by simple pre-processing of the image before segmentation, e.g.\ by calculating the number of clusters, as we have done in our previous work~\cite{ICIP2015a}.

\section{Conclusions}
\label{sec:conc}

We have presented a systematic analysis of color spaces and components, and proposed a probabilistic approach using PLS-based regression for the segmentation of ground-based sky/cloud images. Our approach is entirely learning-based and does not require any manually-defined thresholds, conditions, or parameters at any stage of the algorithm. We also release an extensive sky/cloud image database captured with a calibrated ground-based camera that has been annotated with ground-truth segmentation masks. 

Our future work will include the annotation of a database with probabilistic ground-truth segmentation maps as well as the extension of this method to High-Dynamic-Range (HDR) images.  Going beyond segmentation, it is also important to classify clouds into different types \cite{ICIP2015a} or estimate cloud altitude and movement \cite{IGARSS2015b}, which are both part of our current research.

\section*{Acknowledgment}

This work is supported by a grant from Singapore's Defence Science \& Technology Agency (DSTA).

\ifCLASSOPTIONcaptionsoff
  \newpage
\fi

\bibliographystyle{IEEEbib}

\begin{IEEEbiography}
[{\includegraphics[width=1in,height=1.25in,clip,keepaspectratio]{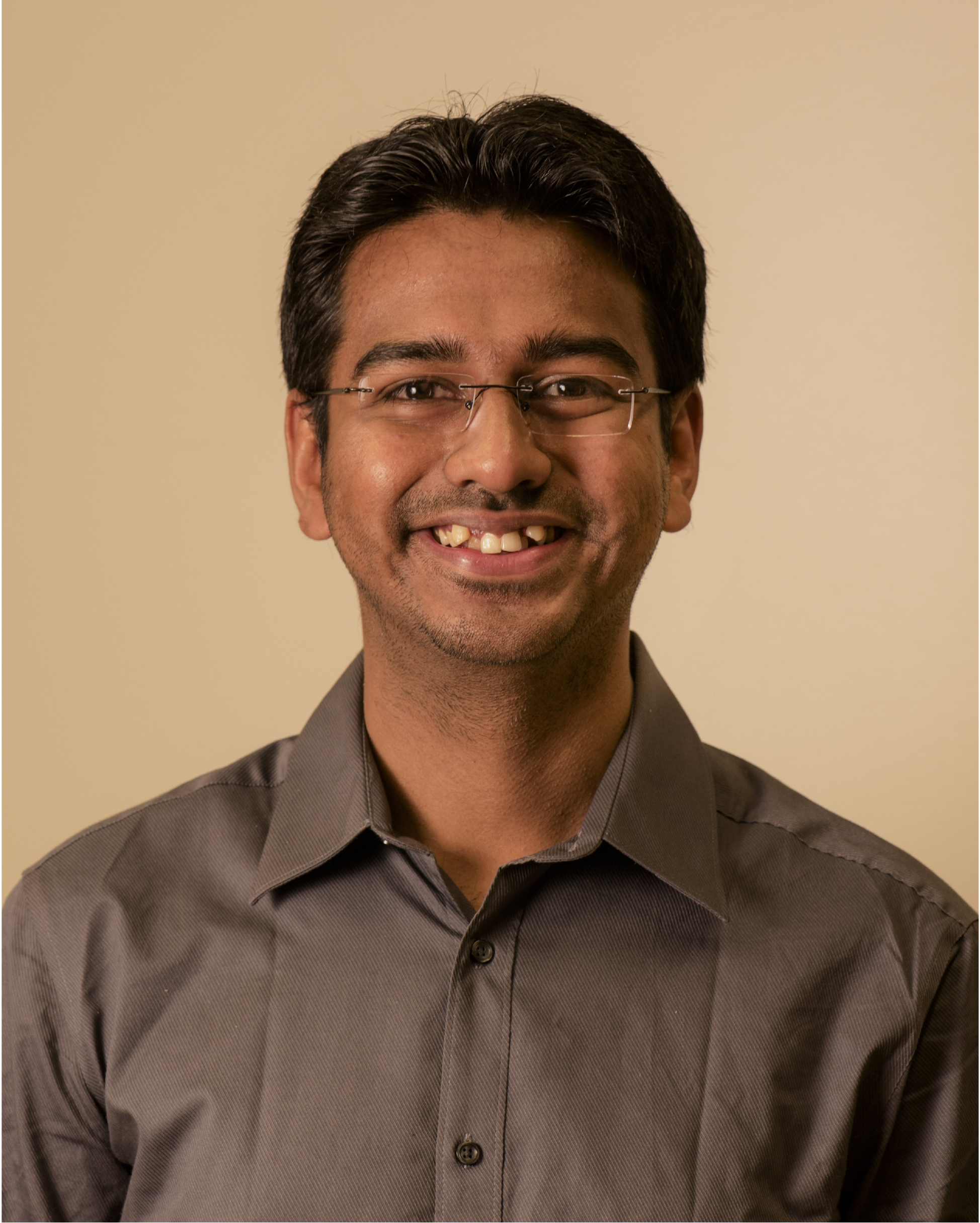}}]{Soumyabrata Dev} (S'09) graduated summa cum laude from National Institute of Technology Silchar, India with a B.Tech.\ in Electronics and Communication Engineering in 2010. Subsequently, he worked in Ericsson as a network engineer from 2010 to 2012. 

Currently, he is pursuing a Ph.D.\ degree in the School of Electrical and Electronic Engineering, Nanyang Technological University, Singapore. From Aug-Dec 2015, he was a visiting student at Audiovisual Communication Laboratory (LCAV), \'{E}cole Polytechnique F\'{e}d\'{e}rale de Lausanne (EPFL), Switzerland. His research interests include remote sensing, statistical image processing and machine learning. 
\end{IEEEbiography}

\begin{IEEEbiography}
[{\includegraphics[width=1in,height=1.25in,clip,keepaspectratio]{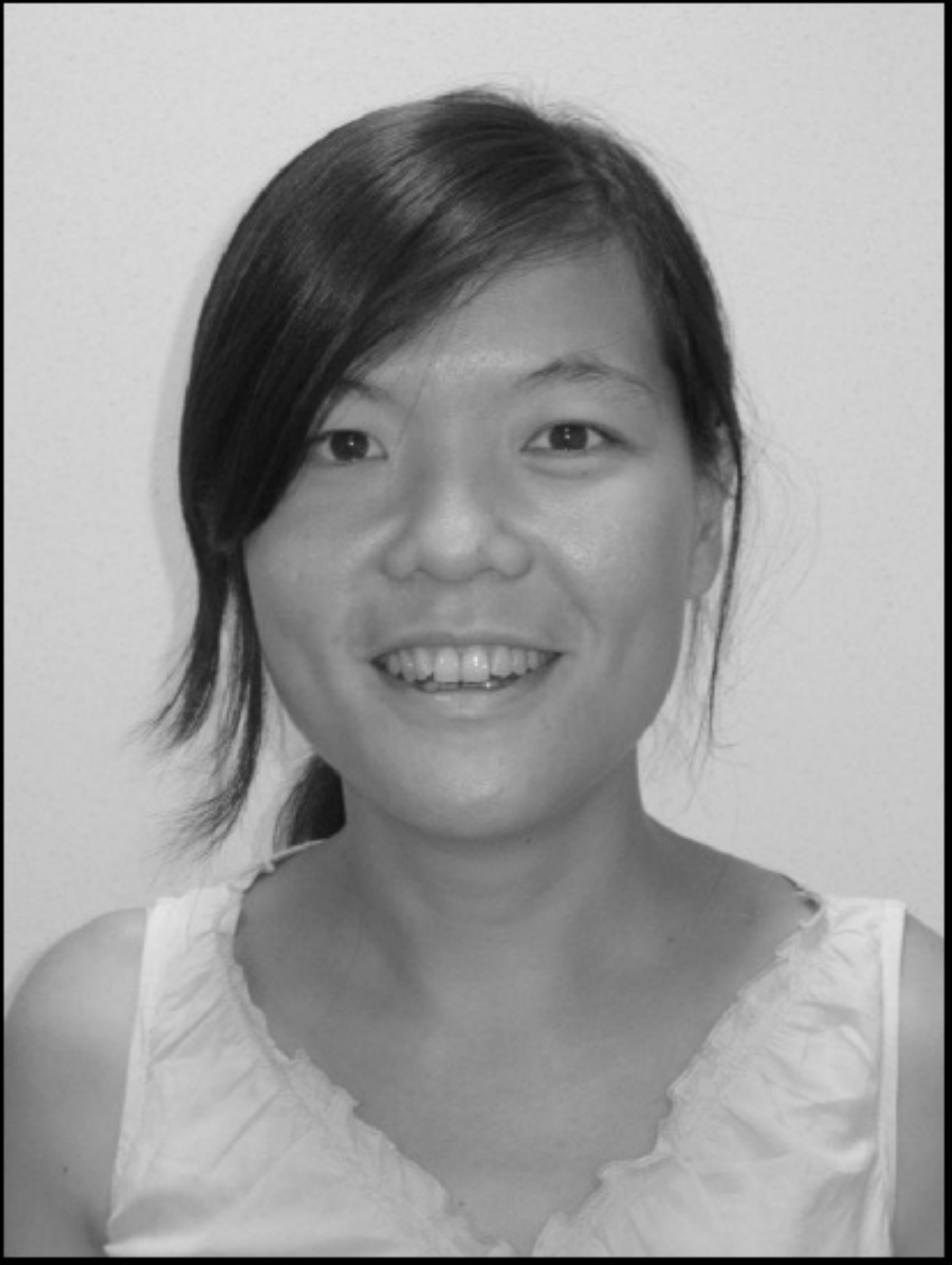}}]{Yee Hui Lee} (S'96-M'02-SM'11) received the B.Eng.\ (Hons.) and M.Eng.\ degrees from the School of Electrical and Electronics Engineering at  Nanyang Technological University, Singapore, in 1996 and 1998, respectively, and the Ph.D.\ degree from the University of York, UK, in 2002. 

Dr.\ Lee is currently Associate Professor and Assistant Chair (Students) at the School of Electrical and Electronic Engineering, Nanyang Technological University, where she has been a faculty member since 2002. Her interests are channel characterization, rain propagation, antenna design, electromagnetic bandgap structures, and evolutionary techniques.
\end{IEEEbiography}

\begin{IEEEbiography}
[{\includegraphics[width=1in,height=1.25in,clip,keepaspectratio]{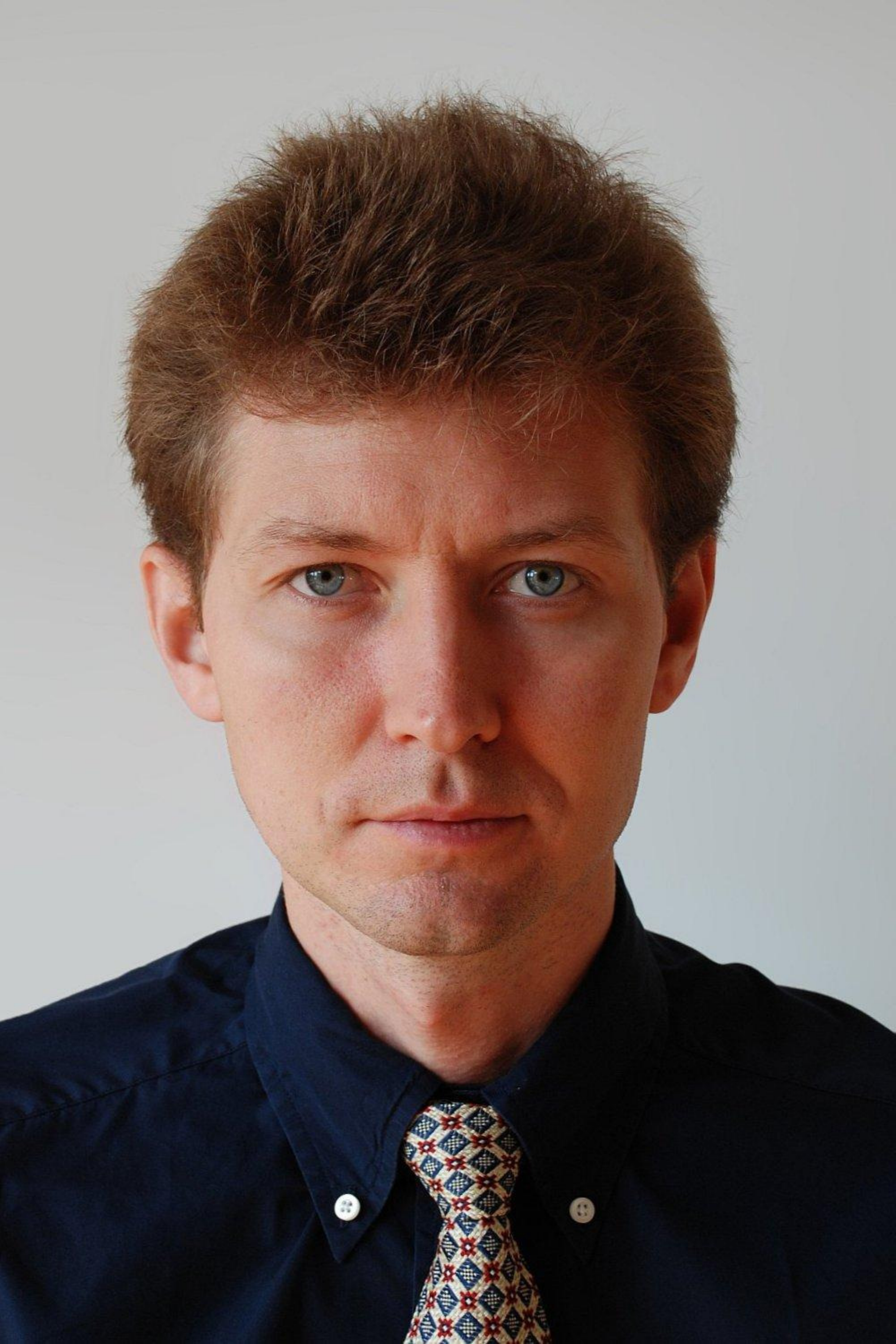}}]{Stefan Winkler} is Distinguished Scientist and Director of the Video \& Analytics Program at the Advanced Digital Sciences Center (ADSC), a joint research center between A*STAR and the University of Illinois. Prior to that, he co-founded Genista, worked for Silicon Valley companies, and held faculty positions at the National University of Singapore and the University of Lausanne, Switzerland.

Dr.\ Winkler has a Ph.D.\ degree from the \'{E}cole Polytechnique F\'{e}d\'{e}rale de Lausanne (EPFL), Switzerland, and an M.Eng./B.Eng. degree from the University of Technology Vienna, Austria. He has published over 100 papers and the book "Digital Video Quality" (Wiley). He is an Associate Editor of the IEEE Transactions on Image Processing, a member of the IVMSP Technical Committee of the IEEE Signal Processing Society, and Chair of the IEEE Singapore Signal Processing Chapter. His research interests include video processing, computer vision, perception, and human-computer interaction.
	
\end{IEEEbiography}

\balance

\end{document}